\documentclass[twoside,11pt]{article}

\usepackage{subcaption}
\usepackage[abbrvbib, preprint]{jmlr2e}
\usepackage[utf8]{inputenc} 
\usepackage[T1]{fontenc}    
\usepackage{url}            
\usepackage{amsfonts}       
\usepackage{nicefrac}       
\usepackage{microtype}      
\usepackage{xcolor}         

\usepackage{amsmath}
\allowdisplaybreaks   
\usepackage{amsthm} 
\usepackage{mathtools}
\usepackage{mathrsfs}
\usepackage{bbm}
\usepackage{tasks}
\usepackage{wrapfig}
\usepackage[capitalise]{cleveref}
\usepackage{enumitem}
\usepackage{parskip}
\usepackage{booktabs}
\usepackage{pifont}
\usepackage{makecell}
\usepackage{multicol}
\usepackage{tikz-cd}
\usepackage{units}
\usepackage[ruled,linesnumbered]{algorithm2e}
\usepackage{titletoc}
\usepackage{appendix}
\usepackage{multirow}
\usepackage{colortbl}
\usepackage{crossreftools}
\usepackage{mdframed} 
\usepackage{chngcntr} 

\newmdenv[
  backgroundcolor=gray!10, 
  linecolor=gray!10,       
  linewidth=0pt,           
  roundcorner=8pt,         
  skipabove=6pt,           
  skipbelow=6pt,           
  innertopmargin=6pt,      
  innerbottommargin=6pt,   
  leftmargin=0pt,          
  rightmargin=0pt          
]{keybox}

\usepackage{listings}
\newcommand{\BlackBox}{\rule{1.5ex}{1.5ex}}  
\ifdefined\proof
    \renewenvironment{proof}{\par\noindent{\bf Proof\ }}{\hfill\BlackBox\\[2mm]}
\else
    \newenvironment{proof}{\par\noindent{\bf Proof\ }}{\hfill\BlackBox\\[2mm]}
\fi

\theoremstyle{plain}
\newtheorem{theorem}{Theorem}            
\newtheorem{proposition}[theorem]{Proposition}
\newtheorem{lemma}[theorem]{Lemma}
\newtheorem{corollary}[theorem]{Corollary}

\theoremstyle{definition}
\newtheorem{definition}[theorem]{Definition}

\theoremstyle{remark}
\newtheorem{remark}[theorem]{Remark}

\providecommand{\wcalM}{\widetilde{\mathcal{M}}}
\providecommand{\calN}{\mathcal{N}}

\providecommand{\bbZ}{\mathbb{Z}}

\providecommand{\bbD}{\mathbb{D}}

\providecommand{\sym}[1]{\mathcal{S}^{#1}}
\providecommand{\spd}[1]{\mathcal{S}^{#1}_{++}}

\providecommand{\bbR}[1]{\mathbb {R}^{#1}}

\providecommand{\bbRscalar}{\mathbb {R}}

\providecommand{\orth}[1]{\mathrm{O}({#1})}

\providecommand{\dist}{\operatorname{d}}
\providecommand{\rieexp}{\operatorname{Exp}}
\providecommand{\rielog}{\operatorname{Log}}

\providecommand{\pt}[2]{\operatorname{PT}_{#1 \rightarrow #2}}
\providecommand{\scrL}{\mathscr{L}}
\providecommand{\bbD}{\mathbb {D}}
\providecommand{\ln}{\operatorname{ln}}

\providecommand{\dlog}{\operatorname{Dlog}}

\providecommand{\sign}{\operatorname{sign}}

\providecommand{\id}{\mathbbm{1}}

\providecommand{\clog}{\psi_{\mathrm{LC}}} 
\providecommand{\fm}{\operatorname{FM}}
\providecommand{\wfm}{\operatorname{WFM}}
\providecommand{\argmin}{\operatorname{argmin}}

\providecommand{\mlog}{\operatorname{mlog}}
\providecommand{\mexp}{\operatorname{mexp}}

\providecommand{\calB}{\mathcal{B}}

\providecommand{\gyr}{\operatorname{gyr}}
\providecommand{\gyrinner}[2]{\left\langle #1, #2 \right\rangle_{\mathrm{gyr}}}
\providecommand{\gyrnorm}[1]{\left\| #1 \right\|_{\mathrm{gyr}}}
\providecommand{\gyrdist}{{\mathrm{d}}_{\mathrm{gyr}}}

\providecommand{\gyrw}{\widetilde{\operatorname{gyr}}}

\providecommand{\gyrnormw}[1]{\left\| #1 \right\|_{\widetilde{\mathrm{gyr}}}}
\providecommand{\gyrdistw}{{\mathrm{d}}_{\widetilde{\mathrm{gyr}}}}

\providecommand{\gE}{g^\mathrm{E}}
\providecommand{\inner}[2]{\left\langle #1,#2 \right\rangle}

\providecommand{\norm}[1]{\left\| #1 \right\|}

\providecommand{\Rzero}{\mathbf{0}}

\providecommand{\calM}{\mathcal{M}}
\providecommand{\calH}{\mathcal{H}}
\providecommand{\stiefel}[1]{\mathrm{St}(#1)}

\providecommand{\barcenter}{\operatorname{Bar}}
\providecommand{\struct}[1]{\left( #1 \right)}

\providecommand{\oplusLieLE}{\oplus^{\mathrm{LE}}}
\providecommand{\oplusLieLC}{\oplus^{\mathrm{LC}}}
\providecommand{\oplusGyrAI}{\oplus^{\mathrm{AI}}}

\providecommand{\chol}{\operatorname{Chol}}

\providecommand{\cor}[1]{\mathrm{Cor}^{+} ({#1})}
\providecommand{\coropt}{\operatorname{Cor}}

\providecommand{\covtocor}{\operatorname{Cor}}

\providecommand{\bbPPB}[1]{\mathbb{PP}^{#1}}

\providecommand{\grasonb}[1]{\mathrm{Gr}(#1)}
\providecommand{\graspp}[1]{\widetilde{\mathrm{Gr}}(#1)}
\providecommand{\rp}[1]{\mathbb{RP}^{#1}}

\providecommand{\rank}{\operatorname{rank}}
\providecommand{\idonb}{I_{p,n}}
\providecommand{\idpp}{\widetilde{I}_{p,n}}
\providecommand{\oplusGyrONB}{\oplus^{\mathrm{Gr}}}
\providecommand{\oplusGyrPP}{\widetilde{\oplus}^{\mathrm{Gr}}}
\providecommand{\ominusGyrONB}{\ominus^{\mathrm{Gr}}}

\providecommand{\odotGyrONB}{\odot^{\mathrm{Gr}}}

\providecommand{\frakU}{\mathfrak{U}}

\providecommand{\stiefel}[1]{\mathrm{St}(#1)}

\providecommand{\ccs}[1]{\mathcal{C}_{K}^{#1}}

\providecommand{\calMK}[1]{\mathcal{M}_{K}^{#1}}
\providecommand{\gMK}{g^{K}}
\providecommand{\oplusMK}{\oplus_{K}}
\providecommand{\oplusMK}{\oplus_{K}}

\providecommand{\stereo}[1]{\mathfrak{st} _{K}^{#1}}
\providecommand{\stoplus}{\oplus _{K}}
\providecommand{\stominus}{\ominus _{K}}
\providecommand{\stodot}{\odot _{K}}
\providecommand{\tank}{\tan_{K}}

\providecommand{\isoSTMK}[1]{\pi_{\stereo{#1} \to \calMK{#1}}}
\providecommand{\isoMKST}[1]{\pi_{\calMK{#1} \to \stereo{#1}}}

\providecommand{\sink}{\sin _{K}}
\providecommand{\cosk}{\cos _{K}}
\providecommand{\tank}{\tan _{K}}

\providecommand{\Knorm}[1]{\norm{#1}_{K}}
\providecommand{\Kinner}[2]{\left\langle #1, #2 \right\rangle_{K}}

\providecommand{\hs}[1]{\mathrm{H}\mathbb{S}^{#1}}

\providecommand{\pball}[1]{\mathbb{P}^{#1}_{K}}
\providecommand{\Moplus}{\oplus_\mathrm{M}}
\providecommand{\Mominus}{\ominus _\mathrm{M}}
\providecommand{\Modot}{\odot _\mathrm{M}}

\providecommand{\klein}[1]{\mathbb{K}^{#1}_K}
\providecommand{\Eoplus}{\oplus_\mathrm{E}}
\providecommand{\Eominus}{\ominus_\mathrm{E}}
\providecommand{\Eodot}{\odot _\mathrm{E}}

\providecommand{\bbh}[1]{\mathbb{H}^{#1}_K}
\providecommand{\Lnorm}[1]{\left\| #1 \right\|_{\mathcal{L}}}
\providecommand{\Linner}[2]{\left\langle #1, #2 \right\rangle_{\mathcal{L}}}
\providecommand{\MKzero}{\overline{\mathbf{0}}}

\providecommand{\MKoplus}{\oplus ^\mathcal{M} _K}
\providecommand{\MKominus}{\ominus ^\mathcal{M} _K}
\providecommand{\MKodot}{\odot ^\mathcal{M} _K}

\providecommand{\projhs}[1]{\mathbb{D}^{#1}_{K}}

\providecommand{\sphere}[1]{\mathbb{S}_{K}^{#1}}

\providecommand{\unitpball}[1]{\mathbb{P}^{#1}_{-1}}
\providecommand{\unitklein}[1]{\mathbb{K}^{#1}_{-1}}
\providecommand{\unitbbh}[1]{\mathbb{H}^{#1}_{-1}}
\providecommand{\unitsphere}[1]{\mathbb{S}^{#1}_{-1}}
\providecommand{\unitprojhs}[1]{\mathbb{D}^{#1}_{-1}}

\providecommand{\norm}[1]{\left\| #1 \right\|}

\providecommand{\gyrobn}{\operatorname{GyroBN}}

\newcommand{\cmark}{\textcolor{green}{\text{\ding{51}}}}%
\newcommand{\xmark}{\textcolor{red}{\text{\ding{55}}}}

\providecommand{\ie}{\emph{i.e.},\xspace}
\providecommand{\eg}{\emph{e.g.},\xspace}
\newcommand{\na}{\textcolor{gray}{N/A}}%

\providecommand{\red}[1]{\textcolor{red}{#1}}
\providecommand{\mypara}[1]{\textit{#1}}

\definecolor{HilightColor}{RGB}{220, 255, 240} 
\definecolor{RowGray}{gray}{0.925}

\crefname{equation}{Equation}{Equations}
\Crefname{equation}{Equation}{Equations}

\crefname{figure}{Figure}{Figures}
\Crefname{figure}{Figure}{Figures}

\crefname{table}{Table}{Tables}
\Crefname{table}{Table}{Tables}

\crefname{algocf}{Algorithm}{Algorithms}
\Crefname{algocf}{Algorithm}{Algorithms}

\crefname{section}{Section}{Sections}
\Crefname{section}{Section}{Sections}

\crefname{appendix}{Appendix}{Appendices}
\Crefname{appendix}{Appendix}{Appendices}

\crefname{theorem}{Theorem}{Theorems}
\Crefname{theorem}{Theorem}{Theorems}
\crefname{lemma}{Lemma}{Lemmas}
\Crefname{lemma}{Lemma}{Lemmas}
\crefname{definition}{Definition}{Definitions}
\Crefname{definition}{Definition}{Definitions}
\crefname{corollary}{Corollary}{Corollaries}
\Crefname{corollary}{Corollary}{Corollaries}
\crefname{remark}{Remark}{Remarks}
\Crefname{remark}{Remark}{Remarks}
\crefname{proposition}{Proposition}{Propositions}
\Crefname{proposition}{Proposition}{Propositions}
\crefname{proof}{Proof}{Proofs}
\Crefname{proof}{Proof}{Proofs}

\makeatletter
\newcommand{\eqnref}[1]{Equation~\ref{#1}}

\newcommand{\eqnrefs}[1]{%
  \begingroup
    \edef\eqn@list{#1}%
    \count@=0\relax
    \@for\eqn@tmp:=\eqn@list\do{\advance\count@ by 1\relax}%
    \ifnum\count@=1
      Equation~\ref{#1}%
    \else
      Equations~%
      \count256=0\relax
      \@for\eqn@tmp:=\eqn@list\do{%
        \advance\count256 by 1\relax
        \ifnum\count256<\numexpr\count@-0\relax
          \ref{\eqn@tmp}%
          \ifnum\count256<\numexpr\count@-1\relax
            ,\ 
          \else
            \ 
          \fi
        \else
          and~\ref{\eqn@tmp}%
        \fi
      }%
    \fi
  \endgroup
}
\makeatother

\input{link_proof}
\usepackage[acronym, toc, automake, hyperfirst=true]{glossaries-extra}

\newglossary*{notation}{List of Symbols}
\setglossarystyle{long} 
\setabbreviationstyle[acronym]{long-short} 

\makeglossaries 

\newacronym[sort=nn]{CNNs}{CNNs}{Convolutional Neural Networks}
\newacronym[sort=nn]{RNNs}{RNNs}{Recurrent Neural Networks}
\newacronym[sort=nn]{DNNs}{DNNs}{Deep Neural Networks}
\newacronym[sort=nn]{FC}{FC}{Fully Connected}
\newacronym[sort=nn]{MLR}{MLR}{Multinomial Logistics Regression}

\newacronym[sort=nn]{SPDNet}{SPDNet}{SPD Neural Network}
\newacronym[sort=nn]{GrNet}{SPDNet}{Grassmann network}
\newacronym[sort=nn]{HNN}{HNN}{Hyperbolic Neural Network}
\newacronym[sort=nn]{HNN++}{HNN++}{Hyperbolic Neural Network++}

\newacronym[sort=bn]{BN}{BN}{Batch Normalization}
\newacronym[sort=bn]{RBN}{RBN}{Riemannian Batch Normalization}
\newacronym[sort=bn]{LieBN}{LieBN}{Lie Group Batch Normalization}
\newacronym[sort=bn]{SPDBN}{SPDBN}{SPD Batch Normalization}
\newacronym[sort=bn]{GyroBN}{GyroBN}{Gyrogroup Batch Normalization}

\newacronym[sort=spd]{SPD}{SPD}{Symmetric Positive Definite}
\newacronym[sort=spd]{AIM}{AIM}{Affine-Invariant Metric}
\newacronym[sort=spd]{LCM}{LCM}{Log-Cholesky Metric}
\newacronym[sort=spd]{LEM}{LEM}{Log-Euclidean Metric}

\newacronym[sort=gr]{ONB}{ONB}{Orthonormal Basis}
\newacronym[sort=gr]{PP}{PP}{Projector Perspective}

\newacronym[sort=cor]{ECM}{ECM}{Euclidean--Cholesky Metric}
\newacronym[sort=cor]{LECM}{LECM}{Log-Euclidean--Cholesky Metric}
\newacronym[sort=cor]{LSM}{LSM}{Log-Scaled Metric}
\newacronym[sort=cor]{OLM}{OLM}{Off-Log Metric}
\newacronym[sort=cor]{PHCM}{PHCM}{Poly-Hyperbolic-Cholesky Metric}

\newglossaryentry{manifold}{
  type=notation, 
  name={$\mathcal{M}$}, 
  text={$\mathcal{M}$}, 
  description={Riemannian manifold}, 
}

\newglossaryentry{metric}{
  type=notation,
  name={$\mathbf{g}$},
  text={$\mathbf{g}$},
  description={Metric on $\mathcal{M}$},
}

\newglossaryentry{reals}{
  type=notation,
  name={$\mathbb{R}^n$},
  text={$\mathbb{R}^n$},
  description={A $n$-dimensional Euclidean space},
}

\newcommand{\glsfull}[1]{\glsentrylong{#1} (\glsentryshort{#1})}

\usepackage{lastpage}
\jmlrheading{26}{2025}{1-\pageref{LastPage}}{9/25; Revised 5/22}{9/22}{21-0000}{Ziheng Chen, Xiao-Jun Wu \& Nicu Sebe}

\ShortHeadings{GyroBN}{Chen, Wu, and Sebe}
\firstpageno{1}

\begin{document}

\title{Riemannian Batch Normalization: A Gyro Approach}

\author{\name Ziheng Chen\thanks{Corresponding Author.} \email ziheng\_ch@163.com \\
       \addr University of Trento, Italy\\
       \AND
       \name Xiao-Jun Wu \email wu\_xiaojun@jiangnan.edu.cn \\
       \addr Jiangnan University, China
       \AND
       \name Bernhard Schölkopf \email bernhard.schoelkopf@tuebingen.mpg.de \\
       \addr Max Planck Institute for Intelligent Systems, Germany
       \AND
       \name Nicu Sebe \email niculae.sebe@unitn.it \\
       \addr University of Trento, Italy\\
       }

\editor{My editor}

\maketitle

\begin{abstract}%
Normalization layers are crucial for deep learning, but their Euclidean formulations are inadequate for data on manifolds. On the other hand, many Riemannian manifolds in machine learning admit gyro-structures, enabling principled extensions of Euclidean neural networks to non-Euclidean domains. Inspired by this, we introduce GyroBN, a principled Riemannian batch normalization framework for gyrogroups. We establish two necessary conditions, namely \emph{pseudo-reduction} and \emph{gyroisometric gyrations}, that guarantee GyroBN with theoretical control over sample statistics, and show that these conditions hold for all known gyrogroups in machine learning. Our framework also incorporates several existing Riemannian normalization methods as special cases. We further instantiate GyroBN on seven representative geometries, including the Grassmannian, five constant curvature spaces, and the correlation manifold, and derive novel gyro and Riemannian structures to enable these instantiations. Experiments across these geometries demonstrate the effectiveness of GyroBN. The code is available at \url{https://github.com/GitZH-Chen/GyroBN.git}.
\end{abstract}

\section{Introduction}
\label{sec:introduction}

\gls{DNNs} on Riemannian manifolds have attracted increasing attention in diverse machine learning applications, such as computer vision~\citep{huang2017deep,huang2017riemannian,huang2018building,skopek2019mixed,chen2023riemannian,gao2023exploring,wang2024spd,chen2025understanding}, natural language processing \citep{ganea2018hyperbolic,shimizu2020hyperbolic}, drone classification \citep{brooks2019riemannian,chen2024spdmlr}, human neuroimaging \citep{pan2022matt,kobler2022spd,bonet2023sliced,ju2024deep,wang2024grassatt,li2025spdim,hu2025correlation}, medical imaging \citep{huang2019manifold,chakraborty2020manifoldnet}, node and graph classification \citep{chami2019hyperbolic,dai2021hyperbolic,zhao2023modeling,chen2023distribution,chen2024rmlr}, and genome sequence modeling \citep{khan2025hyperbolic}. As in the Euclidean setting, normalization layers~\citep{ioffe2015batch,ba2016layer,ulyanov2016instance,wu2018group} play a central role in facilitating training, motivating their adaptation to manifold-valued data.

Despite this need, most existing Riemannian normalization methods are restricted to specific geometries or cannot normalize sample statistics. For example, \citet{brooks2019riemannian,kobler2022spd} introduced \gls{SPDBN} on \gls{SPD} manifolds under the specific \gls{AIM}, while \citet{chakraborty2020manifoldnorm} proposed a \gls{RBN} framework for homogeneous spaces. However, these approaches often fail to guarantee principled control of sample mean and variance. Similar limitations appear in \citet{lou2020differentiating} and \citet{bdeir2024fully}. \citet{chen2024liebn} further developed the RBN for general Lie groups, termed \gls{LieBN}. Although LieBN can normalize sample statistics, many important geometries in machine learning do not admit a Lie group structure. As a result, existing methods still lack a principled solution for Riemannian normalization.

Recently, gyro-structures have emerged as effective tools for building Riemannian networks across various geometries, including SPD~\citep{nguyen2022gyrovector,nguyen2022gyro}, Grassmannian~\citep{nguyen2022gyro}, hyperbolic~\citep{ganea2018hyperbolic}, and spherical manifolds~\citep{skopek2019mixed}. They naturally extend Euclidean vector structures while encompassing Lie groups and non-group geometries. For instance, the Grassmannian, hyperbolic, and spherical manifolds do not form Lie groups but instead form gyrogroups.

\begin{figure}[tbp]
\centering
\includegraphics[width=0.9\linewidth,trim={0cm 0.5cm 0cm 0.5cm}]{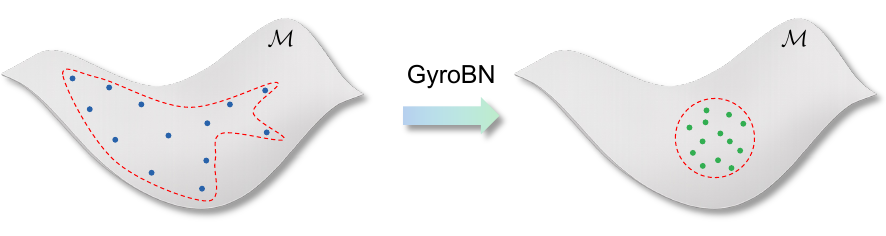}
\caption{Illustration of GyroBN on manifold-valued data. Blue points, green points, and the red dashed curves indicate the input samples, normalized outputs, and data distributions, respectively.}
\label{fig:illustration_gyrobn}
\vspace{-4mm}
\end{figure}

\begin{table}[tbp]
    \centering
    \resizebox{0.99\linewidth}{!}{
    \begin{tabular}{cccc}
        \toprule
         Method & Controllable Statistics & Applied Geometries & Incorporated by GyroBN\\
         \midrule
         \makecell{SPDBN \\ \citep{brooks2019riemannian}} &  M & SPD manifolds under AIM & \cmark\\
         \makecell{SPDBN \\ \citep{kobler2022controlling}}
           & M+V & SPD manifolds under AIM & \cmark\\
         \makecell{SPDDSMBN \\ \citep{kobler2022spd}} & M+V & SPD manifolds under AIM & \cmark\\
         \makecell{ManifoldNorm \\ \citep[Algorithms 1--2]{chakraborty2020manifoldnorm}}  & \textbf{\textcolor{red}{N/A}} & Riemannian homogeneous space & \xmark\\
         \makecell{ManifoldNorm \\ \citep[Algorithms 3--4]{chakraborty2020manifoldnorm}} & M+V & \makecell{Matrix Lie groups under the distance \\
         $d(X, Y)=\left\|\mlog \left(X^{-1} Y\right)\right\|$}  & \cmark\\
         \makecell{ RBN \\ \citep[Algorithm 2]{lou2020differentiating}} & \textbf{\textcolor{red}{N/A}} & Geodesically complete manifolds & \xmark\\
         \makecell{ LieBN  \\ \citep{chen2024liebn}} & M+V & Lie groups & \cmark\\
         \midrule
         GyroBN & M+V & \makecell{ Pseudo-reductive gyrogroups \\ with gyroisometric gyrations } & \na \\
         \bottomrule
    \end{tabular}  
    }
     \caption{Comparison of previous RBN methods with our GyroBN, where M and V denote the sample mean and variance.}
    \label{tab:rbn_summary}
    \vspace{-3mm}
\end{table}

Based on the analysis above, this paper introduces \gls{GyroBN}, a general RBN framework on gyrogroups, as illustrated in \cref{fig:illustration_gyrobn}. We employ gyrosubtraction, gyroaddition, and scalar gyromultiplication to generalize the centering (vector subtraction), biasing (vector addition), and scaling (scalar multiplication) in Euclidean \gls{BN} to curved manifolds in a principled manner. We clarify why centering and biasing in GyroBN rely on \emph{left} gyroaddition, rather than other candidates, such as right gyroaddition or gyrocoaddition~\citep[Definition.~2.9]{ungar2022analytic}. To broaden the scope, we relax the classical gyrogroup to the \emph{pseudo-reductive gyrogroup}, and show that when gyrations are \emph{gyroisometries}, GyroBN enjoys theoretical control over sample statistics. These conditions are satisfied by all known gyrogroups in machine learning, providing a principled and unified normalization mechanism. Moreover, several existing RBN methods arise as special cases of GyroBN, including LieBN on Lie groups and various SPD-based variants, as summarized in \cref{tab:rbn_summary}.

On the implementation side, we instantiate GyroBN on seven representative geometries: the Grassmannian~\citep{bendokat2024grassmann}, five constant curvature spaces~\citep{ganea2018hyperbolic,lee2018introduction,bachmann2020constant}, and the full-rank correlation manifold~\citep{thanwerdas2022theoretically}. For the Grassmannian, we propose an efficient implementation. For constant curvature spaces, we cover five models: Poincaré ball, hyperboloid, Beltrami--Klein, sphere, and projected hypersphere. To enable these instantiations, we refine the projected hypersphere structure~\citep{bachmann2020constant}, derive closed-form gyro-structures for the hyperboloid and sphere, and develop the Riemannian structure of the Beltrami--Klein model. For the correlation manifold, we demonstrate that its gyro-structure can be defined row-wise on its Cholesky factor. To facilitate adoption, we release a PyTorch-compatible toolbox~\citep{paszke2019pytorch} with drop-in GyroBN layers, illustrated in \cref{fig:gyrobn_minimal_examples}. Experiments on networks over these seven geometries validate the effectiveness of our framework.

\begin{figure}[tb]
\begin{lstlisting}[language=Python]
from GyroBN import *
from GyroBN.Geometry import *

# ==== Grassmannian ====
manifold = GrassmannianGyro(n=50, p=10)
X_gr = manifold.random_normal(30, 50, 10)
gybn_gr = GyroBNGr(shape=[50, 10])
out_gr = gybn_gr(X_gr)

# ==== Five Constant Curvature Spaces (CCSs) ====
models = [
    ("Poincare",    Stereographic(K=-1.0)),
    ("Hyperboloid", Hyperboloid(K=-1.0)),
    ("Klein",       Klein(K=-1.0)),
    ("Sphere",      Sphere(K= 1.0)),
    ("ProjSphere",  Stereographic(K= 1.0)),
]
for name, manifold in models:
    X_ccs = manifold.random_normal(30, 16)
    gybn_ccs = GyroBNCCS(shape=[16], model=name, K=manifold.K)
    out_ccs = gybn_ccs(X_ccs)

# ==== Full-Rank Correlation ====
manifold = CorPolyHyperbolicCholeskyMetric(n=10)
X_cor = manifold.random(30, 10, 10)
gybn_cor = GyroBNCor(shape=[10, 10])
out_cor = gybn_cor(X_cor)
\end{lstlisting}
\caption{Minimal examples of applying GyroBN.}
\label{fig:gyrobn_minimal_examples}
\vspace{-6mm}
\end{figure}

In summary, our \emph{main contributions} are
\begin{itemize}[itemsep=0pt, topsep=2pt, parsep=2pt]
    \item Theoretical foundation: pseudo-reductive gyrogroups as a relaxation of classical gyrogroups;
    \item General framework: GyroBN as a plug-and-play normalization mechanism, with pseudo-reduction and gyroisometric gyrations ensuring theoretical control of batch statistics;
    \item Geometric insights: refined projected hypersphere gyro-structure, closed-form hyperboloid and sphere gyro-structures, Riemannian structure of the Beltrami--Klein model, and row-wise correlation manifold gyro-structure;
    \item Practical instantiations: implementations on the Grassmannian, five constant curvature spaces, and the correlation manifold with extensive experiments.
\end{itemize}

\mypara{Outline.}
\cref{sec:preliminaries} reviews the background on Riemannian geometry, gyro-structures, and the manifolds considered in this work. \cref{sec:pseudo-reductive_gyrogroups} introduces pseudo-reductive gyrogroups and analyzes their theoretical properties,  while \cref{sec:gyrobn-general} develops the GyroBN framework. \cref{sec:instantiations} shows that prior RBN methods are our special cases, and instantiate our GyroBN on seven representative geometries. \cref{sec:exp} reports experiments that validate GyroBN across these geometries. \cref{sec:conclusion} concludes the paper. All proofs are deferred to \cref{app:sec:proofs}.

This paper extends our ICLR work~\citep{chen2025gyrogroup} both methodologically and in implementation. On the methodological side, we provide justification for why GyroBN relies on left gyroaddition. On the implementation side, we broaden the framework beyond the Grassmannian and Poincaré geometries to additional constant curvature spaces and the correlation manifold, introducing novel gyro and Riemannian structures. We further expand the experiments and release an open-source PyTorch toolbox.

\section{Preliminaries}
\label{sec:preliminaries}
This section reviews the necessary background on Riemannian geometry, gyro-structures, and concrete gyrogroups in machine learning. For in-depth discussion of Riemannian manifolds and gyrospaces, we refer the reader to \citet{lee2018introduction} and \citet{ungar2022analytic}, respectively.

\subsection{Riemannian Geometry}
\label{subsec:riem-geometry}

\mypara{Notation.}
For the Euclidean space $\bbR{n}$ (resp. $\bbR{n \times n}$), we denote $\inner{\cdot}{\cdot}$ as the standard Euclidean inner product on vectors (resp. Frobenius inner product on matrices), with $\norm{\cdot}$ as the induced norm: the $L_2$ norm for vectors and the Frobenius norm for matrices. The zero vector and zero matrix are denoted uniformly by $\Rzero$. For points on general manifolds, we use lowercase letters (\eg $x,y,z$) unless the elements are matrix-valued, in which case we use uppercase letters (\eg $P,Q,R$). A summary of notations is provided in \cref{app:notations}.

\mypara{Riemannian Manifold.}
A Riemannian manifold $(\calM,g)$, abbreviated as $\calM$, carries a smoothly varying Riemannian metric $g_x: T_x\calM \times T_x\calM \to \bbRscalar$ on each tangent space $T_x\calM$. The induced norm is $\|v\|_x = \sqrt{g_x(v,v)}$. As an inner product, $g_x$ is also denoted as $\inner{\cdot}{\cdot}_x$. For a smooth curve $\gamma:[0,1]\to\calM$, its length is $L(\gamma)=\int_0^1 \norm{\dot\gamma(t) }_{\gamma(t)} \mathrm{d}t$, and the associated geodesic distance is $\dist(x,y) = \inf_{\gamma(0)=x,\,\gamma(1)=y} L(\gamma)$. 

\mypara{Geodesic.}
Straight lines are generalized to constant-speed curves that are locally length-minimizing between points $x, y \in \calM$, known as geodesics:
\begin{equation*}
    \gamma^* = \arg \min_\gamma L(\gamma) 
    \quad \text{subject to } \gamma(0) = x,\; \gamma(1) = y,\; 
    \norm{\dot\gamma(t) }_{\gamma(t)} = 1.
\end{equation*}

\mypara{Exponential and Logarithmic Maps.}
For $x \in \calM$ and $v \in T_x\calM$, let $\gamma_{x,v}$ denote the unique geodesic with $\gamma_{x,v}(0) = x$ and $\dot\gamma_{x,v}(0) = v$. 
The exponential map $\rieexp_x : T_x\calM \supset \mathcal{V} \to \calM$ is defined by $\rieexp_x(v) = \gamma_{x,v}(1)$, where $\mathcal{V}$ is an open neighborhood of the origin in $T_x\calM$. Its local inverse, defined for $y$ in a neighborhood $\mathcal{U} \subset \calM$ of $x$, is the logarithmic map $\rielog_x : \mathcal{U} \to T_x\calM$, satisfying $\rieexp_x \circ \rielog_x = \id_{\mathcal{U}}$. On Cartan--Hadamard manifolds, that is, complete and connected Riemannian manifolds with non-positive sectional curvature, the exponential and logarithmic maps are globally defined~\citep[Proposition~12.19]{lee2018introduction}. Representative cases include SPD, hyperbolic, and correlation manifolds. Throughout this paper, we assume that $\rieexp_x$ and $\rielog_x$ are globally defined, with numerical remedies adopted when singularities arise.

\mypara{Parallel Transport.}
Given a geodesic $\gamma$ from $x$ to $y$, the parallel transport of a tangent vector $v \in T_x\calM$ is the unique vector $\pt{x}{y} (v) \in T_y\calM$ obtained by transporting $v$ along $\gamma$ so that its covariant derivative along $\gamma$ vanishes. Parallel transport defines a linear isometry between $T_x\calM$ and $T_y\calM$.

\cref{tb:reinter_riem_operators} compares the corresponding operators in Euclidean and Riemannian geometries.

\mypara{Isometry.}
The isometries generalize the bijection into the Riemannian geometry. If $\{\calM, g\}$ and $\{\widetilde{\calM}, \widetilde{g}\}$ are both Riemannian manifolds, a smooth map $f: \calM \rightarrow$ $\widetilde{\calM}$ is called a (Riemannian) isometry if it is a diffeomorphism that satisfies 
\begin{equation*}
    g_x(v,w) = \widetilde{g}_{f(x)}(f_{*,x}(v),f_{*,x}(w)),
\end{equation*}
where $f_{*,x}(\cdot): T_x \calM \rightarrow T _{f(x)} \widetilde{\calM}$ is the differential map of $f$ at $x \in \calM$, and $v,w \in T_x\calM$ are two tangent vectors.

\mypara{Lie Group.}
A manifold is a Lie group, if it forms a group with a group operation $\odot$ such that $m(x,y) \mapsto x \odot y$ and $i(x) \mapsto x_{\odot}^{-1}$ are both smooth, where $x_{\odot}^{-1}$ is the group inverse of $x$. 

\begin{table}[t]
    \centering
    \begin{tabular}{ccc}
    \toprule
    Operation & Euclidean space & Riemannian manifold \\
    \midrule
    Straight line & Straight line & Geodesic \\
    Subtraction & $\overrightarrow{x y}=y-x$ & $\overrightarrow{x y}=\rielog _x(y)$ \\
    Addition & $y=x+\overrightarrow{x y}$ & $y=\rieexp _x(\overrightarrow{x y})$ \\
    Parallelly moving & $v \rightarrow v$ & $\pt{x}{y}(v)$\\
    \bottomrule
    \end{tabular}
    \caption{The geometric reinterpretations of Riemannian operators.}
    \label{tb:reinter_riem_operators}
\end{table}

\subsection{Gyro-Structures}

\begin{definition}[Gyrogroups \citep{ungar2022analytic}]
    \label{def:gyrogroups}
    Given a nonempty set $G$ with a binary operation $\oplus: G \times G \rightarrow G$, $(G, \oplus )$ forms a gyrogroup if its binary operation satisfies the following axioms for any $x, y, z \in G$ :
    
    \noindent (G1) There is at least one element $e \in G$ called a left identity (or neutral element) such that $e \oplus x = x$.
    
    \noindent (G2) There is an element $\ominus x \in G$ called a left inverse of $x$ such that $\ominus x \oplus x = e$.
    
    \noindent (G3) There is an automorphism $\gyr[x, y]: G \rightarrow G$ for each $x, y \in G$ such that
    \begin{equation*}
        x \oplus (y \oplus z) = (x \oplus y) \oplus \gyr[x, y] z \quad \text { (Left Gyroassociative Law). }
    \end{equation*}
    The automorphism $\gyr[x, y]$ is called the gyroautomorphism, or the gyration of $G$ generated by $x, y$. 
    
    \noindent (G4) Left reduction law: $\gyr[x, y]=\gyr[x \oplus y, y]$.
\end{definition}

\begin{definition}[Gyrocommutative Gyrogroups \citep{ungar2022analytic}]
    \label{def:gyrocommutative_gyrogroups}
    A gyrogroup $(G, \oplus)$ is gyrocommutative if it satisfies
    \begin{equation*}
        x \oplus y = \gyr[x, y](y \oplus x) \quad \text { (Gyrocommutative Law). }
    \end{equation*}
\end{definition}

\begin{definition}[Nonreductive Gyrogroups \citep{nguyen2022gyro}]
    \label{def:nonreductive_gyrogroup}
    A groupoid $(G, \oplus )$ is a nonreductive gyrogroup if it satisfies axioms (G1), (G2), and (G3).
\end{definition}

\begin{definition}[Gyrovector Spaces]
\label{def:gyrovector_spaces}
A gyrocommutative gyrogroup $(G, \oplus)$ equipped with a scalar gyromultiplication $\odot : \mathbb{R} \times G \rightarrow G$
is called a gyrovector space if it satisfies the following axioms for $s, t \in \mathbb{R}$ and $x, y, z \in G$:
\\ \noindent (V1) Identity Scalar Multiplication:
$1 \odot x = x$.
\\ \noindent (V2) Scalar Distributive Law:
$(s+t) \odot x = s \odot x \oplus t \odot x$.
\\ \noindent (V3) Scalar Associative Law:
$(s t) \odot x = s \odot (t \odot x)$.
\\ \noindent (V4) Gyroautomorphism:
$\gyr[x, y](t \odot z) = t \odot \gyr[x, y] z$.
\\ \noindent (V5) Identity Gyroautomorphism:
$\gyr[s \odot x, t \odot x] = \id$, where $\id$ is the identity map.
\end{definition}
\begin{remark}
    \citet{nguyen2022gyro} presented a similar definition, except that (V1) is defined as $1 \odot x = x, \; 0 \odot x = t \odot e = e, \; \text{and } (-1) \odot x = \ominus x$. However, as implied by \citet[Theorem~6.4]{ungar2022analytic}, $0 \odot x = t \odot e = e, \; (-1) \odot x = \ominus x$ are redundant.
\end{remark}

Intuitively, gyrogroups are natural generalizations of groups. Unlike groups, gyrogroups are non-associative but have gyroassociativity characterized by gyrations. Since gyrations in any (Lie) group are the identity map, every (Lie) group is automatically a gyrogroup. Similarly, the gyrovector space generalizes the vector space, which has shown impressive success in hyperbolic geometry \citep{ungar2022analytic}. Although this work employs scalar gyromultiplication, we do not require it to satisfy the axioms of a gyrovector space. Nevertheless, we present \cref{def:gyrovector_spaces} for completeness.

\subsection{Gyro-Structures over Manifolds}
As shown by \citet{nguyen2023building}, for $x,y,z \in \calM$ and $t \in \bbRscalar$, the gyro-structure is defined as\footnote{\citet{nguyen2023building} assume all involved Riemannian operators are well-defined. For each singular cases in this paper, we will provide numerical solutions, which will be discussed in \cref{sec:instantiations}.}  
\begin{align}
\label{eq:gyro_addtion}
\text{Gyroaddition: } x \oplus y &= \rieexp_{x}\left(\pt{e}{x} \left(\rielog _{e}(y)\right)\right), \\
\label{eq:gyro_scalar_product}
\text{Scalar gyromultiplication: } t \odot x &= \rieexp_{e}\left(t \rielog _{e}(x)\right), \\
\label{eq:gyro_inverse}
\text{Gyroinverse: } \ominus x &= -1 \odot x = \rieexp_{e}\left(- \rielog _{e}(x)\right), \\
\label{eq:gyro_automorphism}
\text{Gyration: }\gyr[x, y] z &= (\ominus(x \oplus y)) \oplus(x \oplus(y \oplus z)),\\
\label{eq:gyro_inner_product}
\text{Gyro inner product: } \gyrinner{x}{y}&=\left\langle \rielog _e (x), \rielog _e (y)\right\rangle_{e},\\
\label{eq:gyro_norm}
\text{Gyronorm: } \gyrnorm{x} &= \gyrinner{x}{x},\\
\label{eq:gyro_distance}
\text{Gyrodistance: } \gyrdist(x, y) &= \gyrnorm{\ominus x \oplus y} ,
\end{align}
where $e$ is the gyro identity element, and $\rielog _e$ and $\langle \cdot,\cdot \rangle_e$ is the Riemannian logarithm and metric at $e$.
A bijection $\omega: G \rightarrow G$ is called gyroisometry, if it preserves the gyrodistance:
\begin{equation*}
    \gyrdist(\omega(x), \omega(y))=\gyrdist(x, y).
\end{equation*}

\subsection{Examples}
\label{subsec:preliminaries-examples}

Several geometries in machine learning admit (nonreductive) gyrogroups.

\mypara{SPD Manifold \citep{pennec2006riemannian}.}
The set $\spd{n}$ of $n \times n$ SPD matrices form a manifold, named the SPD manifold. We focus on three popular Riemannian metrics on the SPD manifold: Affine-Invariant Metric (AIM) \citep{pennec2006riemannian}, \gls{LEM} \citep{arsigny2005fast}, and \gls{LCM} \citep{lin2019riemannian}. As shown by \citet[Section 3.1]{nguyen2022gyro}, each metric can induce a distinct gyro-structure via \cref{eq:gyro_addtion}--\cref{eq:gyro_distance}. Besides, the gyrogroups under the LEM and LCM coincide with the Lie groups proposed by \citet{arsigny2005fast,lin2019riemannian}, respectively.

\mypara{Grassmannian Manifold \citep{bendokat2024grassmann}.}
The Grassmannian is the set of $p$-dimensional subspace of $n$-dimensional vector space. It has two matrix representations, the \gls{PP} and \gls{ONB} perspective:
\begin{equation*}
    \begin{aligned}
        \text{PP: }& \graspp{p,n} = \{P \in \sym{n} : P^2=P, \quad \rank(P)=p \}, \\
        \text{ONB: }& \grasonb{p,n} = \{ [U] : [U]:= \{\widetilde{U} \in \stiefel{p, n} \mid \widetilde{U}=U R, \quad R \in \orth{p}\} \},
    \end{aligned}
\end{equation*}
where $\sym{n}$ is the Euclidean space of symmetric matrices, $\stiefel{p, n}$ is the Stiefel manifold, and $\orth{p}$ is the orthogonal group.
By abuse of notations, we use $[U]$ and $U$ interchangeably for the element of $\grasonb{p,n}$, where the $n \times p$ column-wise orthonormal matrix $U$ should be taken as a representative of an equivalence class. \citet{helmke2012optimization} show that the ONB perspective is diffeomorphic to the PP by 
\begin{equation} \label{eq:iso_grass}
    \pi: \grasonb{p,n} \ni U \mapsto UU^\top \in \graspp{p,n}.
\end{equation}
As shown by \citet[Section 3.2]{nguyen2022gyro} and \citet[Section 2.3.1]{nguyen2023building}, the Grassmannian admits a gyrocommutative and nonreductive gyrogroup defined by \cref{eq:gyro_addtion}--\cref{eq:gyro_distance}.

\mypara{$K$-Stereographic Model \citep{bachmann2020constant}.}
It is defined as
\begin{equation*}
    \stereo{n} = \left\{x \in \bbR{n} : \norm{x}^2 < -\nicefrac{1}{K} \right\}, \text{ with }\gMK_x = \left(\lambda_{x}^K\right)^2 \gE, 
\end{equation*}
where $K \in \bbRscalar$ is the constant curvature, $\lambda^K_x =\frac{2}{\left(1+K\|x\|^2\right)}$ is a conformal factor, $\gMK$ is its Riemannian metric, and $\gE=\inner{}{}$ is the standard inner product. Particularly, $\stereo{n}$ is the scaled $\bbR{n}$ when $K = 0$. It unifies the spherical projected hypersphere $\projhs{n}$, Euclidean $\bbR{n}$, and hyperbolic Poincaré ball $\pball{n}$:
\begin{equation*}
    \stereo{n} = 
    \begin{cases}
    \projhs{n} = \bbR{n}, &  \text{For } K>0, \text{ spherical geometry,} \\ 
    \bbR{n}, &  \text{For } K=0, \text{ Euclidean geometry,}\\
    \pball{n} = \left\{x\in \bbR{n}: \norm{x}^2 < -\nicefrac{1}{K} \right \}, &  \text{For } K<0, \text{ hyperbolic geometry.}\\
    \end{cases}
\end{equation*}
Although $\projhs{n}=\bbR{n}$ for $K < 0$, its metric is conformal to the Euclidean one. We will abbreviate the $K$-stereographic model as the stereographic model. \citet[Equations. 2--3]{bachmann2020constant} shows that this model admits a gyro-structure.

\mypara{$K$-Radius Model \citep{skopek2019mixed}.} 
It provides an extrinsic representation of the space with constant curvature 
$K \in \bbRscalar$, encompassing the sphere $\sphere{n}$, Euclidean space $\bbR{n}$, and hyperboloid $\bbh{n}$:
\begin{equation*}
    \calMK{n} = 
    \begin{cases}
    \sphere{n} =\left\{x\in \bbR{n+1}: \norm{x}^2 = \frac{1}{K} \right\}, &  \text{For } K>0, \text{ spherical geometry,} \\ 
    \bbR{n}, &  \text{For } K=0, \text{ Euclidean geometry,}\\
    \bbh{n} = \left\{x\in \bbR{n+1}: \Lnorm{x}^2 = \frac{1}{K}, \, x_t > 0 \right \}, &  \text{For } K<0, \text{ hyperbolic geometry,}\\
    \end{cases}
\end{equation*}
where $\Lnorm{x}^2 = \norm{x_s}^2 - x _t ^2$ is the Lorentz inner product. Following the conventions of the hyperboloid, we write $x=(x_t, x_s^\top)^\top$, with $x_t \in \bbRscalar$ as the time component and $x_s \in \bbR{n}$ as the spatial component \citep{ratcliffe2006foundations}. When $K \neq 0$, the model can be written compactly as
\begin{equation*}
\calMK{n}= \left\{x \in \bbR{n+1}: \Kinner{x}{x} = 1 / K \right\},
\text{ with }
\Kinner{\cdot}{\cdot}=
\begin{cases}
\inner{\cdot}{\cdot}, & K>0,\\
\Linner{\cdot}{\cdot}, & K<0 \land x_t > 0.
\end{cases}
\end{equation*}
We abbreviate the $K$-radius model as the radius model. Its gyro-structure is developed in \cref{subsubsec:radius_gyro_space}.

\mypara{Beltrami--Klein Model.}
There are five models over the hyperbolic space \citep{cannon1997hyperbolic}. Apart from the above Poincaré ball and hyperboloid models, we further study the Beltrami--Klein model: 
\begin{equation*}
    \klein{n} =\left\{x \in \bbR{n} : \norm{x}^2 < -\frac{1}{K} \right\}, \text{ with } g ^{\mathbb{K}}_{x}(v,w) = \frac{\inner{v}{w}}{1+ K \norm{x} ^2} - \frac{K \inner{x}{v} \inner{x}{w}}{\left(1 + K \norm{x} ^2\right)^2},
\end{equation*}
where $K<0$ is the constant curvature and $g ^{\mathbb{K}}$ is its Riemannian metric. Although the Poincaré ball and Beltrami–Klein models share the same underlying set, their Riemannian metrics differ. This model admits an Einstein gyrovector space \citep[Chapter 6.18]{ungar2022analytic}.

\mypara{Full-Rank Correlation.}
The correlation matrix of a covariance matrix $\Sigma$ is defined as
$C = \coropt(\Sigma) = \bbD(\Sigma)^{-\nicefrac{1}{2}} \Sigma \bbD(\Sigma)^{-\nicefrac{1}{2}}$, where $\bbD(\cdot)$ returns a diagonal matrix with diagonal elements of $\Sigma$. The space of $n \times n$ full-rank correlation matrices, denoted as $\cor{n}$, forms a manifold \citep[Theorem 1]{david2019riemannian}, referred to as the correlation manifold. This manifold can be viewed as a compact representation of SPD matrices that preserves scale-invariant information. However, its Riemannian structure has been less studied than SPD matrices. Recently, \citet{thanwerdas2022theoretically,thanwerdas2024permutation} developed five convenient Riemannian metrics. Among them, \gls{PHCM} \citep{thanwerdas2022theoretically} identifies each correlation matrix with a product of hyperbolic spaces via the Cholesky decomposition. In \cref{subsec:cor_manifolds}, we construct the corresponding gyro-structure based on this identification.

\begin{table}[t]
  \centering
  \resizebox{0.99\linewidth}{!}{
    \begin{tabular}{cccccccc}
        \toprule
        Geometry & Symbol & $P \oplus Q$ or $x \oplus y$ & $E$ & $\ominus P$ or $\ominus x$ & Lie group & Gyrogroup & References \\
        \midrule
        AIM $\spd{n}$  & $\oplusGyrAI$ &  $P^{\frac{1}{2}} Q P^{\frac{1}{2}}$ & $I_n$ & $P^{-1}$   & \xmark &  \cmark  & \citep{nguyen2022gyrovector}        \\
        LEM $\spd{n}$ & $\oplusLieLE$  &   $\mexp (\mlog(P) + \mlog(Q))$    & $I_n$  & $P^{-1}$ & \cmark & \cmark & \makecell{\citep{arsigny2005fast} \\ \citep{nguyen2022gyrovector}}       \\
        LCM $\spd{n}$ & $\oplusLieLC$  &   $\clog^{-1}(\clog(P) + \clog(Q))$   & $I_n$ & $\clog(-\clog(P))$ &   \cmark & \cmark & \makecell{\citep{lin2019riemannian} \\ \citep{nguyen2022gyrovector} \\ \citep{chen2024adaptive}} \\
        \midrule
        $\graspp{p,n}$ & $\oplusGyrPP$     & $\mexp(\Omega) Q \mexp(-\Omega)$ &  $\idpp$ & $\mexp(-\Omega) \idpp \mexp(\Omega) $ &  \multirow{2}[2]{*}{\xmark} & \multirow{2}[2]{*}{Non-reductive} &\citep{nguyen2022gyro}  \\
        $\grasonb{p,n}$  & $\oplusGyrONB$   & $\mexp(\Omega) V$ & $\idonb$ & $\mexp(-\Omega) \idonb $ &  &  & \citep{nguyen2023building} \\
        \midrule
        $\stereo{n}$  &  $\oplusMK$ & $\frac{\left(1-2 K\langle x, y\rangle-K\|y\|^2\right) x+\left(1+K \|x\|^2 \right) y}{1 -2 K\langle x, y\rangle+K^2 \|x\|^2\|y\|^2}$ & $\Rzero$  & $-x$ & \xmark (\cmark for $K{=}0$) & \cmark & 
        \makecell{\citep{ungar2022analytic} \\ \citep{bachmann2020constant} \\ \cref{thm:stereographic_gyro}}
        \\
        \midrule
        $\klein{n}$ & $\Eoplus$ & $\frac{1}{1 - K\inner{x}{y}}\left(x+\frac{1}{\gamma_{x}} y -K \frac{\gamma_{x}}{1+\gamma_{x}} \inner{x}{y} x\right)$ & $\Rzero$ & $-x$ & \xmark & \cmark & \citep{ungar2022analytic}\\
        \bottomrule
    \end{tabular}
    }
    \caption{Gyrogroup on several geometries.}
    \label{tab:gyrogroup_opeartions}%
\end{table}

\cref{tab:gyrogroup_opeartions} summarizes the existing gyrogroups on the above geometries. The associated notations are defined as follows. 
\begin{itemize}
    \item 
    Let $\calM \in \{\spd{n}, \graspp{p,n}\}$ be a matrix manifold, and consider $P,Q \in \calM$. For the Grassmannian, $U=\pi^{-1}(P)$ and $V=\pi^{-1}(Q)$ denote the ONB representations. The matrix exponential, logarithm, and Cholesky decomposition are denoted by $\mexp$, $\mlog$, and $\scrL$, respectively. We further define $\clog = \dlog \circ \scrL$, where $\clog$ is the diagonal logarithm. We write $I_n$ for the $n \times n$ identity matrix, $\idonb=(I_{p},0)^\top \in \bbR{n \times p}$ for the ONB identity, and $\idpp=\pi(\idonb)$ for the PP identity. For the Grassmannian $\graspp{p,n}$, we define $\Omega=[\overline{P}, \idpp]$, where $\overline{P}=\rielog_{\idpp}(P)$ and $[\cdot,\cdot]$ denote the matrix commutator. 
    \item 
    Let $\ccs{n} \in \{\stereo{n}, \klein{n}\}$ be a constant curvature space, and consider $x, y \in \ccs{n}$. The gamma factor is defined as $\gamma_{x}=1 / \sqrt{1+ K \|x\|^2}$. 
\end{itemize}
\section{Pseudo-Reductive Gyrogroups}
\label{sec:pseudo-reductive_gyrogroups}

\begin{figure}[t]
\centering
\includegraphics[width=0.8\linewidth,trim={0cm 0cm 0cm 0cm}]{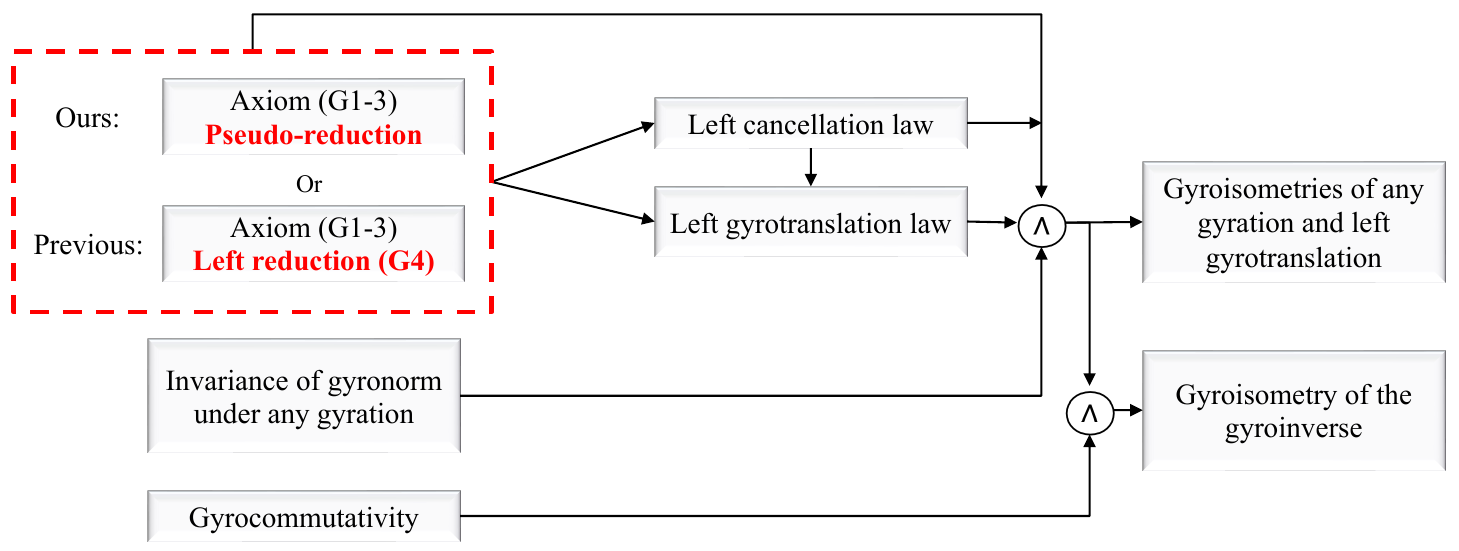}
\caption{The conceptual comparison of derivation logic of our work against previous work \citep{nguyen2023building}, where the left gyrotranslation law is presented in \cref{app:lem:left_gyrotranslation}.
The previous work proves the results on the SPD and Grassmannian manifolds in a case-by-case manner. In contrast, we relax the left reduction into pseudo-reduction and give a general analysis. Our framework also corrects the proof for the Grassmannian cases.
}
\label{fig:logics_isometry}
\end{figure}

Given a gyrogroup $(G, \oplus)$, the left gyrotranslation by $x \in G$ is defined as
\begin{equation*}
     L_x : G \rightarrow G, \quad L_{x}(y) = x \oplus y, \quad \forall y \in G.
\end{equation*}
If any gyrotranslation is a gyroisometry, we can use gyrotranslation to center manifold-valued samples for the normalization layer. \citet{nguyen2023building} show that any left gyrotranslation on the SPD and Grassmannian manifolds is a gyroisometry. However, the proof relies on the left cancellation law of gyrogroups, which does not hold for nonreductive gyrogroups, such as the Grassmannian. Therefore, the proof is questionable for the Grassmannian. We propose an intermediate structure, referred to as \emph{pseudo-reductive gyrogroups}, which can support the left cancellation law and, therefore, the gyroisometry of gyrotranslation. This will form the algebraic foundation for building the normalization layer. As illustrated in \cref{fig:logics_isometry}, our derivation extends the the case-by-case approach by \citet{nguyen2023building} into general pseudo-reductive gyrogroups.

\subsection{From Gyrogroups to Pseudo-Reductive Gyrogroups}

\begin{definition}[Pseudo-Reductive Gyrogroups] 
\label{def:pseudo_reductive_gyrogroup}
    A groupoid $(G, \oplus)$ is a pseudo-reductive gyrogroup if it satisfies the axioms (G1), (G2), (G3) and the following pseudo-reductive law:
    \begin{equation} \label{eq:pseudo_reduction}
        \gyr[a, x] = \id, \text{ for any left inverse } a \text{ of } x \text{ in } G,
    \end{equation}
    where $\id$ is the identity map.
\end{definition}

\cref{eq:pseudo_reduction} can be intuitively viewed as the intermediate between reduction and non-reduction. For gyrogroups, \cref{eq:pseudo_reduction} can be directly obtained from left gyroassociativity (G3) and reduction (G4) \citep[Theorem~2.10, item~3]{ungar2022analytic}. However, there is no theoretical guarantee that \cref{eq:pseudo_reduction} holds for non-reductive gyrogroups. Therefore, we name \cref{eq:pseudo_reduction} pseudo-reduction. Nevertheless, for the specific non-reductive Grassmannian, it is indeed pseudo-reductive.

\begin{proposition}\label{prop:grassmannian_pseudo_reductive_gyrogroups} 
    \linktoproof{prop:grassmannian_pseudo_reductive_gyrogroups}
    $\grasonb{p,n}$ and $\graspp{p,n}$ are pseudo-reductive gyrocommutative gyrogroups.
\end{proposition}

Our pseudo-reductive gyrogroup naturally generalizes the vanilla gyrogroup, as it shares most of the basic properties of gyrogroups \citep[Theorems. 2.10--2.11]{ungar2022analytic}.

\begin{theorem}[First Pseudo-Reductive Gyrogroups Properties]
    \label{thm:pseudo_reductive_gyrogroups_properties} 
    \linktoproof{thm:pseudo_reductive_gyrogroups_properties}
    Let $(G, \oplus)$ be a pseudo-reductive gyrogroup. For any elements $x, y, z, a \in G$, we have:
    \begin{enumerate}
        \item \label{enu:prgp_1}
        If $x \oplus y = x \oplus z$, then $y = z$ (General Left Cancellation law; see (\ref{enu:prgp_8}) below).
        \item \label{enu:prgp_2}
        $\gyr[e, x] = \id$ for any left identity $e$ in $G$.
        \item \label{enu:prgp_3}
        $\gyr[a, x] = \id$ for any left inverse $a$ of $x$ in $G$.
        \item \label{enu:prgp_4}
        There is a left identity that is a right identity.
        \item \label{enu:prgp_5}
        There is only one left identity.
        \item \label{enu:prgp_6}
        Every left inverse is a right inverse.
        \item \label{enu:prgp_7}
        There is only one left inverse, $\ominus x$, of $x$, and $\ominus(\ominus x) = x$.
        \item \label{enu:prgp_8}
        The left cancellation law: $\ominus x \oplus (x \oplus y) = y$.
        \item \label{enu:prgp_9}
        The gyrator identity: $\gyr[x, y] a = \ominus(x \oplus y) \oplus \{x \oplus (y \oplus a)\}$.
        \item \label{enu:prgp_10}
        $\gyr[x, y] e = e$.
        \item \label{enu:prgp_11}
        $\gyr[x, y](\ominus a) = \ominus \gyr[x, y] a$.
        \item \label{enu:prgp_12}
        $\gyr[x, e] = \id$.
        \item \label{enu:prgp_13}
        The gyrosum inversion law: $\ominus(x \oplus y) = \gyr[x, y](\ominus y \ominus x)$.
    \end{enumerate}
\end{theorem}
\begin{remark}
In non-reductive gyrogroups, \ref{enu:prgp_2} and \ref{enu:prgp_3} are undefined. Consequently, any property relying on them, such as~\ref{enu:prgp_4} and those from~\ref{enu:prgp_6} to~\ref{enu:prgp_10}, is not guaranteed to hold. The absence of these basic properties undermines the rationality of non-reductive gyrogroups. In contrast, our pseudo-reductive gyrogroups preserve most of the fundamental properties of gyrogroups.
\end{remark}

\subsection{Isometries over Pseudo-Reductive Gyrogroups}
The gyro-structure in the following is assumed to be defined as \cref{eq:gyro_addtion}--\cref{eq:gyro_distance}. We first clarify that the Riemannian distance agrees with the gyrodistance, and the Riemannian isometry agrees with the gyroisometry. These justify gyrodistance and gyroisometry for gyrospaces over manifolds.
\begin{lemma}[Distances]
\label{lem:gyro_geodesic_dist}
\linktoproof{lem:gyro_geodesic_dist}
    Given a pseudo-reductive gyrogroup $(\calM, \oplus)$, we have
    \begin{equation}
            \dist(x,y) = \norm{\rielog_x (y)}_x = \gyrnorm{ \ominus x \oplus y} = \gyrdist(x,y), \quad \forall x,y \in \calM,
    \end{equation}
    where $e \in \calM$ is the origin of the manifold, and $\dist$ denotes the geodesic distance.\footnote{On Cartan--Hadamard manifolds, the statement holds for all $x,y \in \calM$. More generally, the equality requires $x,y$ to lie within a geodesic ball of convexity radius to ensure the well-definedness of the minimizing geodesic and logarithm. In this paper, we implicitly assume these conditions are satisfied.}
\end{lemma}
\begin{lemma}[Isometries]
\label{lem:isometry_for_gyro}
\linktoproof{lem:isometry_for_gyro}
Let $(\calM, \oplus)$ and $(\widetilde{\calM}, \widetilde{\oplus})$ be two pseudo-reductive gyrogroups. Their gyro identity elements are $e \in \calM$ and $\widetilde{e} \in \widetilde{\calM}$, respectively. Suppose $\phi: \calM \to \widetilde{\calM}$ is a Riemannian isometry with $\widetilde{e}=\phi(e)$, we have the following.
\begin{enumerate}
    \item 
    The Riemannian isometry is a gyroisometry:
    \begin{equation}
        \gyrdist(x, y) = \widetilde{\gyrdist}(\phi(x), \phi(y)),
    \end{equation}
    where $\gyrdist$ and $\widetilde{\gyrdist}$ are the gyrodistances over $\calM$ and $\widetilde{\calM}$, respectively.
    \item 
    If the gyroinverse, gyration, or left gyrotranslation over $\calM$ is a gyroisometry, its counterpart over $\widetilde{\calM}$ is also a gyroisometry.
\end{enumerate}
\end{lemma}
\cref{lem:gyro_geodesic_dist} implies that, as long as the gyro-structure is defined by \cref{eq:gyro_addtion}--\cref{eq:gyro_distance}, the gyrodistance coincides with the geodesic distance. Unless otherwise specified, we shall not distinguish between the two and uniformly denote them by $\dist(\cdot,\cdot)$. Besides, the second result in \cref{lem:isometry_for_gyro} is particularly useful, as several geometries are isometric, such as the ONB and PP Grassmannian, as well as different models in hyperbolic geometry.

Now, we analyze gyroisometries over pseudo-reductive gyrogroups. The most related property in \cref{thm:pseudo_reductive_gyrogroups_properties} is the left cancellation law, one of the key prerequisites for gyro translation to be gyroisometry. Note that the left cancellation comes from left gyro associativity and \cref{eq:pseudo_reduction} \citep[Theorem~2.10, item~9]{ungar2022analytic}. Therefore, left cancellation does not generally hold for non-reductive gyrogroups but exists in pseudo-reductive gyrogroups. We first present an iff statement about gyroisometry, which will be useful in the following.

\begin{theorem}
    \label{thm:iff_gyroauto_gyroisometries} 
    \linktoproof{thm:iff_gyroauto_gyroisometries}
    Given a pseudo-reductive gyrogroup $(G, \oplus)$, $\gyr[x, y]$ preserves the gyronorm for any $x,y \in G$, iff $\gyr[x, y]$ is a gyroisometry for any $x,y \in G$.
\end{theorem}

The gyroisometry of any gyration is a prerequisite for other operators to be gyroisometries.
\begin{theorem}[Gyroisometries]
    \label{thm:gyroisometries} 
    \linktoproof{thm:gyroisometries}
    Given a pseudo-reductive gyrogroup $(G, \oplus)$ with any $\gyr[\cdot, \cdot]$ as a gyroisometry, we have the following.
    \begin{enumerate}
        \item 
        The left gyrotranslation is a gyroisometry.
        \item 
        If $(G, \oplus)$ is gyrocommutative, then any gyroinverse is a gyroisometry.
    \end{enumerate}
\end{theorem}
Now, we can discuss the gyroisometries w.r.t. \cref{tab:gyrogroup_opeartions}.
\begin{theorem} 
    \label{thm:gyroinvariance} 
    \linktoproof{thm:gyroinvariance}
    For the pseudo-reductive gyrogroups corresponding to the SPD manifold (under AIM, LEM, and LCM), the ONB and PP Grassmannian, and the stereographic model with $K \leq 0$ (the Poincaré ball for $K<0$ and Euclidean space for $K=0$), the gyrodistance coincides with the geodesic distance. Moreover, the gyroinverse, any gyration, and any left gyrotranslation are gyroisometries. 
\end{theorem}
\begin{proof}[Credit and Sketch of the Proof] 
As \cref{lem:gyro_geodesic_dist} already shows that the gyrodistance coincides with the geodesic distance, it remains to establish the isometries. For the Grassmannian and SPD manifolds, these were proved by \citet[Theorems~2.12--2.14 and~2.16--2.18]{nguyen2023building}, although their arguments implicitly treated the non-reductive Grassmannian as a gyrogroup by using left cancellation. Our \cref{prop:grassmannian_pseudo_reductive_gyrogroups,thm:pseudo_reductive_gyrogroups_properties} confirms that the Grassmannian is pseudo-reductive and does satisfy left cancellation, thereby validating their results. However, we can directly establish these properties from \cref{thm:iff_gyroauto_gyroisometries,thm:gyroisometries}. The complete proof is given in \cref{app:subsec:proof_gyroinvariance}.
\end{proof}
\begin{remark}
The remaining manifolds in \cref{subsec:preliminaries-examples}, including the stereographic model with $K>0$, radius model, and Beltrami--Klein model, also satisfy these properties. Their verifications will be presented in \cref{sec:instantiations}.
\end{remark}
\section{GyroBN on Pseudo-Reductive Gyrogroups}
\label{sec:gyrobn-general}
Building on \cref{thm:gyroinvariance}, which establishes that several geometries admit isometric gyrotranslations, we develop RBN in a principled way for general pseudo-reductive gyrogroups, referred to as \glsfull{GyroBN}. Throughout, we assume $(\calM, \oplus)$ is a pseudo-reductive gyrogroup with the gyro-structure defined by \cref{eq:gyro_addtion}--\cref{eq:gyro_distance}.\footnote{In GyroBN, $\odot$ is not required to satisfy the axioms of a gyrovector space (\cref{def:gyrovector_spaces}).} As \cref{lem:gyro_geodesic_dist} establishes the equivalence between gyrodistance and geodesic distance, we use the terms ``gyromean'' and ``gyrovariance'' interchangeably with their Riemannian counterparts.

\subsection{Euclidean Batch Normalization Revisited}
As the core operations of different Euclidean normalization variants \citep{ioffe2015batch, ba2016layer, ulyanov2016instance, wu2018group} are similar, this paper focuses on BN. Given a batch of activations $\{x_{i \ldots N} \}$, the BN update can be expressed as
\begin{equation} \label{eq:ebn}
    \forall i \leq N, \quad x_i \gets \gamma \frac{x_i-\mu}{\sqrt{v^2+\epsilon}} + \beta
\end{equation}
where $\mu$ and $v^2$ denote the mean and variance of the sample, and $\gamma$ and $\beta$ are the learnable scaling and bias parameters.

\subsection{GyroBN}
To generalize the Euclidean BN into gyrogroups, we first define sample mean, sample variance, centering, biasing, and scaling over gyrogroups. Then, we introduce our GyroBN framework with a theoretical analysis of the ability to normalize sample statistics.

We define the gyromean as the Fréchet mean \citep{frechet1948elements} under gyrodistance:
\begin{equation} \label{eq:gyromean}
    \mu = \fm(\{ x_i \in \calM \}_{i=1}^N) = \underset{y \in \calM}{\argmin} \frac{1}{N} \sum\nolimits_{i=1}^N  \dist^2\!\left(x_i, y \right).
\end{equation}
The gyrovariance is the corresponding Fréchet variance. By \cref{lem:gyro_geodesic_dist}, the gyromean and gyrovariance coincide with the Riemannian mean and variance, \ie the Fréchet mean and variance under geodesic distance. For completeness, \cref{app:sec:exist_unique_wfm} reviews the existence and uniqueness of the Fréchet mean.

Easy computation shows that centering, biasing, and scaling in the Euclidean BN (\eqnref{eq:ebn}) correspond to gyrosubtraction (\eqnrefs{eq:gyro_addtion,eq:gyro_inverse}), gyroaddition (\eqnref{eq:gyro_addtion}), and scalar gyromultiplication (\eqnref{eq:gyro_scalar_product}) in $\bbR{n}$. Motivated by this, we define normalization layer over gyrogroups via gyro operations. Given a batch of activations $\{x_i\}_{i=1}^N \subset \calM$, the core operations of GyroBN are
\begin{equation}
    \label{eq:gyrobn_centering}
    \forall i \leq N, \quad \tilde{x}_i \gets  \overbrace{ \beta \oplus}^{\text{Biasing}} \left( \overbrace{\frac{s}{\sqrt{v^2+\epsilon}} \odot }^{\text{Scaling}} \left( \overbrace{\ominus \mu  \oplus x_i}^{\text{Centering}} \right) \right),
\end{equation}
where $\mu \in \calM$ and $v^2$ are gyromean and gyrovariance, $\beta \in \calM$ is the bias parameter, $s \in \bbRscalar$ is the scaling parameter, and $\epsilon$ is a small value for numerical stability. The following theorem shows that \cref{eq:gyrobn_centering} can normalize manifold-valued data.

\begin{theorem}[Homogeneity]
    \label{thm:gyrobn} \linktoproof{thm:gyrobn}
    Let $(\calM, \oplus)$ be a pseudo-reductive gyrogroup in which every gyration $\gyr[\cdot,\cdot]$ is a gyroisometry. For $N$ samples $\{x_i\}_{i=1}^N \subset \calM$, we have
    \begin{align}
        \label{eq:hom_fm_gyrogroup}
        & \text{Homogeneity of gyromean: }
        \fm( \{\beta \oplus x_i \}_{i=1}^N ) = \beta \oplus \fm( \{ x_i \}_{i=1}^N ), \quad \forall \beta \in \calM,\\
        \label{eq:variance_gyrogroup}
        & \text{Homogeneity of dispersion from $e$: } 
        \frac{1}{N} \sum\nolimits_{i=1}^N \dist^2(t \odot x_i, e) = \frac{t^2}{N} \sum\nolimits_{i=1}^N \dist^2(x_i, e),
    \end{align}
\end{theorem}

The most important property of the Euclidean BN~\citep{ioffe2015batch} lies in its ability to normalize the sample mean and variance. Theorem~\ref{thm:gyrobn} shows that our formulation in \cref{eq:gyrobn_centering} enjoys the same property: \cref{eq:hom_fm_gyrogroup} guarantees that centering and biasing shift the gyromean, while \cref{eq:variance_gyrogroup} ensures that scaling controls the sample variance. As a result, GyroBN provides a theoretical guarantee of normalization on any pseudo-reductive gyrogroup with isometric gyrations. Moreover, since the gyromean and gyrovariance coincide with their Riemannian counterparts, GyroBN also normalizes Riemannian statistics. 

\begin{algorithm}[t] 
\SetKwInOut{Input}{Require}
\SetKwInOut{Output}{Return}
\SetKwInOut{Parameters}{Parameters}
\caption{\glsfull{GyroBN}}
\label{alg:gyrobn}
\Input{
batch of activations $\{x_{1 \ldots N} \in \calM\}$, small positive constant $\epsilon$, and momentum $\eta \in [0,1]$, running mean $\mu_r$, running variance $v^2_r$, bias parameter $\beta \in \calM$, scaling parameter $s \in \bbRscalar $.
}
\Output{normalized batch $\{\tilde{x}_{1 \ldots N} \in \calM \}$}
\BlankLine
\If{training}{
    Compute batch mean $\mu_b$ and variance $v_b^2$ of $\{x_{1 \ldots N}\}$;\\
    Update running statistics 
    $\mu_r = \barcenter_\gamma (\mu_b,\mu_r)$, and $v^2_r = \gamma v^2_b + (1-\gamma)v^2_r$;
}

$(\mu,v^2) = (\mu_b, v^2_b )$ \textbf{if} \textit{training} \textbf{else} $(\mu_r, v_r^2)$

$\forall i \leq N, \ \tilde{x}_i = \beta \oplus \left( \frac{s}{\sqrt{v^2+\epsilon}} \odot \left( \ominus \mu \oplus x_i \right) \right)$
\end{algorithm}

To finalize our GyroBN, we define the running mean updates over gyrogroups as the binary barycenter based on gyrodistance: 
\begin{equation*}
        \barcenter_\eta (x_1, x_2) = {\argmin}_{y \in \calM}  \left( \eta \dist^2\!\left(x_1, y \right) + (1-\eta) \dist^2\!\left(x_2, y \right)  \right), \quad \eta \in [0,1],
\end{equation*}
which can be calculated by the geodesic. With these ingredients, the general framework for GyroBN is presented in \cref{alg:gyrobn}. In particular, it recovers the classic Euclidean BN~\citep{ioffe2015batch} when $\calM=\bbR{n}$.

\begin{remark}
We make the following two remarks w.r.t. left translation.
\begin{itemize}[leftmargin=*, itemsep=0pt, topsep=2pt, parsep=2pt]
    \item 
    \mypara{Other Candidates.}
    There are three alternatives of left gyrotranslation. However, they are not necessarily
    gyroisometries and therefore are not suitable for building the normalization layer.
    Specifically, analogous to the left gyrotranslation, the right gyrotranslation is
    \begin{equation*}
         R_x : G \rightarrow G, \quad R_{x}(y) = y \oplus x, \quad \forall y \in G.
    \end{equation*}
    Along with the gyroaddition, there is the gyrogroup coaddition \citep[Definition 2.9]{ungar2022analytic}:
    \begin{equation*}
        x \boxplus y = x \oplus \gyr[x, \ominus y] y, \quad \forall x, y \in G.
    \end{equation*}
    Coaddition is symmetric to gyroaddition in many ways; for instance, when gyroaddition is
    gyrocommutative, coaddition is commutative \citep[Theorem~3.3]{ungar2022analytic}.
    However, the right gyrotranslation, as well as the left and right translations by coaddition,
    are not guaranteed to be gyroisometries. Numerical experiments confirm that these three
    translations on the hyperbolic Poincaré ball fail to preserve gyrodistance (see
    \texttt{gyrocoadd.py}). A theoretical reason is that they all lack a counterpart of the left
    gyrotranslation law, which is important for the translation to be a gyroisometry
    (see~\cref{fig:logics_isometry}).

    \item 
    \mypara{Special Cases.} 
    For Lie groups with a right-invariant Riemannian metric, right translations are isometries, and GyroBN can then be formulated using them, with \cref{thm:gyrobn} extending directly. In general, however, only left gyrotranslations are guaranteed to be isometries.
\end{itemize}
\end{remark}
\section{Instantiations}
\label{sec:instantiations}
As indicated by \cref{thm:gyroinvariance,thm:gyrobn}, our GyroBN can be applied to different geometries with guaranteed normalization of the sample statistics. Once the required operators are specified, \cref{alg:gyrobn} can be used in a plug-and-play manner. We first show that existing RBN methods with control over sample statistics, such as LieBN on Lie groups and AIM-based SPDBNs, are special cases of our framework. We then instantiate GyroBN on seven representative geometries: the Grassmannian, five constant curvature models (Poincaré ball, projected hypersphere, hyperboloid, sphere, and Beltrami–Klein), and the full-rank correlation manifold. To support these instantiations, we simplify the Grassmannian operators for efficient computation, refine the gyro-structures on the Poincaré ball and projected hypersphere, establish new gyro-structures for the hyperboloid and sphere, characterize the Beltrami–Klein geometry, and formulate a row-wise realization for the correlation manifold.

\subsection{LieBN as a Special Case}
\label{subsec:liebn-as-gyrobn}
\citet[Algorithms~3--4]{chakraborty2020manifoldnorm} introduced the Riemannian normalization on matrix Lie groups under a specific distance. \citet{chen2024liebn} extended this framework to general Lie groups, yielding LieBN, which provides theoretical normalization over the Riemannian mean and variance. This subsection shows that LieBN is a special case of our GyroBN.

LieBN is formulated with a left-invariant metric on a Lie group. Centering and biasing are performed through left group translations, while scaling is defined in the tangent space at the identity element \citep[Equations~13--15]{chen2024liebn}, which coincides with our gyromultiplication in \cref{eq:gyro_scalar_product}. Since every Lie group is automatically a gyrogroup, gyrotranslation reduces exactly to the group translation. Consequently, the centering, biasing, and scaling in LieBN are identical to those in GyroBN. Moreover, as shown in \cref{lem:gyro_geodesic_dist}, the mean, variance, and running mean update defined via the geodesic distance in LieBN are equivalent to their counterparts based on gyrodistance in GyroBN. Therefore, LieBN is a special case of our GyroBN. \citet{chen2024liebn} implemented LieBN on three left-invariant metrics of the SPD manifold (AIM\footnote{AIM is left-invariant with respect to the Lie group operation $P \oplus^{\text{Lie}}_{\text{AI}} Q = LQL^\top$, where $L$ is the Cholesky factor of $P=LL^\top$ \citep{thanwerdas2022theoretically}. This group structure differs from the one in \cref{tab:gyrogroup_opeartions}.}, LEM, and LCM) and on a bi-invariant metric over the special orthogonal group. These are therefore incorporated by our GyroBN.

\subsection{AIM-Based SPDBNs as Special Cases}
\label{subsec:spdbn-as-gyrobn}
Several RBNs on the SPD manifold were developed based on AIM \citep{brooks2019riemannian,kobler2022controlling,kobler2022spd}.
The core operations of these approaches can be expressed as
\begin{equation} \label{eq:spdbn_aim}
    \text{Normalization: } 
    \forall i \leq N, \quad \widetilde{P}_i \gets B^{\frac{1}{2}} \left( M^{-\frac{1}{2}} P_i M^{-\frac{1}{2}} \right)^{\frac{s}{\sqrt{v^2+\epsilon}}} B^{\frac{1}{2}},
\end{equation}
where $M$ and $v^2$ are the Riemannian mean and variance. The running mean is updated by the binary barycenter under the geodesic distance.

As gyrodistance is identical to the geodesic distance, the gyromean, gyrovariance, and running mean updates are identical to the Riemannian ones. Recalling \cref{tab:gyrogroup_opeartions}, \cref{eq:spdbn_aim} is exactly the specific implementation of \cref{eq:gyrobn_centering} under AIM-based gyrogroup on the SPD manifold.\footnote{\citet[Lemma 4]{nguyen2022gyrovector} show that the AIM gyromultiplication is $t \odot^{\mathrm{AI}} S = S^t, \forall t \in \bbRscalar, \forall S \in \spd{n}$.} Therefore, the SPDBNs developed by \citet{brooks2019riemannian,kobler2022controlling,kobler2022spd} are also special cases of our GyroBN.
\begin{remark}
\citet{brooks2019riemannian} only consider centering and biasing.
\citet{kobler2022controlling} use running mean for centering during the training. \citet{kobler2022spd} use different momentum to update running statistics for training and testing and multi-channel mechanisms for domain adaptation. Nevertheless, all of them are based on \cref{eq:spdbn_aim}.
Therefore, tricks such as multi-channel and separate momentum can also be applied to our GyroBN. This is what we mean by claiming that our GyroBN incorporates their approaches.
\end{remark}

\subsection{Grassmannian Manifold}
\label{subsec:grassmannian_gyrobn}

\begin{table}[tbp]
    \centering
    \begin{subtable}[t]{0.55\linewidth}
        \centering
        \resizebox{\linewidth}{!}{
        \begin{tabular}{cc}
            \toprule
            Operator & Expression \\
            \midrule
            $\dist(U,V)$ & 
            \makecell{$\|\arccos (\Sigma)\|$ \\ $U^\top V \stackrel{\mathrm{SVD}}{:=} O\Sigma R^\top$} \\[6pt]
            $\rielog_U V$ & 
            \makecell{$O \arctan(\Sigma) R^\top$ \\ $(I_n - UU^\top)V(U^\top V)^{-1} \stackrel{\mathrm{SVD}}{:=} O\Sigma R^\top$} \\[6pt]
            $\rieexp_U \Delta$ & 
            \makecell{$UR \cos(\Sigma) R^\top + O \sin(\Sigma) R^\top$ \\ $\Delta \stackrel{\mathrm{SVD}}{:=} O\Sigma R^\top$} \\
            \bottomrule
        \end{tabular}}
        \caption{Riemannian operators}
    \end{subtable}%
    \hfill
    \begin{subtable}[t]{0.42\linewidth}
        \centering
        \begin{tabular}{cc}
            \toprule
            Operator & Expression \\
            \midrule
            $U \oplusGyrONB V$ & $\mexp(\Omega) V$ \\[6pt]
            Identity & $\idonb$ \\[6pt]
            $t \odotGyrONB U$ & $\mexp(t\Omega)\idonb$ \\[6pt]
            $\ominusGyrONB U$ & $\mexp(-\Omega)\idonb$ \\
            \bottomrule
        \end{tabular}
        \caption{Gyro operators}
    \end{subtable}
    \caption{Operators on the ONB Grassmannian.}
    \label{tab:riem-gyro-operators-grass-onb}
\end{table}

The Grassmannian $\grasonb{p,n}$, representing the space of $p$-dimensional subspaces in $\mathbb{R}^n$, has been widely applied in machine learning, ranging from action recognition \citep{huang2018building,nguyen2023building} to question answering \citep{nguyen2023building}, shape generation \citep{yataka2023grassmann}, image classification \citep{wang2023get}, and signal analysis \citep{wang2024grassatt}. We focus on the ONB perspective. Let $U,V \in \grasonb{p,n}$, $t \in \bbRscalar$, and $\Delta \in T _U \grasonb{p,n}$. Following the notation in \cref{tab:gyrogroup_opeartions}, \cref{tab:riem-gyro-operators-grass-onb} summarizes the key operators.

\mypara{Instantiation.}
Building on \cref{tab:riem-gyro-operators-grass-onb}, we now implement the ONB Grassmannian GyroBN. Given a batch of activations $\{U_{1 \cdots N}\}$, the three core steps of GyroBN are
\begin{align}
    \text{Centering to the identity $\idonb$: }&
    U^1_i = \mexp \left( -[\overline{M M^\top}, \idpp]\right) U_i,\\
    \text{Scaling the dispersion from $\idonb$: }&
     U^2_i = \mexp \left(  \frac{s}{\sqrt{v^2+\epsilon}} [\overline{U^1_i(U^1_i)^\top}, \idpp] \right) \idonb,\\
     \label{eq:gyrobn_grass_biasing}
     \text{Biasing towards $B \in \calM$: }&
     U^3_i = \mexp \left([\overline{B}, \idpp] \right)U^2_i .
\end{align}
Here $\overline{(\cdot)} = \widetilde{\rielog}_{\idpp}(\cdot)$ is the Riemannian logarithm under the PP Grassmannian, $M$ (resp. $v^2$) is the Riemannian batch mean (resp. variance), and $\idpp=\idonb \idonb^\top$ is the PP identity. The mean $M$ can be obtained by the Karcher flow~\citep{karcher1977riemannian}, with the Riemannian logarithm computed by the efficient and stable Algorithm~5.3 of \citet{bendokat2024grassmann}.

\mypara{Efficient Computation.}
The commutators $[\overline{M M^\top}, \idpp]$ and $[\overline{U^1_i(U^1_i)^\top},\idpp]$ can be efficiently computed by the following result.

\begin{proposition} 
    \label{prop:fast_bracket_grassmannian}
    \linktoproof{prop:fast_bracket_grassmannian}
    Given $U=(U_1^\top,U_2^\top)^\top \in \grasonb{p,n}$ with $U_1 \in \bbR{p \times p}$ and $U_2 \in \bbR{(n-p) \times p}$, then
    \begin{equation*}
        [\overline{U U^\top},\idpp] = 
        \begin{pmatrix}
             \Rzero & -\widetilde{U}_2^T \\
             \widetilde{U}_2 & \Rzero
        \end{pmatrix},
    \end{equation*}
    where $\widetilde{U}_2 = U_2 Q \frac{\arcsin(\hat{S})}{\hat{S}}R^\top$ and $U_1^\top \stackrel{\mathrm{SVD}}{:=} QSR^\top$. 
    Here $S$ is in ascending order, $Q$ and $R$ are flipped column-wise, and $\hat{S} = \sqrt{1-S^2}$.
\end{proposition}

\begin{remark}
Two technical issues are worth noting.  
(1) \mypara{Cut locus:} the logarithm $\rielog_U(V)$ exists only when $U$ and $V$ are not in each other's cut locus~\citep{bendokat2024grassmann}. Similarly, gyroaddition and gyromultiplication are not globally defined due to the cut locus~\citep[Section 3.2]{nguyen2022gyro}. However, Algorithm~5.3 of \citet{bendokat2024grassmann} provides a numerical remedy.  
(2) \mypara{PP Grassmannian:} although our derivation is based on the ONB Grassmannian, GyroBN under the PP Grassmannian can be obtained by mapping data via $\pi^{-1}: \graspp{p,n} \to \grasonb{p,n}$, normalizing, and mapping back via $\pi$. Details are deferred to \cref{app:sec:gyro_grass_pp}.
\end{remark}

\subsection{Stereographic Model}
\label{subsec:stereo_model}

\begin{table}[tbp]
    \centering
    \begin{subtable}[t]{0.52\linewidth}
        \centering
        \resizebox{\linewidth}{!}{
        \begin{tabular}{cc}
            \toprule
            Operator & Expression \\
            \midrule
            $\dist(x,y)$ & 
            $\displaystyle \frac{2}{\sqrt{|K|}} \tanh_K^{-1}\!\left(\sqrt{|K|}\,\|-x \stoplus y\|\right)$ \\
            $\rielog_x y$ & 
            $\displaystyle \frac{2}{\sqrt{|K|}\,\lambda_x^K}\,
            \tanh_K^{-1}\!\left(\sqrt{|K|}\,\|-x \stoplus y\|\right)\,
            \frac{-x \stoplus y}{\|-x \stoplus y\|}$ \\
            $\rieexp_x v$ & 
            $\displaystyle x \stoplus \left(
            \tanh_K \left(\sqrt{|K|}\,\frac{\lambda_x^K \norm{v}}{2}\right)
            \frac{v}{\sqrt{|K|}\,\norm{v}}
            \right)$ \\
            $\pt{x}{y}(v)$ & $\displaystyle \frac{\lambda_{x}^K}{\lambda_{y}^K} \gyr[y,-x] v$ \\
            \bottomrule
        \end{tabular}}
        \caption{Riemannian operators}
    \end{subtable}%
    \hfill
    \begin{subtable}[t]{0.45\linewidth}
        \centering
        \resizebox{\linewidth}{!}{
        \begin{tabular}{cc}
            \toprule
            Operator & Expression \\
            \midrule
            $x \stoplus y$ & 
            $\displaystyle \frac{(1-2K\langle x,y\rangle - K\|y\|^2)x + (1+K\|x\|^2)y}{1 - 2K\langle x,y\rangle + K^2\|x\|^2\|y\|^2}$ \\
            Identity & $\Rzero$ \\
            $t \stodot x$ & 
            $\displaystyle \frac{\tanh_K \!\left(t \tanh_K^{-1}(\sqrt{|K|}\|x\|)\right)}{\sqrt{|K|}} \frac{x}{\|x\|}$ \\
            $\stominus x$ & $-x$ \\
            \bottomrule
        \end{tabular}
        }
        \caption{Gyro operators}
    \end{subtable}
    \caption{Operators on the stereographic model.}
    \label{tab:riem-gryo-operators-stereo}
\end{table}

The stereographic model has shown success in different applications, including computer vision \citep{van2023poincare}, natural language processing \citep{ganea2018hyperbolic,shimizu2020hyperbolic}, graph learning \citep{bachmann2020constant,grover2025curvgad,grover2025spectro}, and astronomy \citep{chen2025galaxy}. We begin by analyzing its gyro-structure and then instantiate GyroBN.

\subsubsection{Stereographic Gyrovector Space}
As introduced in \cref{subsec:preliminaries-examples}, the stereographic model $\stereo{n}$ unifies constant curvature geometries: the hyperbolic Poincaré ball $\pball{n}$ for $K<0$, Euclidean space $\bbR{n}$ for $K=0$, and the spherical projected hypersphere $\projhs{n}$ for $K>0$. For $x,y,z \in \stereo{n}$, $t \in \bbRscalar$, and $v \in T_x\stereo{n}$, its Riemannian and gyro operators are summarized in \cref{tab:riem-gryo-operators-stereo}, where $\tank=\tanh$ if $K<0$ and $\tank=\tan$ if $K>0$. The gyration is given by \citet[Appendix~C.2.6]{bachmann2020constant}:
\begin{equation} \label{eq:stereographic_gyration}
    \gyr[x,y]z \;=\; z + 2 \frac{Ax+By}{D},
\end{equation}
with
\begin{equation*}
\begin{aligned}
   A &= -K^2 \inner{x}{z}\|y\|^2 -K\inner{y}{z} + 2K^2\inner{x}{y}\inner{y}{z}, \\
   B &= -K^2 \inner{y}{z}\|x\|^2 + K\inner{x}{z}, \\
   D &= 1 - 2K\inner{x}{y} + K^2\|x\|^2\|y\|^2 > (1-K\inner{x}{y})^2 > 0 .
\end{aligned}
\end{equation*}

For $K<0$, the Poincaré ball $(\pball{n},\stoplus,\stodot)$ forms a Möbius gyrovector space \citep[Theorem~6.85]{ungar2022analytic}. For $K>0$, however, the situation is subtler \citep{bachmann2020constant}:
\begin{itemize}[itemsep=0pt,topsep=2pt,parsep=2pt]
    \item gyroaddition is definite except $x = \frac{y}{K\norm{y}^2} = 0$;  
    \item gyromultiplication is definite except $r\tan^{-1}(\sqrt{K}\norm{x}) = \nicefrac{\pi}{2} + k\pi$ for some $k \in \bbZ$.
\end{itemize}
Even if assumed well-defined, it remains unclear whether $(\stereo{n},\stoplus,\stodot)$ with $K>0$ satisfies the axioms of a gyrovector space, in contrast to the Poincaré ball. Closing this gap is part of our contribution: under the assumption of well-definedness, we show that the stereographic gyro operations coincide with those in \cref{eq:gyro_addtion,eq:gyro_scalar_product} and prove that the stereographic model satisfies all axioms of a gyrovector space.

\begin{proposition} \label{prop:stereographic_gyro_from_riemannian}
    \linktoproof{prop:stereographic_gyro_from_riemannian}
    The stereographic gyroaddition and gyromultiplication coincide with the Riemannian definitions:
    \begin{align}
        x \stoplus y &= \rieexp_{x}\!\left(\pt{\Rzero}{x}(\rielog_{\Rzero}(y))\right), \quad \forall x,y \in \stereo{n}, \\
        t \stodot x &= \rieexp_{\Rzero}(t \,\rielog_{\Rzero}(x)), \quad \forall t \in \bbRscalar,\; x \in \stereo{n}.
    \end{align}
\end{proposition}

\begin{theorem} \label{thm:stereographic_gyro}
    \linktoproof{thm:stereographic_gyro}
    For any $K\in\bbRscalar$, the stereographic model $(\stereo{n},\stoplus)$ satisfies all axioms of a gyrocommutative gyrogroup. When further endowed with gyromultiplication $\stodot$, it satisfies all axioms of a gyrovector space.
\end{theorem}

\subsubsection{Stereographic GyroBN}

Next, we extend \cref{thm:gyroinvariance} to the stereographic model with arbitrary curvature.

\begin{theorem} 
    \label{thm:gyroinvariance_stereo} 
    \linktoproof{thm:gyroinvariance_stereo}
    For the stereographic model, the gyrodistance coincides with the geodesic distance. Moreover, the gyroinverse, any gyration, and any left gyrotranslation are gyroisometries. 
\end{theorem}

Combining \cref{thm:gyrobn} and \cref{thm:gyroinvariance_stereo}, GyroBN in the stereographic model is theoretically guaranteed to normalize sample statistics. Practically, implementation only requires substituting the operators in \cref{tab:riem-gryo-operators-stereo} into \cref{alg:gyrobn}. For efficient computation, the Poincaré Fréchet mean can be obtained using the algorithm of \citet[Algorithm~1]{lou2020differentiating}, while the mean on the sphere is computed via the Karcher flow \citep{karcher1977riemannian}.

\subsection{Radius Model}
\label{subsec:radius-model}
As introduced in \cref{subsec:preliminaries-examples}, the radius model $\calMK{n}$ is another representation of constant-curvature spaces, unifying the hyperboloid $\bbh{n}$ for $K<0$, the sphere $\sphere{n}$ for $K>0$, and the Euclidean space $\bbR{n}$ for $K=0$. Although this model has been effective in various applications \citep{chami2019hyperbolic,chen2021fully,bdeir2024fully,pal2024compositional,he2025lorentzian,khan2025hyperbolic}, its gyro-structure has not been formalized. We first analyze the gyro-structure on $\calMK{n}$ and then instantiate GyroBN.

\subsubsection{Radius Gyrovector Space}
\label{subsubsec:radius_gyro_space}

\begin{table}[tbp]
    \centering
    \begin{subtable}[t]{0.48\linewidth}
        \centering
        \renewcommand{\arraystretch}{1.3}
        \resizebox{\linewidth}{!}{
        \begin{tabular}{cc}
            \toprule
            Operator & Expression \\
            \midrule
            $\dist(x,y)$ &
            \makecell{$\displaystyle \frac{1}{\sqrt{|K|}} \cosk^{-1}\!\left(\beta\right)$ \\ $\beta = K \Kinner{x}{y}$} \\
            $\rielog_x y$ &
            \makecell{$\displaystyle \frac{\cosk^{-1}(\beta)}{\sqrt{\sign(K)\,(1-\beta^2)}}\,(y-\beta x)$ \\ $\beta = K \Kinner{x}{y}$} \\
            $\rieexp_x v$ &
            \makecell{$\displaystyle \cosk(\alpha)\,x+ \frac{\sink(\alpha)}{\alpha}\, v$ \\ $\alpha=\sqrt{|K|}\Knorm{v}$} \\[6pt]
            $\pt{x}{y}(v)$ & $\displaystyle v- \frac{K \inner{y}{v}_K}{1+K \inner{x}{y}_K}(x+y)$ \\
            \bottomrule
        \end{tabular}}
        \caption{General form}
    \end{subtable}
    \hfill
    \begin{subtable}[t]{0.5\linewidth}
        \centering
        \renewcommand{\arraystretch}{1.3}
        \resizebox{\linewidth}{!}{
        \begin{tabular}{cc}
            \toprule
            Operator & Expression \\
            \midrule
            $\dist(\MKzero,x)$ &
            $\displaystyle \tfrac{1}{\sqrt{|K|}}\cosk^{-1}(\sqrt{|K|}\,x_t)$ \\
            $\rielog_{\MKzero}(x)$ &
            $\displaystyle
              \begin{bmatrix}
                0 \\
                \tfrac{\cosk^{-1}(\sqrt{|K|}\,x_t)}{\sqrt{|K|}\,\|x_s\|}\,x_s
              \end{bmatrix}$ \\
            $\rieexp_{\MKzero}(v)$ &
            $\displaystyle \tfrac{1}{\sqrt{|K|}}
              \begin{bmatrix}
                \cosk(\sqrt{|K|}\|v_s\|) \\
                \tfrac{\sink(\sqrt{|K|}\|v_s\|)}{\|v_s\|}\,v_s
              \end{bmatrix}$ \\
            $\pt{\MKzero}{x}(v)$ &
            $\displaystyle v-\tfrac{K\,\inner{x_s}{v_s}}{1+\sqrt{|K|}\,x_t}
              \begin{bmatrix}
                x_t+1/\sqrt{|K|}\\[2pt]
                x_s
              \end{bmatrix}$ \\
            \bottomrule
        \end{tabular}}
        \caption{At the origin $\MKzero$}
    \end{subtable}
    \caption{Riemannian operators on the radius model.}
    \label{tab:riem-operator-radius}
\end{table}

The radius model $\calMK{n}$ is isometric to the stereographic model $\stereo{n}$ via stereographic projection fixing the south pole \citep{skopek2019mixed}:
\begin{align}
    \label{eq:iso_calMK_to_stereo}
    \isoMKST{n}
    &: \calMK{n} \ni 
    \begin{bmatrix}
    \xi \in \bbRscalar \\
    x \in \bbR{n}
    \end{bmatrix} \longmapsto \frac{x}{1+\sqrt{|K|} \xi} \in \stereo{n}, \\
    \label{eq:iso_stereo_to_calMK}
    \isoSTMK{n}
    &: \stereo{n} \ni y \longmapsto 
    \begin{bmatrix}
    \frac{1}{\sqrt{|K|}} \frac{1-K\| y \|^2}{1+K\| y \|^2} \\
    \frac{2 y}{1+ K\| y \|^2}
    \end{bmatrix} \in \calMK{n}.
\end{align}
The origin in $\calMK{n}$ is defined as $\MKzero=[\sqrt{1/|K|}, 0, \ldots, 0]^\top$, corresponding to $\Rzero \in \stereo{n}$. For $K>0$, we have $\calMK{n}=\sphere{n}$ and $\stereo{n}=\projhs{n}$. Since \cref{eq:iso_calMK_to_stereo} is undefined at the south pole $-\MKzero$ when $K>0$, we use the one-point compactification $\projhs{n}\cup\{\infty\}$ with the identification $\pi_{\calMK{n}\to\stereo{n}}(-\MKzero)=\infty$ \citep[Remark~A.9]{skopek2019mixed}. For simplicity, we use $\projhs{n}$ and $\projhs{n}\cup\{\infty\}$ interchangeably.

\cref{tab:riem-operator-radius} lists the Riemannian operators \citep[Table~1]{skopek2019mixed}, together with the simplified expressions at $\MKzero$. We adopt the following curvature-aware functions:
\begin{equation*}
    \sin _K
    =\left\{\begin{array}{ll}
    \sin & \text { if } K>0 \\
    \sinh & \text { if } K<0
    \end{array} 
    \quad 
    \cos _K
    =\left\{\begin{array}{ll}
    \cos & \text { if } K>0 \\
    \cosh & \text { if } K<0
    \end{array} 
    \quad 
    \Kinner{\cdot}{\cdot}=
    \begin{cases}
    \inner{\cdot}{\cdot} & \text { if } K>0\\
    \Linner{\cdot}{\cdot} & \text { if } K<0 
    \end{cases} \right.\right.
\end{equation*}
Besides, $\Knorm{\cdot}$ is the norm induced by $\Kinner{\cdot}{\cdot}$, $(\cdot)_s$ denotes the space vector, and $(\cdot)_t$ denotes the time scalar.

By \cref{tab:riem-operator-radius}, we define the gyroaddition and gyromultiplication as \cref{eq:gyro_addtion,eq:gyro_scalar_product}:
\begin{align} 
    \label{eq:calmk_gyroadd_def}
    x \MKoplus y 
    &= \rieexp_{x}\left(\pt{\MKzero}{x} \left(\rielog _{\MKzero}(y)\right)\right), \quad \forall x,y \in \calMK{n}, \\
    \label{eq:calmk_gyro_prod_def}
    t \MKodot x
    &= \rieexp _{\MKzero} \left( t \rielog _{\MKzero} (x) \right), \quad \forall t \in \bbRscalar, \forall x \in \calMK{n}.
\end{align}

We give the following clarifications regarding the above two gyro operations.
\begin{itemize}
    \item 
    For hyperbolic geometry ($K < 0$), \cref{eq:calmk_gyroadd_def} has been employed in prior work~\citep{chami2019hyperbolic,he2025lorentzian}. However, there exists no closed-form expression, which could be more efficient than the composition of Riemannian operators. Besides, the underlying gyro-structure has not been formally discussed.
    \item 
    For spherical geometry ($K>0$), the geodesic between antipodal points ($x$ and $-x$) is not unique, making the logarithm and parallel transport along such a geodesic ill-defined. For the sphere, \cref{eq:calmk_gyroadd_def} assumes that $x \neq -\MKzero$ and $y \neq -\MKzero$, while \cref{eq:calmk_gyro_prod_def} assumes that $x \neq -\MKzero$. Like before, we always make these assumptions implicitly.
\end{itemize}

In the following, we first give the closed-form expressions of \cref{eq:calmk_gyroadd_def,eq:calmk_gyro_prod_def}. Then, we show that \cref{eq:calmk_gyroadd_def,eq:calmk_gyro_prod_def} conform with all the axioms of a gyrovector space.

\begin{proposition}[Gyromultiplication and Gyroinverse]
    \label{prop:calmk_scalar_prod_inv}
    \linktoproof{prop:calmk_scalar_prod_inv}
    Let $x=[x_t,x_s^\top]^\top$ be a point in $\calMK{n}$, where $x_t \in \mathbb{R}$ is the time scalar, and $x_s \in \mathbb{R}^n$ is the spatial part. The gyromultiplication and inverse have closed-form expressions:
    \begin{align}
    \label{eq:gyro_prod_calmk}
    t \MKodot x 
    &=
    \begin{cases}
    \MKzero, & t = 0 \lor x=\MKzero, \\
    \frac{1}{\sqrt{|K|}} 
    \begin{bmatrix}
    \cosk \!\left( t \cosk^{-1}(\sqrt{|K|} \, x_t) \right) \\
    \dfrac{\sink \!\left( t \cosk^{-1}(\sqrt{|K|} \, x_t) \right)}{\norm{x_s}} \, x_s
    \end{bmatrix}, & t \neq 0, 
    \end{cases} \\
    \MKominus x &
    = -1 \MKodot x
    = 
    \begin{bmatrix}
    x_t \\
    -x_s
    \end{bmatrix}.
    \end{align}
    In particular, the gyro identity is $\MKzero$. Besides, the following shows that the sphere gyromultiplication is still valid for the singular cases in the projected hypersphere. Assume $K>0$ and $x \neq \pm \MKzero$, and let $\stereo{n} \ni u=\isoMKST{n}(x)$ be its stereographic image. For $t\in \bbRscalar$, set $\theta = \cos^{-1} \left( \sqrt{K} x_t \right) \in (0,\pi)$. Then the following are equivalent:
    \begin{equation*}
    \textup{(i) } t \tan^{-1}\!\left(\sqrt{K} \norm{u} \right) = \tfrac{\pi}{2}+k\pi
      \Longleftrightarrow  
    \textup{(ii) } t \theta = (2k+1)\pi, \quad k \in \bbZ.
    \end{equation*}
    In these singular cases, \cref{eq:gyro_prod_calmk} is still valid, while the stereographic gyromultiplication returns infinity:
    \begin{equation} \label{eq:singular-case-radius}
    t \MKodot x = -\MKzero \in \calMK{n},
    \quad
    t \stodot u = \infty \in \stereo{n},
    \end{equation}
    where $\infty$ denotes the added point in the one-point compactification $\projhs{n}\cup\{\infty\}$ ($\stereo{n}=\projhs{n}$ for $K>0$) and $\isoMKST{n}(-\MKzero)=\infty$.
\end{proposition}

\begin{proposition}[Gyroaddition]
\label{prop:calmk_gyroaddition}
\linktoproof{prop:calmk_gyroaddition}
Let $x=[x_t,x_s^\top]^\top$ and $y=[y_t,y_s ^\top]^\top$ be points in $\calMK{n}$, where $x_t,y_t\in\mathbb{R}$ are the time scalars, and $x_s,y_s\in\mathbb{R}^n$ are the spatial parts. Then, the gyroaddition on $\calMK{n}$ admits the closed form:
\begin{equation} \label{eq:gyroadd_calmk}
    x\MKoplus y = 
    \begin{cases}
    x, & y=\MKzero, \\
    y, & x=\MKzero, \\
    \begin{bmatrix}
    \frac{1}{\sqrt{|K|}} \frac{D - K N}{D + K N} \\
    \frac{2 \left( A_s x_s + A_y y_s \right)}{ D + K N }
    \end{bmatrix},  & \text{Others}.
    \end{cases}
\end{equation}
Here, $A_s = ab^2 - 2K b s_{xy} - K a n_y$ and $A_y = b(a^2 + K n_x)$ with the following notations:
\begin{equation*}
a=1+\sqrt{|K|}x_t, \;
b=1+\sqrt{|K|} y_t,\;
n_x=\norm{x_s}^2, \; 
n_y=\norm{y_s}^2, \; 
s_{xy}=\langle x_s,y_s\rangle.
\end{equation*}
\begin{equation*}
D = a^2b^2 - 2K ab s_{xy} + K^2 n_x n_y, \;
N = a^2 n_y + 2ab s_{xy} + b^2 n_x .
\end{equation*}
Besides, the following shows that the sphere gyroaddition is still valid under the singular cases in the projected hypersphere. Assume $K>0$ and $x,y \neq \pm \MKzero$, and let $u=\isoMKST{n}(x)$ and $v=\isoMKST{n}(y)$ be the stereographic images. The following statements are equivalent:
\begin{enumerate}
\item $u=\dfrac{v}{K\norm{v}^2}$ ($v\neq \Rzero$);
\item $x_s=y_s$ and $x_t=-\,y_t$ (same meridian, mirrored across the equator);
\item $D=0$.
\end{enumerate}
In such singular cases, we have $N>0$ and \cref{eq:gyroadd_calmk} is still valid, while the stereographic gyroaddition returns infinity:
\begin{equation*}
    x \MKoplus y= -\MKzero,
    \quad 
    u \stoplus v=\infty,
\end{equation*}
where $\infty$ is the point added in the one–point compactification of $\projhs{n}$.
\end{proposition}

The above two propositions immediately imply that $\isoMKST{n}$ preserves the gyro operations. 
\begin{corollary}[Isomorphism]
    \label{cor:isomorphism_calmk_stereo}
    \linktoproof{cor:isomorphism_calmk_stereo}
    For the hyperbolic geometry ($K<0$), the isometry $\isoMKST{n}: \bbh{n} \to \pball{n}$ preserves the gyro operations:
    \begin{equation}
        \begin{aligned}
            x \MKoplus y &= \isoSTMK{n} \left( \isoMKST{n}(x) \stoplus \isoMKST{n}(y) \right), \quad \forall x, y \in \calMK{n},\\
            r \MKodot y &= \isoSTMK{n} \left( r \stodot \isoMKST{n}(y) \right), \quad \forall r \in \bbRscalar, \forall x \in \calMK{n}.
        \end{aligned}
    \end{equation}
    For the spherical geometry ($K>0$), the isometry $\isoMKST{n}: \sphere{n} \to \projhs{n} \cup \{\infty\}$ preserves the gyro operations:
    \begin{equation}
        \begin{aligned}
            x \MKoplus y &= \isoSTMK{n} \left( \isoMKST{n}(x) \stoplus \isoMKST{n}(y) \right), \quad \forall x, y \in \calMK{n} / \{ -\MKzero \},\\
            r \MKodot y &= \isoSTMK{n} \left( r \stodot \isoMKST{n}(y) \right), \quad \forall r \in \bbRscalar, \forall x \in \calMK{n} / \{ -\MKzero \}.
        \end{aligned}
    \end{equation}
\end{corollary}

\begin{remark}
    We provide two clarifications regarding \cref{cor:isomorphism_calmk_stereo}.
    \begin{itemize}
        \item 
        \mypara{Compactified projected hypersphere.}
        Although stereographic operations for $K>0$ may be undefined in certain cases, they become well-defined on the one-point compactification $\projhs{n}\cup\{\infty\}$, where undefined cases corresponds to $\infty$.
        
        \item 
        \mypara{Sphere vs.\ Projected Hypersphere.}
        \cref{cor:isomorphism_calmk_stereo} suggests a numerical advantage of the sphere: gyro-operations are well-defined on $\sphere{n}$ at all points except the single south pole $-\MKzero$, including cases corresponding to singularities on the projected hypersphere. This broader domain can make computations on $\sphere{n}$ more stable.
    \end{itemize}
\end{remark}

From the above corollary, it is expected that the operations $\MKoplus$ and $\MKodot$ also satisfy the axioms of a gyrovector space for both negative and positive curvature $K$. 
\begin{theorem}[Radius Gyrovector Spaces] 
\label{thm:calmk_gyrovector}
\linktoproof{thm:calmk_gyrovector}
$(\calMK{n},\MKoplus)$ forms a gyrocommutative gyrogroup, and $(\calMK{n},\MKoplus,\MKodot)$ forms a gyrovector space.\footnote{For $K>0$, we implicitly assume the addition and multiplication are well-defined; whenever they are, all corresponding axioms hold.}
\end{theorem}

Due to the isometry, \cref{lem:isometry_for_gyro} implies gyroisometries over $\calMK{n}$.
\begin{theorem} \label{thm:radius_iso}
On the radius model, the gyrodistance is identical to the geodesic distance, whereas the gyroinverse, gyration, and left gyrotranslation are gyroisometries.
\end{theorem}

\subsubsection{Radius GyroBN}

\cref{thm:radius_iso} guarantees that GyroBN on the radius model normalizes the sample mean and variance. Substituting the operators from \cref{tab:riem-operator-radius,prop:calmk_scalar_prod_inv,prop:calmk_gyroaddition} into \cref{alg:gyrobn} can directly yield the radius GyroBN. For $K<0$ (hyperboloid), the Fréchet mean can be computed efficiently by Algorithm~3 of \citet{lou2020differentiating}; for $K>0$ (projected hypersphere), we compute the Fréchet mean using the Karcher flow \citep{karcher1977riemannian}.

\subsection{Hyperbolic Beltrami--Klein}
\label{subsec:klein_einstein}
The Beltrami--Klein model has recently emerged as a promising alternative to the Poincaré ball for representing hyperbolic geometry \citep{mao2024klein}. As the Poincaré ball admits the Möbius gyrovector space, the Beltrami--Klein model admits the Einstein gyrovector space \citep[Chapter~6.18]{ungar2022analytic}. While prior studies mainly focused on the case $K=-1$ \citep{mao2024klein}, we study the Einstein gyrospace and the Beltrami--Klein geometry under arbitrary negative curvature. This allows us to establish the equivalence between their gyro and Riemannian formulations and, ultimately, to instantiate GyroBN on this model.

\subsubsection{Einstein Gyrovector Space and Beltrami--Klein Geometry}
\label{subsubsec:einstein-gyro-riem}

For $x,y,z \in \klein{n}$ and $t \in \bbRscalar$, the Einstein gyro operations are given by
\begin{align}
    x \Eoplus y 
    &=\frac{1}{1 - K\inner{x}{y}}\left(x+\frac{1}{\gamma_{x}} y -K \frac{\gamma_{x}}{1+\gamma_{x}} \inner{x}{y} x\right), \\
    t \Eodot x
    &=\frac{\tanh \left(t \tanh ^{-1}\left( \sqrt{-K} \|x\| \right)\right)}{\sqrt{-K}}  \frac{x}{\|x\|}, \\
    \gyr[x, y] z
    &= z+\frac{A x+B y}{D},
\end{align}
where $\gamma^K_x =\frac{1}{\sqrt{1+K\|x\|^2}}$ is the gamma factor, and
\begin{equation}
    \begin{aligned}
    A= & K \frac{\gamma_{x}^2}{\left(\gamma_{x}+1\right)}\left(\gamma_{y}-1\right)(\inner{x}{z})-K \gamma_{x} \gamma_{y}(\inner{y}{z}) \\
    & +2K^2 \frac{\gamma_{x}^2 \gamma_{y}^2}{\left(\gamma_{x}+1\right)\left(\gamma_{y}+1\right)}(\inner{x}{y})(\inner{y}{z}), \\
    B= & K \frac{\gamma_{y}}{\gamma_{y}+1}\left\{\gamma_{x}\left(\gamma_{y}+1\right)(\inner{x}{z})+\left(\gamma_{x}-1\right) \gamma_{y}(\inner{y}{z})\right\}, \\
    D= & 1+\gamma_{x} \gamma_{y}\left(1 - K\inner{x}{y}\right)=1+\gamma_{x \Eoplus y}>1.
    \end{aligned}
\end{equation}
Note that the Einstein gyromultiplication coincides with the Möbius gyromultiplication.

For the Poincaré ball, Möbius operations are equivalent to those defined by \cref{eq:gyro_addtion,eq:gyro_scalar_product} \citep[Section 2.4]{ganea2018hyperbolic}. Similarly, under $K=-1$, the Einstein operations coincide with \cref{eq:gyro_addtion,eq:gyro_scalar_product} \citep[Section~4.2]{mao2024klein}. We extend these results to arbitrary $K<0$ and further show that the Beltrami--Klein Riemannian operators can be equivalently expressed in terms of the Einstein gyrospace. To this end, we first establish the isometry between the Beltrami--Klein and Poincaré models.

\begin{proposition}[Beltrami--Klein Isometries]
    \label{props:klein_poincare_isometry_differentials}
    \linktoproof{props:klein_poincare_isometry_differentials}
    The following maps are Riemannian isometries between the Beltrami--Klein and Poincaré ball models:
    \begin{align*}
        \pi_{\klein{n} \to \pball{n}}
        &: \klein{n} \ni x  \longmapsto
        \frac{1}{ 1 +  \sqrt{1 + K \norm{x}^2} } x \in \pball{n},\\
        \pi_{\pball{n} \to \klein{n}} 
        &: \pball{n} \ni x \longmapsto \frac{2}{1 - K \norm{x}^2} x \in \klein{n}.
    \end{align*}
    Particularly, $\pi_{\pball{n} \to \klein{n}}(\Rzero)=\Rzero$. Given $x$ in the hyperbolic model $\calH \in \{ \klein{n}, \pball{n}\}$ and tangent vector $v \in T _x \calH$, the differential maps of $\pi_{\klein{n} \to \pball{n}}$ and $\pi_{\pball{n} \to \klein{n}}$ are
    \begin{align*}
    (\pi_{\klein{n} \to \pball{n}})_{*,x} (v)
    &= \frac{1}{1 + \sqrt{1+K\norm{x}^2}} v 
    - \frac{K \inner{x}{v}}{\left(1 + \sqrt{1+K\norm{x}^2}\right)^2 \sqrt{1+K\norm{x}^2}} x, \\
    (\pi_{\pball{n} \to \klein{n}})_{*,x} (v) 
    &= \frac{2}{\left(1-K\norm{x}^2\right)}v
    + \frac{4 K \inner{x}{v} }{\left(1-K\norm{x}^2\right)^2 }x.
    \end{align*}
    Especially, the differential maps at the zero vector are 
    \begin{align}
        \label{app:eq:k2p_diff_0}
        (\pi_{\klein{n} \to \pball{n}})_{*,{\Rzero}} (v) 
        &= \frac{1}{2} v, \\
        \label{app:eq:p2k_diff_0}
        (\pi_{\pball{n} \to \klein{n}})_{*,{\Rzero}} (v) 
        &= 2 v.        
    \end{align}
    Moreover, these isometries preserve gyroaddition and gyromultiplication:
    \begin{equation}
        \label{eq:iso_addition}
        \begin{aligned}
        \pi_{\pball{n} \to \klein{n}}(x \Moplus y) &= \pi_{\pball{n} \to \klein{n}}(x) \Eoplus \pi_{\pball{n} \to \klein{n}}(y), \quad \forall x,y \in \pball{n}, \\
        \pi_{\pball{n} \to \klein{n}} (t \Modot x) &= t \Eodot \pi_{\pball{n} \to \klein{n}}(x), \quad \forall t \in \bbRscalar, \forall x \in \pball{n},
    \end{aligned}
    \end{equation}
    where $\Moplus$ and $\Modot$ are Möbius operations, while $\Eoplus$ and $\Eodot$ are the Einstein counterparts.
\end{proposition}

\begin{theorem}[Einstein by Beltrami--Klein]
    \label{thm:einstein_klein}
    \linktoproof{thm:einstein_klein}
    The Einstein gyro operations can be rewritten as \cref{eq:gyro_addtion,eq:gyro_scalar_product}: 
    \begin{align*}
        x \Eoplus y &= \rieexp _x \left(\pt{{\Rzero}}{x} (\rielog _{\Rzero} (y)) \right), \quad \forall x,y \in \klein{n}, \\
        t \Eodot x &= \rieexp _{\Rzero} ( t \rielog _{\Rzero} (x)), \quad \forall x \in \klein{n}, \forall t \in \bbRscalar.
    \end{align*}
\end{theorem}

\cref{thm:einstein_klein} demonstrates that the Einstein gyro-structure can be expressed by the Beltrami--Klein geometry. Conversely, the Beltrami--Klein geometry can also be formulated by the Einstein gyro-structure.

\begin{theorem} [Beltrami--Klein by Einstein]
    \label{thm:riem_klein}
    \linktoproof{thm:riem_klein}
    Given $x, y \in \klein{n}$ and $v \in T _x \klein{n}$, the distance, exponential, and logarithmic operators under the Beltrami--Klein geometry are
    \begin{align}
        \dist (x, y) 
        & =\frac{2}{\sqrt{|K|}} \tanh ^{-1}\left(\sqrt{|K|} \frac{ \norm{-x \Eoplus y}}{1+\sqrt{1+K\left\|-x \Eoplus y\right\|^2}} \right), \\
        \label{eq:exp_klein}
        \rieexp _{x}(v) 
        &= x \Eoplus \rieexp _{\Rzero}\left( \frac{1}{\sqrt{1+K\norm{x}^2}} v 
            - \frac{K \inner{x}{v}}{\left(1 + \sqrt{1+K\norm{x}^2}\right) (1+K\norm{x}^2)} x  \right), \\ 
        \label{eq:log_klein}
        \rielog _x (y) 
        &= \frac{1}{\lambda_{\widetilde{x}}^K} (\pi_{\pball{n} \to \klein{n}}) _{*, \widetilde{x}}\left(\rielog _{\Rzero}\left( -x \Eoplus  y\right)\right),
    \end{align}
    where $\widetilde{x} = \pi_{\klein{n} \to \pball{n}}(x)$. Particularly, the exponential and logarithmic maps at the zero vector $\Rzero$ are identical across the Beltrami--Klein and Poincaré ball models:
    \begin{align}
        \label{eq:exp_0_klein}
        \rieexp_{\Rzero}(v) &= \tanh (\sqrt{|K|}\norm{v}) \frac{v}{\sqrt{|K|}\norm{v}}, \quad \forall v \in T_\Rzero \calH,\\
        \label{eq:log_0_klein}
        \rielog_{\Rzero}(x) &= \tanh ^{-1}(\sqrt{|K|}\norm{x}) \frac{x}{\sqrt{|K|}\norm{x}}, \quad \forall x \in \calH,
    \end{align}
    with $\calH \in \{ \klein{n}, \pball{n} \}$.
\end{theorem}
\begin{remark}
    Since the Beltrami--Klein and Poincaré ball models share the same $\rieexp_{\Rzero}$ and $\rielog_{\Rzero}$, it naturally follows that the Einstein and Möbius gyromultiplication coincide.
\end{remark}

As the Beltrami--Klein is isometric to the Poincaré ball, \cref{lem:isometry_for_gyro} implies gyroisometries.
\begin{theorem} \label{thm:klein_iso}
On the Beltrami--Klein model, the gyrodistance is identical to the geodesic distance, whereas the gyroinverse, gyration, and left gyrotranslation are gyroisometries.
\end{theorem}

\subsubsection{Beltrami--Klein GyroBN}
\label{subsubsec:klein-gyrobn}

\cref{thm:klein_iso} ensures that GyroBN on the Beltrami--Klein model normalizes the sample mean and variance. Owing to the isometry between the Beltrami--Klein and Poincaré models, the Fréchet mean can be computed via the Poincaré ball: map the data to the Poincaré model using $\pi_{\klein{n} \to \pball{n}}$, compute the Poincaré Fréchet mean \citep[Algorithm~1]{lou2020differentiating}, and map the result back using $\pi_{\pball{n} \to \klein{n}}$. Together with the gyro and Riemannian operators in \cref{subsubsec:einstein-gyro-riem}, we have all the ingredients to implement \cref{alg:gyrobn}.

\subsection{Correlation Manifolds}
\label{subsec:cor_manifolds}

\citet[Section~4.1]{thanwerdas2022theoretically} show that any correlation matrix can be identified with a product of hyperbolic spaces via its Cholesky decomposition. Given $C \in \cor{n}$, let $L=\chol(C)$ be its Cholesky factor. The $k$-th row of $L$ has the form $(L_{k1},\ldots,L_{k,k-1},L_{kk},0,\ldots,0)$ with $L_{kk}>0$, which belongs to the hyperbolic open hemisphere\footnote{Also known as the Jemisphere model, where the ``J'' is pronounced as in Spanish \citep[Section~7]{cannon1997hyperbolic}.} $\mathrm{HS}^{k-1} =\left\{x \in \mathbb{R}^k \mid\|x\|=1, x_k>0 \right\}$. As shown by \citet{chen2025cornet}, $\hs{n}$ is isometric to the unit Poincaré ball $\unitpball{n}=\left\{x \in \mathbb{R}^{n} \mid \|x\| < 1 \right\}$ by 
\begin{equation*}
    \pi _{\hs{n} \rightarrow \unitpball{n}} \left(
    \begin{bmatrix}
        x \\
        x_{n+1}
    \end{bmatrix} \right)
    = \frac{x}{1+x_{n+1}}.
\end{equation*}
Therefore, each correlation matrix can be identified with $n-1$ Poincaré vectors \citep{chen2025cornet}:
\begin{equation} \label{eq:iso_cor_phc}
    \cor{n} \ni C {\mapsto} 
    \begin{bmatrix}
    1 & 0 & \cdots & 0 \\
    L_{21} & L_{22} & \cdots & 0 \\
    \vdots & \vdots & \ddots & \vdots \\
    L_{n1} & L_{n2} & \cdots & L_{nn}
    \end{bmatrix} 
    {\mapsto} 
    \begin{bmatrix}
    x_1 \in \unitpball{1} \\
    \vdots \\
    x_{n-1} \in \unitpball{n-1} 
    \end{bmatrix}.
\end{equation}
where $x_{i} = \pi _{\hs{i} \rightarrow \unitpball{i}} \left(L_{(i+1, 1)}, \cdots, L_{(i+1, i+1)} \right)^\top$ corresponds to the $(i+1)$-th row of the Cholesky factor. Let $\bbPPB{n-1} = \prod_{i=1}^{n-1}\unitpball{i}$ denote the product of unit Poincaré balls. We denote the identification in \cref{eq:iso_cor_phc} by $\Phi: \cor{n} \to \bbPPB{n-1}$. GyroBN on the correlation manifold can then be realized via the Poincaré GyroBN applied row-wise: first map $C$ to $\bbPPB{n-1}$ via $\Phi$, apply $\text{GyroBN}_i$ independently on each $\unitpball{i}$, and finally map back with $\Phi^{-1}$. For a batch of activations $\{C^i\}_{i=1}^N \subset \cor{n}$, the process can be expressed as
\begin{equation}
    \forall i \leq N, \quad C^i \overset{\Phi}{\longmapsto}  
    \begin{bmatrix}
    x^i_1 \in \unitpball{1} \\
    \vdots \\
    x^i_{n-1} \in \unitpball{n-1} 
    \end{bmatrix}
    \begin{array}{c}
         \overset{\text{GyroBN}_1}{\longmapsto}  \\
         \vdots \\
         \overset{\text{GyroBN}_{n-1}}{\longmapsto}  \\ 
    \end{array}
    \begin{bmatrix}
    \widetilde{x}^i_1 \in \unitpball{1} \\
    \vdots \\
    \widetilde{x}^i_{n-1} \in \unitpball{n-1} 
    \end{bmatrix}
    \overset{\Phi^{-1}}{\longmapsto}
    \widetilde{C}^i.
\end{equation}

\subsection{Summary}

\begin{table}[t]
\centering
\renewcommand{\arraystretch}{1.2}
\begin{subtable}[t]{\linewidth}
    \centering
    \begin{tabular}{cccc}
    \toprule
    $U \oplusGyrONB V$ & $\ominusGyrONB U$ & $t \odotGyrONB U$ & Fréchet mean \\
    \midrule
     $\mexp(\Omega) V$  
    & $\mexp(-\Omega) \idonb$
    & $ \mexp(t\Omega) \idonb$
    & Karcher flow \\
    \bottomrule
    \end{tabular}
    \caption{The ONB Grassmannian}
    \end{subtable}
    
    
    \begin{subtable}[t]{\linewidth}
    \centering
    \resizebox{\linewidth}{!}{
    \begin{tabular}{cccc}
    \toprule
    Operator & $\stereo{n}$ & $\calMK{n}$ & $\klein{n}$ \\
    \midrule
    $x \oplus y$
    & $\displaystyle \frac{(1-2K\langle x,y\rangle - K\|y\|^{2})\,x + (1+K\|x\|^{2})\,y}{1 - 2K\langle x,y\rangle + K^{2}\|x\|^{2}\|y\|^{2}}$
    & \cref{eq:gyroadd_calmk}
    & $\displaystyle \frac{1}{1 - K\inner{x}{y}}\left(x+\frac{1}{\gamma_{x}} y -K \frac{\gamma_{x}}{1+\gamma_{x}} \inner{x}{y} x\right)$ \\
    $\ominus x$ & $-x$ 
    & $\displaystyle \begin{bmatrix}
    x_t \\
    -x_s
    \end{bmatrix}$ & $-x$ \\
    $t \odot x$ 
    & $\displaystyle \frac{\tanh \left(t \tanh ^{-1}\left( \sqrt{-K} \|x\| \right)\right)}{\sqrt{-K}}  \frac{x}{\|x\|}$
    & \cref{eq:gyro_prod_calmk}
    & $\displaystyle \frac{\tanh \left(t \tanh ^{-1}\left( \sqrt{-K} \|x\| \right)\right)}{\sqrt{-K}}  \frac{x}{\|x\|}$\\
    Fréchet mean 
    & \makecell{$K<0$: Algorithm 1 \citep{lou2020differentiating} \\ $K>0$: Karcher flow}
    & \makecell{$K<0$: Algorithm 3 \citep{lou2020differentiating} \\ $K>0$: Karcher flow}
    & \makecell{Via Poincaré ball \\ (see \cref{subsubsec:klein-gyrobn})} \\
    \bottomrule
    \end{tabular}
    }
    \caption{Constant curvature spaces}
    \end{subtable}
    \caption{Summary of operators for GyroBN across representative manifolds.}
\label{tab:gyrobn-summary}
\end{table}

To conclude this section, \cref{tab:gyrobn-summary} summarizes the key gyro operators needed to implement GyroBN on representative manifolds. The correlation manifold is excluded, since its GyroBN is realized row-wise through the Poincaré ball.

\section{Experiments}
\label{sec:exp}
Our GyroBN layers are model-agnostic and can be seamlessly integrated into networks over gyrogroups. This section evaluates GyroBN on the Grassmannian, five constant curvature spaces, and the correlation manifold. The main findings are summarized as follows.
\begin{keybox}
\begin{itemize}
    \item 
    \mypara{Numerical Experiments (\cref{subsec:exp:numerical}).} 
    The closed-form expression derived for radius gyroaddition in \cref{eq:gyroadd_calmk} significantly accelerate the computation compared to their definition-based counterpart in \cref{eq:calmk_gyroadd_def}, achieving $2 \times$–$3 \times$ speedups. Furthermore, visualizations demonstrate that GyroBN effectively normalizes sample distributions across diverse geometries.      
    \item 
    \mypara{Performance (\cref{subsec:exp:grass_nn,subsec:exp:ccs_nn,subsec:exp:cor_nn}).} 
    On Grassmannian, constant curvature, and correlation networks, GyroBN consistently improves backbone networks, whereas existing RBN methods often degrade performance. Compared to these methods, GyroBN is generally faster or comparable in runtime, requires fewer or equal parameters, and improves robustness.
    \item 
    \mypara{Discussions (\cref{subec:exp:discussions}).} 
    Ablations further validate the design:
    (i) covariate shift is evident in manifold networks, underscoring the necessity of normalization; 
    (ii) GyroBN consistently reduces the condition numbers of both weight matrices and network Jacobians, thereby stabilizing optimization; 
    (iii) Fréchet mean iterations need not to be run to full convergence, since one or two iterations are generally sufficient, which provides a favorable trade-off between efficiency and accuracy.
\end{itemize}
\end{keybox}

\subsection{Numerical Experiments}
\label{subsec:exp:numerical}

\subsubsection{Efficiency of the Closed-form Radius Gyroaddition}
\label{subsubsec:efficiency_radius_gyroadd}

\begin{table}[t]
  \centering
  \resizebox{0.7\linewidth}{!}{
    \begin{tabular}{c|cc|cc}
    \toprule
    Geometry & \multicolumn{2}{c|}{Hyperboloid} & \multicolumn{2}{c}{Sphere} \\
    \midrule
    Dim & Riemannian & \cellcolor{HilightColor}Close-form & Riemannian & \cellcolor{HilightColor}Close-form \\
    \midrule
    16    & 361.22 & \cellcolor{HilightColor}\textbf{121.68 (33.69\%)} & 323.58 & \cellcolor{HilightColor}\textbf{122.02 (37.71\%)} \\
    32    & 363.41 & \cellcolor{HilightColor}\textbf{123.08 (33.87\%)} & 327.75 & \cellcolor{HilightColor}\textbf{123.71 (37.74\%)} \\
    64    & 387.94 & \cellcolor{HilightColor}\textbf{181.50 (46.78\%)} & 453.42 & \cellcolor{HilightColor}\textbf{181.05 (39.93\%)} \\
    128   & 574.37 & \cellcolor{HilightColor}\textbf{272.10 (47.37\%)} & 644.18 & \cellcolor{HilightColor}\textbf{270.21 (41.95\%)} \\
    256   & 1149.58 & \cellcolor{HilightColor}\textbf{534.09 (46.46\%)} & 1137.62 & \cellcolor{HilightColor}\textbf{538.58 (47.34\%)} \\
    1024  & 3364.23 & \cellcolor{HilightColor}\textbf{1414.67 (42.05\%)} & 3551.69 & \cellcolor{HilightColor}\textbf{1421.67 (40.03\%)} \\
    2048  & 6479.95 & \cellcolor{HilightColor}\textbf{2497.47 (38.54\%)} & 6930.69 & \cellcolor{HilightColor}\textbf{2449.56 (35.34\%)} \\
    \bottomrule
    \end{tabular}%
    }
    \caption{Efficiency (in~$\mu$s) of gyroaddition on the radius manifold: closed form v.s. Riemannian definition. Values in parentheses indicate the runtime of the closed-form implementation as a percentage of the corresponding Riemannian implementation.}
  \label{tab:res:efficiency_gyroadd}%
\end{table}%

Recalling \cref{subsubsec:radius_gyro_space}, we derive closed-form expressions for the radius gyrooperations to improve computational efficiency. To assess this, we compare two variants of radius gyroaddition: (i) the definition-based operator \cref{eq:calmk_gyroadd_def}, implemented via a composition of the Riemannian logarithm, parallel transport, and exponential map; and (ii) the closed-form operator \cref{eq:gyroadd_calmk}. We report the mean wall-clock time (in~$\mu$s), averaged over 100 runs with a batch size of 10,000, across varying dimensions. As shown in \cref{tab:res:efficiency_gyroadd}, our closed-form implementation consistently outperforms its definition-based counterpart, achieving speedups of roughly $2{\times}$–$3{\times}$ across the hyperboloid and sphere.

\subsubsection{Visualization of GyroBN on Different Geometries}
\label{subsubsec:exp:visualization}

\begin{figure}[t]
\centering
\includegraphics[width=\linewidth,trim={0cm 1.5cm 0cm 0.5cm}]{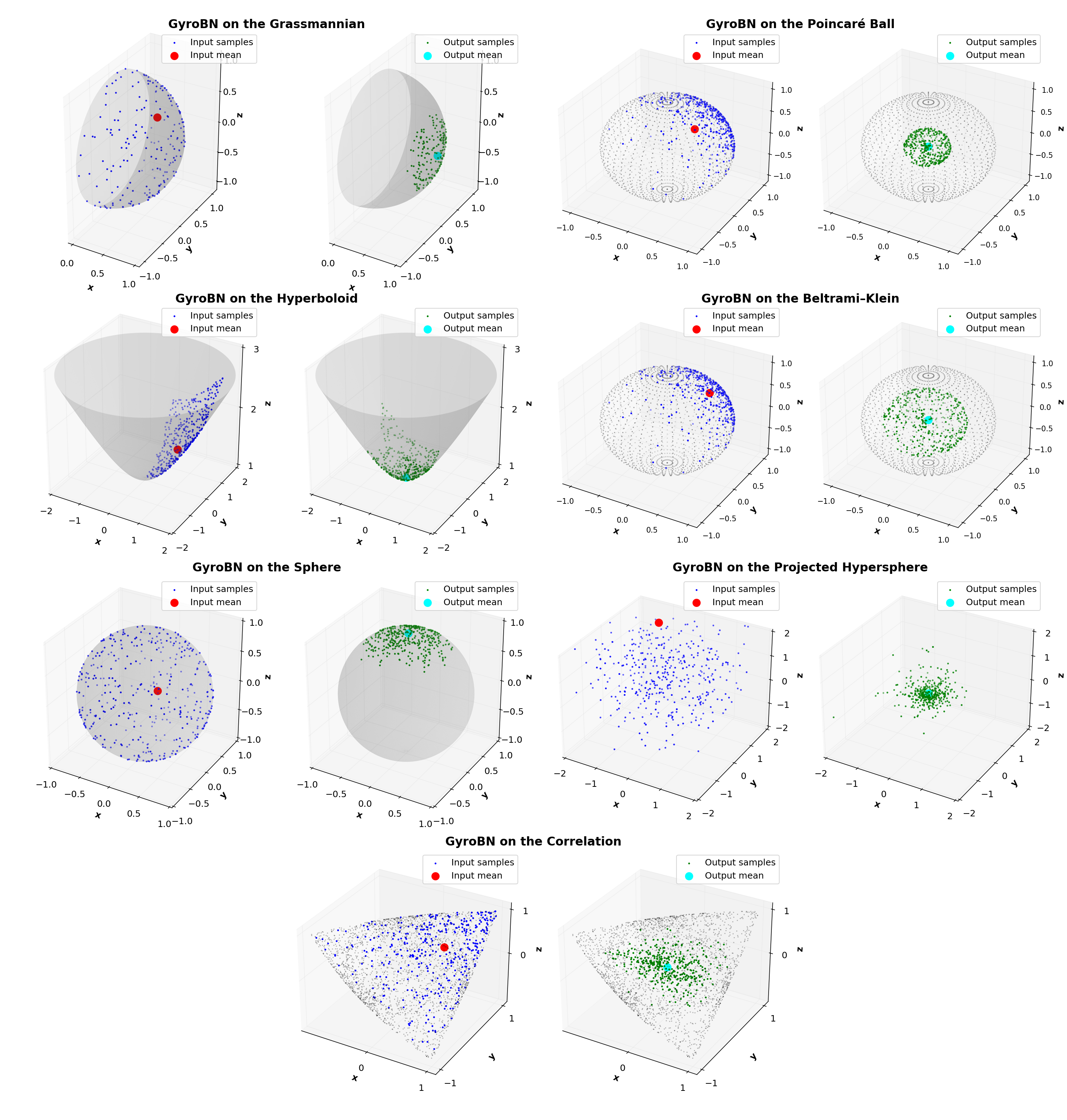}
\caption{Visualization of GyroBN across different geometries. Blue and green points represent input and normalized data, respectively. Red and cyan points denote the input and output batch means. Black points mark the manifold boundary, and the gray surface depicts the manifold.}
\label{fig:illustration_gyrobn_grass_ccs_cor}
\vspace{-5mm}
\end{figure}

To intuitively illustrate the effect of GyroBN, we visualize its behavior on the ONB Grassmannian $\grasonb{1,3}$, five constant curvature spaces with $|K|=1$, and the correlation manifold $\cor{3}$. For each geometry, we randomly generate a batch of points, compute their batch mean, apply GyroBN, and then plot the normalized batch along with the resulting mean. The visualizations are constructed using the following embeddings:
\begin{itemize}
    \item 
    \mypara{Grassmannian.}
    Since $\grasonb{1,3}$ is homeomorphic to the real projective space $\rp{2}$, it is depicted as the unit hemisphere with antipodal points identified.
    \item 
    \mypara{Constant Curvature Spaces.}
    The unit Poincaré ball $\unitpball{3}$ and unit Beltrami--Klein ball $\unitklein{3}$ are shown as the interior of the unit ball in $\bbR{3}$. The unit hyperboloid $\unitbbh{2}$ is visualized as the upper sheet of a two-sheeted hyperboloid in $\bbR{3}$. The unit sphere $\unitsphere{2}$ is embedded as a $2$-sphere in $\bbR{3}$, while the projected hypersphere $\unitprojhs{2}$ coincides with $\bbR{3}$ itself.
    \item 
    \mypara{Correlation Manifold.}
    $\cor{3}$ is embedded in $\bbR{3}$ as an open elliptope using its strictly lower triangular part.
\end{itemize}

For better visualization, we fix the bias parameter to the gyro identity element and set the scaling parameter to $0.4$ for the Grassmannian, correlation manifold, and sphere; $0.7$ for the Poincaré ball, Beltrami--Klein ball, and hyperboloid; and $1$ for the projected hypersphere. As shown in \cref{fig:illustration_gyrobn_grass_ccs_cor}, GyroBN consistently normalizes data distributions across these geometries. Notably, although the Poincaré and Beltrami--Klein inputs are identical, their GyroBN behavior and resulting sample distributions differ due to their distinct Riemannian metrics, underscoring that GyroBN faithfully respects the underlying geometry.

\subsection{Experiments on Grassmannian Neural Networks}
\label{subsec:exp:grass_nn}

\subsubsection{Setup}

\mypara{Data sets and Preprocessing.}
In line with previous work \citep{huang2018building,nguyen2023building}, we evaluate our method on three skeleton-based action recognition tasks, including HDM05 \citep{muller2007documentation}, NTU60 \citep{shahroudy2016ntu}, and NTU120 \citep{liu2019ntu} data sets, focusing on mutual actions for NTU60 and NTU120. Each sequence is represented as a Grassmannian matrix of size $93 \times 10$, $150 \times 10$, and $150 \times 10$ for HDM05, NTU60, and NTU120, respectively. Additional details are provided in \cref{app:subsubsec:datasets_preprocessing_grass}.

\mypara{Comparative Methods.}
Since no Grassmannian-specific batch normalization methods exist, we adapt two previous approaches to the Grassmannian: ManifoldNorm~\citep[Algorithms~1--2]{chakraborty2020manifoldnorm} and the Riemannian batch normalization method of \citet[Algorithm~2]{lou2020differentiating}, which we denote LRBN for clarity. Both methods are briefly reviewed in \cref{app:subsec:manifoldnorm_lrbn}. Although neither was originally designed for the Grassmannian, they can be adapted by employing Riemannian operators such as geodesics, exponential/logarithmic maps, and parallel transport. \textit{The key difference is that our GyroBN can normalize data distributions across different geometries, whereas the other two methods cannot.}

\mypara{Backbone Networks.}
We adopt the recently proposed GyroGr architecture \citep{nguyen2023building} as our backbone, which is briefly reviewed in \cref{app:subsubsec:gyrogr}. GyroGr replaces the non-intrinsic FRMap + ReOrth block in GrNet \citep{huang2018building} with Grassmannian left gyrotranslation, thereby improving numerical stability and performance. It consists of three basic components: left gyrotranslation, pooling \citep{huang2018building}, and the Projection Map (ProjMap) \citep{huang2018building}, where ProjMap maps Grassmannian points to symmetric matrices for classification. We consider both the 1-block and $L$-block variants. The 1-block version is structured as: gyrotranslation $\rightarrow$ pooling $\rightarrow$ ProjMap $\rightarrow$ classification, where the classification is implemented as a \gls{FC} layer with softmax. The $L$-block version stacks $L$ blocks of gyrotranslation and pooling, followed by a final ProjMap and classification layer. Since each pooling step approximately halves the dimensionality, we omit the pooling operation in the last block when $L > 1$. Following \citet{huang2018building}, the number of channels is fixed to $8$.

\mypara{Implementation Details.}
Following \citet{nguyen2023building}, we use the Cayley map to approximate the matrix exponential of skew-symmetric matrices, and apply the trivialization trick \citep{lezcano2019trivializations} to parameterize the Grassmannian variables in both the gyrotranslation and our GyroBN layers. Specifically, each Grassmannian parameter is trivialized via the exponential map at the identity, as detailed in \cref{app:subsubsec:trivialization_details_grass}, ensuring that all parameters lie in Euclidean spaces. This allows direct use of PyTorch optimizers \citep{paszke2019pytorch} and avoids the additional cost of Riemannian optimization. In contrast, we find that the Grassmannian LRBN benefits from Riemannian optimization. Thus, we employ \texttt{geoopt} \citep{kochurov2020geoopt} to optimize its Grassmannian bias parameter. Similarly, we use \texttt{geoopt} to update the orthogonal bias parameter in ManifoldNorm. For all models, the BN layer is inserted after the first pooling layer with a momentum of $0.1$. Training uses SGD with a learning rate of $5e^{-2}$, batch size $30$, and $400$, $200$, and $200$ epochs for HDM05, NTU60, and NTU120, respectively. All models are optimized with a standard cross-entropy loss. Following previous normalization methods on matrix manifolds \citep{kobler2022spd,chen2024liebn,wang2025gbwmbn}, we adopt a single Fréchet mean iteration.

\subsubsection{Main Results}
\label{subsubsec:main_results_grass}
\begin{table}[tbp]
  \centering
  \resizebox{0.99\linewidth}{!}{
    \begin{tabular}{c|ccc|ccc|ccc}
    \toprule
    \multirow{2}[4]{*}{Method} & \multicolumn{3}{c|}{\makecell{HDM05 \\ (47 × 10)}} & \multicolumn{3}{c|}{\makecell{NTU60 \\ (75 × 10)}} & \multicolumn{3}{c}{\makecell{NTU120 \\ (75 × 10)}} \\
    \cmidrule{2-10}          & Acc   & Fit Time &  \#Params & Acc   & Fit Time &  \#Params & Acc   & Fit Time &  \#Params \\
    \midrule
    GyroGr & 48.97±0.24 & 2.09  & 2.0744 & 70.13±0.16 & 28.16  & 0.5062 & 53.76±0.18 & 49.62  & 1.1812 \\
    GyroGr-ManifoldNorm & 49.67±0.76 & 32.90  & \red{2.0921} & 68.56±0.43 & 232.60  & \red{0.5512} & 51.41±0.38 & 399.78  & \red{1.2262} \\
    GyroGr-LRBN & 48.64±0.77 & 33.31  & 2.0781 & 67.77±0.52 & 238.53  & 0.5122 & 50.56±0.22 & 403.40  & 1.1872 \\
    \midrule
    \rowcolor{HilightColor} GyroGr-GyroBN & \textbf{51.89±0.37} & 3.04  & 2.0773 & \textbf{72.60±0.04} & 35.85  & 0.5114 & \textbf{55.47±0.10} & 67.37  & 1.1864 \\
    \bottomrule
    \end{tabular}%
    }
    \caption{Comparison of GyroBN against other Grassmannian BN methods under the GyroGr backbone. Here, fit time denotes the average training time per epoch (s/epoch), and \#Params denotes the number of parameters. Values in parentheses specify the dimension of the Grassmannian input to the BN layer. The best results are marked in \textbf{bold}, while the largest number of parameters is marked in \red{red}.}
  \label{tab:grass_results}
    \vspace{-2mm}
\end{table}

\begin{figure}[tbp]
\centering
\includegraphics[width=0.9\linewidth,trim={0cm 0cm 0cm 0cm}]{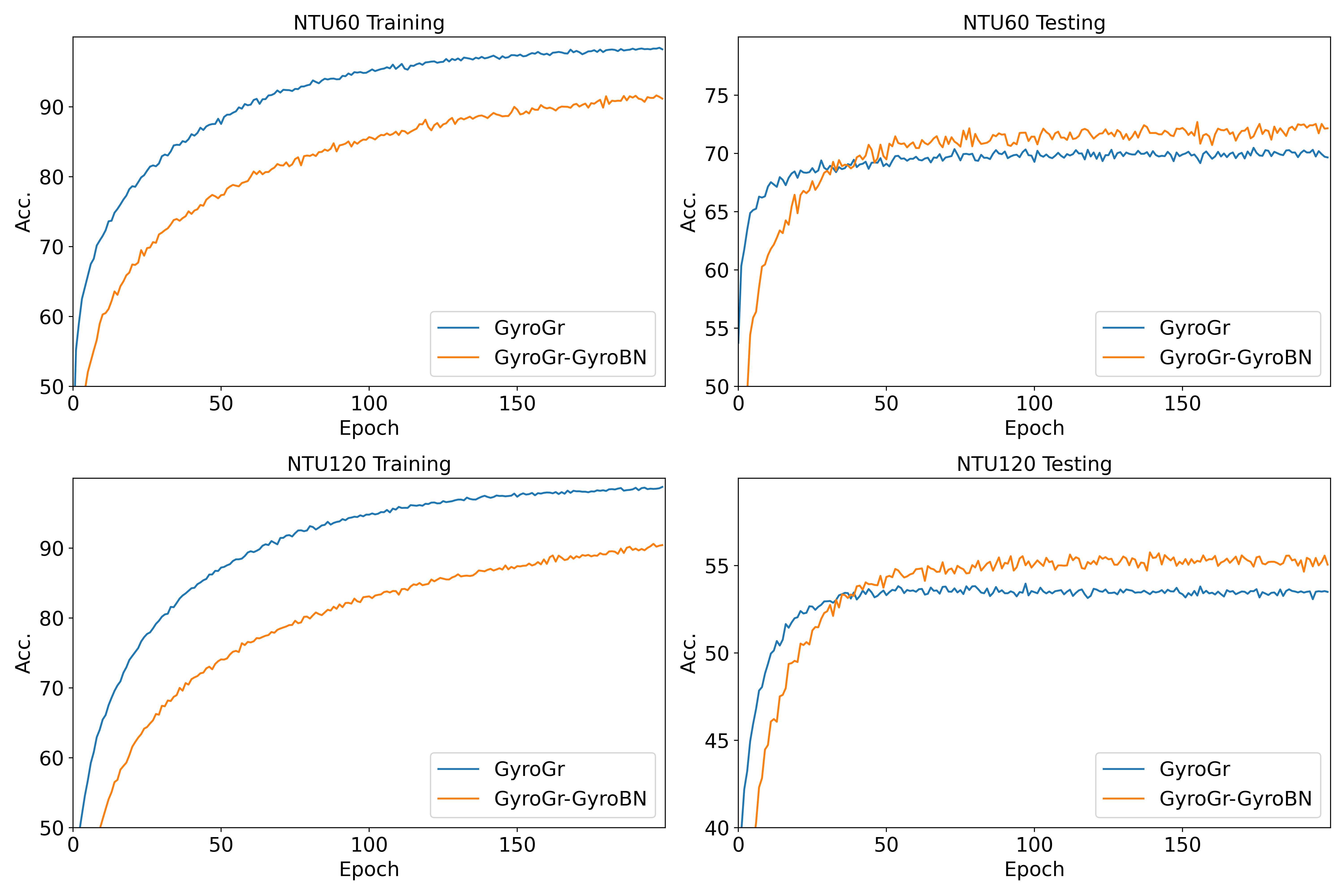}
\caption{Training and testing curves of 1-block GyroGr on two NTU data sets.
}
\label{fig:gyrobn_generalization}
\end{figure}

We compare our GyroBN with ManifoldNorm and LRBN under the 1-block GyroGr backbone. The 5-fold results are presented in \cref{tab:grass_results}. We have the following four findings, which highlight the effectiveness of our GyroBN in facilitating network training.
\begin{itemize}
    \item 
    \mypara{Improved Accuracy.}
    Across all three data sets, GyroBN consistently improves performance, enhancing the accuracy of the vanilla GyroGr by +2.92\%, +2.47\%, and +1.71\% on HDM05, NTU60, and NTU120, respectively. In contrast, both ManifoldNorm and LRBN often degrade performance, particularly on NTU60 and NTU120. This advantage comes from the theoretical guarantee of GyroBN for normalizing sample statistics, which is absent in the other two methods (see \cref{tab:rbn_summary}).
    
    \item 
    \mypara{Enhanced Efficiency.}
    GyroBN is substantially more efficient than ManifoldNorm and LRBN. 
    The efficiency gain is mainly attributed to: (i) replacing computationally expensive Riemannian operators (e.g., parallel transport, exponential/logarithmic maps) with simpler gyro operations, 
    (ii) reducing matrix multiplications from $n \times p$ to $(n-p)\times p$ or $p \times p$ (see \cref{prop:fast_bracket_grassmannian}), 
    and (iii) applying trivialization to avoid costly Riemannian optimization. 
    A detailed analysis is provided in \cref{app:subsubsec_analysis_gyrobn_efficiency}.
    
    \item
    \mypara{Improved Parameter Economy.}
    GyroBN also requires fewer parameters than LRBN and ManifoldNorm. The key difference lies in the bias parameter. GyroBN, due to the trivialization in \cref{app:subsubsec:trivialization_details_grass}, only needs an $(n-p)\times p$ Euclidean matrix, whereas LRBN requires an $n\times p$ Grassmannian matrix 
    and ManifoldNorm requires an $n\times n$ orthogonal matrix. 
    
    \item  
    \mypara{Stronger Generalization.}
    As shown in \cref{fig:gyrobn_generalization}, we observe that GyroBN can narrow the gap between training and testing accuracy, indicating a stronger generalization ability. 
\end{itemize}

\subsubsection{Ablations on Architectures}
\begin{table}[tbp]
  \centering
  \resizebox{0.99\linewidth}{!}{
    \begin{tabular}{c|cccc|cccc|cccc}
    \toprule
    \multirow{2}[4]{*}{Method} & \multicolumn{4}{c|}{HDM05}    & \multicolumn{4}{c|}{NTU60}    & \multicolumn{4}{c}{NTU120} \\
    \cmidrule{2-13}
    & 1-Block & 2-Block & 3-Block & 4-Block & 1-Block & 2-Block & 3-Block & 4-Block & 1-Block & 2-Block & 3-Block & 4-Block \\
    \midrule
    GyroGr & 49.23 & 49.09 & 47.02 & 27.36 & 70.32 & 70.14 & 70.23 & 65.03 & 53.96 & 54.1  & 54.59 & 47.59 \\
    \midrule
    \rowcolor{HilightColor} GyroGr-GyroBN & \textbf{52.43} & \textbf{50.62} & \textbf{51.56} & \textbf{30.29} & \textbf{72.65} & \textbf{71.93} & \textbf{72.25} & \textbf{66.67} & \textbf{55.59} & \textbf{56.15} & \textbf{54.63} & \textbf{48.9} \\
    \bottomrule
    \end{tabular}%
    }
    \caption{Ablation of Grassmannian GyroBN under various network architectures.}
  \label{tab:grass_ablation_architectures}%
\end{table}

We further validate GyroBN across different GyroGr architectures with up to four blocks of gyrotranslation and pooling. As shown in \cref{tab:grass_results}, both GyroGr and GyroGr-GyroBN exhibit relatively small variances, so we perform ablations using a single trial. The results of all three data sets are reported in \cref{tab:grass_ablation_architectures}. We find that GyroBN consistently improves the vanilla GyroGr baseline, highlighting the effectiveness of the proposed framework. Notably, as network depth increases, the performance of GyroGr, with or without GyroBN, declines. This degradation arises because the dimensionality of the final feature representation becomes excessively low, leading to underfitting. For example, in the deepest four-block architecture on the HDM05 data set, the final feature dimension is $12 \times 10$, which is insufficient to capture discriminative information. Nevertheless, GyroBN still provides consistent improvements over GyroGr across all settings.

\subsection{Experiments on Constant Curvature Neural Networks}
\label{subsec:exp:ccs_nn}

\subsubsection{Setup}

\mypara{Data sets.}
Following \citet{lou2020differentiating}, we focus on the link prediction task on four graph data sets: Cora \citep{sen2008collective}, Disease \citep{anderson1991infectious}, Airport \citep{zhang2018link}, and Pubmed \citep{namata2012query}. Details can be found in \cref{app:subsec:datasets_ccs}.

\mypara{Comparative Methods.}
We compare our GyroBN with LRBN \citep[Algorithm 2]{lou2020differentiating}. As the original LRBN is only implemented in the Poincaré ball, we extend it to the other four constant curvature spaces. In particular, the core difference between LRBN and our GyroBN lies in normalization: Our GyroBN can normalize sample statistics across different geometries, while LRBN lacks this guarantee. 

\mypara{Backbone Networks.}
We use HNN \citep{ganea2018hyperbolic} for the Poincaré ball and Klein HNN (KNN) \citep{mao2024klein} for the Beltrami--Klein. For the sphere and projected hypersphere, we mimic the transformation and activation in HNN \citep[Section 3.2]{ganea2018hyperbolic} to build the corresponding layers. The layers on the four spaces above can be expressed as
\begin{align}
    \label{eq:tans_ccnn}
    \text{Transformation: } x^k &= \rieexp _e (M^k \rielog _e(x^{k-1})) \oplus b^k, \text{ with } b^k \in \calN, \text{ and } M^k \in \bbR{m \times n}, \\
    \label{eq:act_ccnn}
    \text{Activation: } x^k &= \rieexp _e ( \phi(\rielog _e(x^{k-1}))), \text{ with } \phi \text{ as an activation},
\end{align}
where $e$ is the origin, $\oplus$ is the gyroaddition, and $\calN \in \{\pball{n}, \klein{n}, \sphere{n}, \projhs{n}\}$. For the hyperboloid, we use the Lorentz fully-connected layer \citep[Equation 3]{chen2021fully} and Lorentz activation layer \citep[Equation 13]{bdeir2024fully}, which are briefly reviewed in \cref{app:subsubsec:review_hnns_layers}. The above backbone network is referred to as HNN, KNN, SNN, PHNN, and LNN, respectively. We collectively call them Constant Curvature Neural Networks (CCNNs). We also adopt Riemannian ResNet (RResNet) \citep{katsman2024riemannian} and HNN++ \citep{shimizu2020hyperbolic} as additional backbones.

\mypara{Implementation Details on CCNNs.}
We follow the official implementations of HGCN\footnote{\url{https://github.com/HazyResearch/hgcn}} \citep{chami2019hyperbolic}, LRBN\footnote{\url{https://github.com/CUAI/Differentiable-Frechet-Mean}} \citep{lou2020differentiating}, and HCNN\footnote{\url{https://github.com/kschwethelm/HyperbolicCV}} \citep{bdeir2024fully} to conduct experiments, where we adopt the same training settings as \citet[Section~H.1]{lou2020differentiating}. Specifically, the baseline encoder is a CCNN with two transformation layers: the first maps the input feature dimension to 128, and the second maps 128 to 128. After each transformation layer, we use a ReLU activation (in \eqnrefs{eq:act_ccnn,app:eq:act_lnn}), except for the Cora data set where activation is omitted. A BN layer, GyroBN or LRBN, is inserted after each transformation layer. The curvature is set as $|K|=1$. For the manifold-valued bias parameter in GyroBN and LRBN, we apply the exponential map $\rieexp_e(v)$ to trivialize it via a Euclidean parameter $v$. The optimization is performed with Adam \citep{kingma2014adam}, using a learning rate of $1e^{-2}$ and a weight decay of $1e^{-3}$, except for the Cora data set, where weight decay is set to 0. The Fréchet mean iterations are performed until convergence.

\begin{table}[tbp]
  \centering
  \begin{subtable}[t]{0.99\linewidth}
    \centering
    \resizebox{\linewidth}{!}{
    \begin{tabular}{c|c|ccc|ccc|ccc}
    \toprule
    \multicolumn{2}{c|}{Space} & \multicolumn{3}{c|}{Poincaré Ball $\pball{n}$} & \multicolumn{3}{c|}{Hyperboloid $\bbh{n}$} & \multicolumn{3}{c}{Beltrami--Klein $\klein{n}$} \\
    \midrule
    \multicolumn{2}{c|}{Method} & HNN   & HNN-LRBN & \cellcolor{HilightColor} HNN-GyroBN & LNN   & LNN-LRBN & \cellcolor{HilightColor} LNN-GyroBN & KNN   & KNN-LRBN & \cellcolor{HilightColor} KNN-GyroBN \\
    \midrule
    \multirow{3}[2]{*}{Disease} & Roc   & 79.21 ± 2.14 & \red{76.58 ± 2.15} & \cellcolor{HilightColor} \textbf{81.18 ± 0.93} & 87.71 ± 1.42 & \red{85.31 ± 0.95} & \cellcolor{HilightColor} \textbf{88.87 ± 0.33} & 81.31 ± 1.37 & \red{78.98 ± 1.51} & \cellcolor{HilightColor} \textbf{81.56 ± 0.70} \\
    & Fit Time & 0.027  & 0.088  & \cellcolor{HilightColor} 0.084  & 0.012  & 0.057  & \cellcolor{HilightColor} 0.057  & 0.020  & 0.089  & \cellcolor{HilightColor} 0.077  \\
    &  \#Params & 0.0180  & 0.0183  & \cellcolor{HilightColor} 0.0183  & 0.0184  & 0.0187  & \cellcolor{HilightColor} 0.0187  & 0.0180  & 0.0183  & \cellcolor{HilightColor} 0.0183  \\
    \midrule
    \multirow{3}[2]{*}{Airport} & Roc   & 94.63 ± 0.19 & \red{94.17 ± 0.40} & \cellcolor{HilightColor} \textbf{ 95.40 ± 0.17} &  93.86 ± 0.21 & \red{ 93.05 ± 1.00} & \cellcolor{HilightColor} \textbf{ 95.06 ± 0.12} & 95.00 ± 0.05 & \red{94.47 ± 0.33} & \cellcolor{HilightColor} \textbf{96.14 ± 0.05} \\
    & Fit Time & 0.054  & 0.122  & \cellcolor{HilightColor} 0.119  & 0.047  & 0.089  & \cellcolor{HilightColor} 0.092  & 0.056  & 0.136  & \cellcolor{HilightColor} 0.126  \\
    &  \#Params & 0.0182  & 0.0184  & \cellcolor{HilightColor} 0.0184  & 0.0186  & 0.0188  & \cellcolor{HilightColor} 0.0188  & 0.0182  & 0.0184  & \cellcolor{HilightColor} 0.0184  \\
    \midrule
    \multirow{3}[2]{*}{\textcolor[rgb]{ .122,  .137,  .157}{Pubmed}} & Roc   & 95.02± 0.42 & \red{93.40 ± 0.20} & \cellcolor{HilightColor} \textbf{95.83 ± 0.11} & 95.36 ± 0.10 & 95.88 ± 0.09 & \cellcolor{HilightColor} \textbf{95.89 ± 0.11} & 95.87 ± 0.11 & \red{89.83 ± 0.27} & \cellcolor{HilightColor} \textbf{96.23 ± 0.12} \\
    & Fit Time & 0.125  & 0.342  & \cellcolor{HilightColor} 0.335  & 0.111  & 0.242  & \cellcolor{HilightColor} 0.249  & 0.125  & 0.353  & \cellcolor{HilightColor} 0.337  \\
    &  \#Params & 0.0806  & 0.0809  & \cellcolor{HilightColor} 0.0809  & 0.0815  & 0.0818  & \cellcolor{HilightColor} 0.0818  & 0.0806  & 0.0809  & \cellcolor{HilightColor} 0.0809  \\
    \midrule
    \multirow{3}[2]{*}{\textcolor[rgb]{ .122,  .137,  .157}{Cora}} & Roc   & 89.96 ± 0.44 & 93.47 ± 0.49 & \cellcolor{HilightColor} \textbf{94.32 ± 0.22} & 91.84 ± 1.01 & 92.62 ± 0.17 & \cellcolor{HilightColor} \textbf{93.66 ± 0.30} & 90.03 ± 0.32 & 93.38 ± 0.12 & \cellcolor{HilightColor} \textbf{93.48 ± 0.25} \\
    & Fit Time & 0.032  & 0.091  & \cellcolor{HilightColor} 0.076  & 0.114  & 0.244  & \cellcolor{HilightColor} 0.242  & 0.038  & 0.091  & \cellcolor{HilightColor} 0.076  \\
    &  \#Params & 0.2001 & 0.2003 & \cellcolor{HilightColor} 0.2003 & 0.2019 & 0.2021 & \cellcolor{HilightColor} 0.2021 & 0.2001 & 0.2003 & \cellcolor{HilightColor} 0.2003 \\
    \bottomrule
    \end{tabular}%
    }
    \caption{Results on three hyperbolic spaces.}
    \label{tab:res:hyperbolic}
  \end{subtable}
  
  \begin{subtable}[t]{0.8\linewidth}
    \centering
    \resizebox{\linewidth}{!}{
    \begin{tabular}{c|c|ccc|ccc}
    \toprule
    \multicolumn{2}{c|}{Space} & \multicolumn{3}{c|}{Projected Hypersphere $\projhs{n}$} & \multicolumn{3}{c}{Sphere $\sphere{n}$} \\
    \midrule
    \multicolumn{2}{c|}{Method} & PHNN  & PHNN-LRBN & \cellcolor{HilightColor} PHNN-GyroBN & SNN   & SNN-LRBN & \cellcolor{HilightColor} SNN-GyroBN \\
    \midrule
    \multirow{3}[2]{*}{Disease} & Roc   & 69.70 ± 2.01 & \red{60.25 ± 1.25} & \cellcolor{HilightColor} \textbf{72.26 ± 0.61} & 54.19 ± 2.21 & \red{53.38 ± 4.07} & \cellcolor{HilightColor} \textbf{71.84 ± 0.89} \\
    & Fit Time & 0.022  & 0.066  & \cellcolor{HilightColor} 0.063  & 0.029  & 0.042  & \cellcolor{HilightColor} 0.043  \\
    &  \#Params & 0.0180  & 0.0183  & \cellcolor{HilightColor} 0.0183  & 0.0181  & 0.0183  & \cellcolor{HilightColor} 0.0183  \\
    \midrule
    \multirow{3}[2]{*}{Airport} & Roc   & 89.60 ± 0.99 & \red{87.06 ± 0.46} & \cellcolor{HilightColor} \textbf{90.44 ± 0.93} & 83.63 ± 0.77 & 86.14 ± 0.79 & \cellcolor{HilightColor} \textbf{91.12 ± 1.57} \\
    & Fit Time & 0.056  & 0.102  & \cellcolor{HilightColor} 0.095  & 0.053  & 0.070  & \cellcolor{HilightColor} 0.071  \\
    &  \#Params & 0.0182  & 0.0184  & \cellcolor{HilightColor} 0.0184  & 0.0182  & 0.0184  & \cellcolor{HilightColor} 0.0184  \\
    \midrule
    \multirow{3}[2]{*}{\textcolor[rgb]{ .122,  .137,  .157}{Pubmed}} & Roc   & 89.86 ± 0.39 & 90.06 ± 0.23 & \cellcolor{HilightColor} \textbf{92.06 ± 0.64} & 79.94 ± 1.76 & 90.10 ± 0.21 & \cellcolor{HilightColor} \textbf{93.31 ± 0.08} \\
    & Fit Time & 0.121  & 0.175  & \cellcolor{HilightColor} 0.174  & 0.120  & 0.140  & \cellcolor{HilightColor} 0.156  \\
    &  \#Params & 0.0806  & 0.0809  & \cellcolor{HilightColor} 0.0809  & 0.0806  & 0.0809  & \cellcolor{HilightColor} 0.0809  \\
    \midrule
    \multirow{3}[2]{*}{\textcolor[rgb]{ .122,  .137,  .157}{Cora}} & Roc   & 92.88 ± 0.26 & \red{92.03 ± 0.42} & \cellcolor{HilightColor} \textbf{93.26 ± 0.42} & 92.10 ± 0.40 & \red{82.01 ± 0.71} & \cellcolor{HilightColor} \textbf{93.16 ± 0.32} \\
    & Fit Time & 0.026  & 0.067  & \cellcolor{HilightColor} 0.063  & 0.025  & 0.044  & \cellcolor{HilightColor} 0.045  \\
    &  \#Params & 0.2001 & 0.2003 & \cellcolor{HilightColor} 0.2003 & 0.2001 & 0.2003 & \cellcolor{HilightColor} 0.2003 \\
    \bottomrule
    \end{tabular}%
    }
    \caption{Results on two spherical spaces.}
    \label{tab:res:sphere}
  \end{subtable}
    \caption{Comparison of GyroBN against LRBN across five constant curvature spaces. The best results are highlighted with \textbf{bold}. When LRBN degenerates the backbone network, the results are highlighted with \red{red}.}
  \label{tab:res:ccs}
\end{table}%

\subsubsection{Main Results}

We compare our GyroBN with LRBN across five constant curvature spaces under the CCNN backbone. \cref{tab:res:ccs} reports the 5-fold average testing AUC on four data sets. We highlight the following findings.
\begin{itemize}
    \item 
    \mypara{Improved Performance.}
    GyroBN consistently improves performance over the vanilla CCNNs across all data sets and geometries, whereas LRBN degrades performance in several cases (highlighted in \red{red}). The gains are especially pronounced on the sphere (SNN), where GyroBN achieves improvements of +17.65\% (Disease), +7.49\% (Airport), and +13.37\% (Pubmed). This contrast underscores the advantage of GyroBN’s theoretical guarantee of normalizing sample statistics. 
    \item 
    \mypara{Efficiency.}
    As shown in the ``Fit Time'' rows of \cref{tab:res:ccs}, GyroBN is more efficient than LRBN on the Poincaré ball, Beltrami--Klein and projected hypersphere, due to the simplicity of gyro operations. On the hyperboloid and sphere, GyroBN and LRBN exhibit comparable efficiency.
    \item 
    \mypara{Parameter Equivalence.}
    GyroBN and LRBN require the same number of parameters, only marginally more than those of the vanilla backbone. Thus, the performance gains of GyroBN cannot be attributed to parameter size, but rather to its principled normalization mechanism.
    \end{itemize}
    
\subsubsection{Ablations}

\begin{figure}[tbp]
\centering
\includegraphics[width=\linewidth,trim={0cm 1cm 0cm 1cm}]{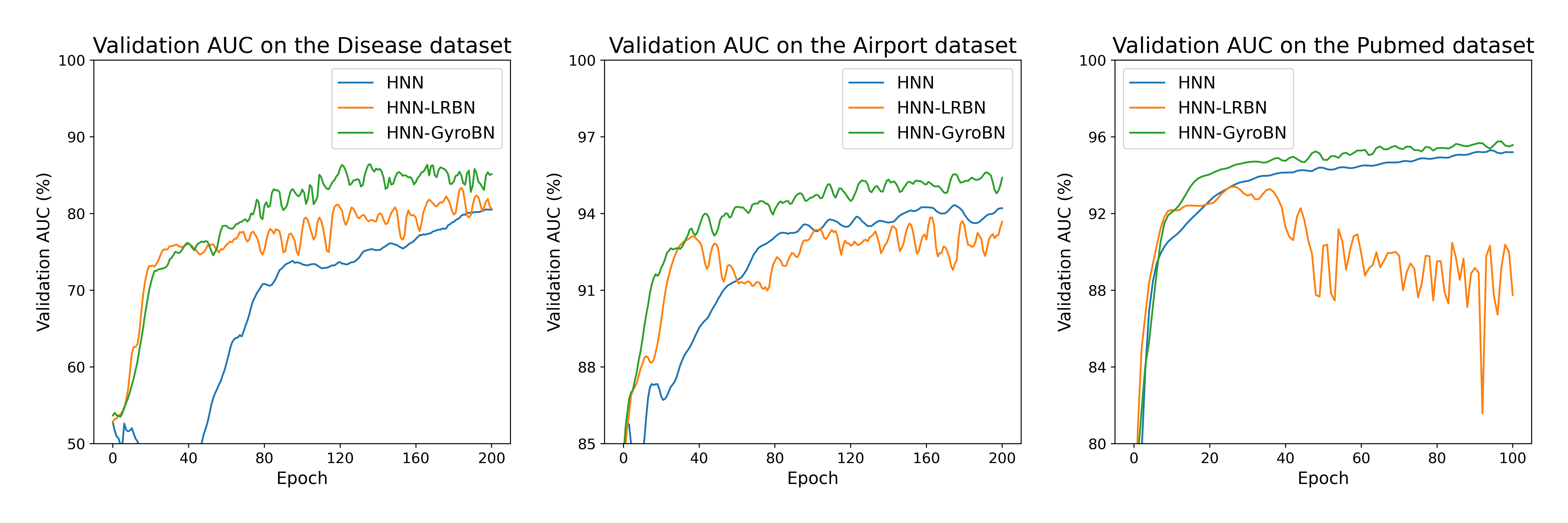}
\caption{Validation AUC of HNN with or without GyroBN or LRBN.}
\label{fig:acc_gyrobn_hyperbolic}
\end{figure}

\mypara{Training Dynamics.}
\cref{fig:acc_gyrobn_hyperbolic} presents the validation AUC curves on the Poincaré HNN. LRBN can even degrade baseline performance. In contrast, GyroBN consistently accelerates convergence and enhances overall performance.

\begin{table}[tbp]
  \centering

  \begin{subtable}[t]{0.48\linewidth}
    \centering
    \resizebox{\linewidth}{!}{
        \begin{tabular}{c|ccc}
        \toprule
        Method & Disease & Airport & Cora \\
        \midrule
        HNN++ & 80.19 ± 0.51 & 94.74 ± 0.24 & 91.06 ± 0.47 \\
        HNN++-LRBN & \red{61.18 ± 1.00} & 95.21 ± 0.65 & \red{88.22 ± 0.67} \\
        \midrule
       \rowcolor{HilightColor}  HNN++-GyroBN & \textbf{81.06 ± 0.95} & \textbf{95.77 ± 0.18} & \textbf{92.83 ± 0.32} \\
        \bottomrule
        \end{tabular}%
    }
    \caption{Main Results.}
    \label{tab:res:hnnpp}
  \end{subtable}
  \hfill
  \begin{subtable}[t]{0.48\linewidth}
    \centering
    \resizebox{\linewidth}{!}{
    \begin{tabular}{c|ccc}
    \toprule
    Weight Decay & $0$ & $1e^{-3}$ & $5e^{-4}$ \\
    \midrule
    HNN++ & \textbf{96.03 ± 0.13} & 69.26 ± 0.47 & 70.75 ± 0.33 \\
    HNN++-LRBN & \red{92.33 ± 0.32} & 92.18 ± 0.18 & 92.16 ± 0.57 \\
    \midrule
   \rowcolor{HilightColor}  HNN++-GyroBN & \red{95.48 ± 0.20} & \textbf{95.36 ± 0.18} & \textbf{95.56 ± 0.10} \\
    \bottomrule
    \end{tabular}%
    }
    \caption{Pubmed with varying weight decay.}
    \label{tab:res:hnnpp_wd}
  \end{subtable}
  \caption{LRBN vs. GyroBN on Poincaré HNN++.}
  \label{tab:res:hnn++}
\end{table}

\mypara{Results on HNN++.}  
We further validate our GyroBN on Poincaré HNN++\footnote{\url{https://github.com/mil-tokyo/hyperbolic_nn_plusplus}} \citep{shimizu2020hyperbolic}, which are briefly reviewed in \cref{app:subsubsec:review_hnns_layers}. It replaces the transformation layer in \cref{eq:tans_ccnn} with a Poincaré FC layer \citep[Section~3.2]{shimizu2020hyperbolic} for a more intrinsic transformation. Following the HNN experiments, we adopt a 2-block HNN++ as the encoder, consisting of two blocks of Poincaré FC and activation layers. Similarly, the BN layer is inserted after each Poincaré FC layer. Other settings remain the same. The 5-fold average results are reported in \cref{tab:res:hnn++}. We highlight the following findings.  
\begin{itemize}
    \item 
    \mypara{Improved Performance and Robustness.}
    While LRBN can degrade HNN++’s performance, particularly on the Disease data set, GyroBN consistently enhances it. We also observe that HNN++ could be sensitive to weight decay. For instance, on the Pubmed, setting the weight decay to $1e^{-3}$ reduces drastically the performance, from 96.03\% to 69.26\%. In contrast, HNN++ with GyroBN is robust to weight decay. 
    \item 
    \mypara{Accelerated convergence.}
    On Cora, vanilla HNN++ requires more than 400 epochs to converge, whereas HNN++-GyroBN converges in roughly 150 epochs, substantially speeding up training.  
\end{itemize}

\begin{table}[tbp]
    \centering
    \begin{subtable}[t]{0.8\linewidth}
    \centering
    \resizebox{\linewidth}{!}{
    \begin{tabular}{c|ccccc}
    \toprule
    Dim   & 8     & 16    & 32    & 64    & 128 \\
    \midrule
    RResNet & 68.32 ± 11.31 & 67.01 ± 9.58 & \textbf{71.48 ± 8.63} & 67.03 ± 2.91 & 54.19 ± 3.43 \\
    RResNet-LRBN & \red{62.96 ± 7.27} & 69.77 ± 10.27 & \red{65.72 ± 0.91} & 71.62 ± 3.04 & 63.06 ± 3.88 \\
    \midrule
    \rowcolor{HilightColor} RResNet-GyroBN & \textbf{76.18 ± 2.98} & \textbf{77.32 ± 1.59} & \red{69.87 ± 2.97} & \textbf{76.38 ± 2.15} & \textbf{66.04 ± 4.04} \\
    \bottomrule
    \end{tabular}
    }
    \caption{Results on the Disease data set.}
    \label{tab:res:rresnet:disease}
    \end{subtable}
    
    \begin{subtable}[t]{0.8\linewidth}
    \centering
    \resizebox{\linewidth}{!}{
    \begin{tabular}{c|ccccc}
    \toprule
    Dim   & 8     & 16    & 32    & 64    & 128 \\
    \midrule
    RResNet & 92.90 ± 0.45 & 91.20 ± 2.72 & \textbf{91.30 ± 0.82} & 89.70 ± 1.90 & 89.34 ± 0.66 \\
    RResNet-LRBN & \red{90.42 ± 1.43} & \red{88.56 ± 2.36} & \red{89.05 ± 1.25} & 89.65 ± 1.56 & \red{88.77 ± 2.22} \\
    \midrule
    \rowcolor{HilightColor} RResNet-GyroBN & \textbf{94.28 ± 1.51} & \textbf{91.65 ± 2.30} & \textbf{92.37 ± 1.07} & \textbf{90.41 ± 0.65} & \textbf{89.65 ± 0.32} \\
    \bottomrule
    \end{tabular}
    }
    \caption{Results on the Airport data set.}
    \label{tab:res:rresnet:airport}
    \end{subtable}
    
    \begin{subtable}[t]{0.8\linewidth}
    \centering
    \resizebox{\linewidth}{!}{
    \begin{tabular}{c|ccccc}
    \toprule
    Dim   & 8     & 16    & 32    & 64    & 128 \\
    \midrule
    RResNet &  71.77 ± 6.67 &  77.38 ± 10.82 &  77.14 ± 7.48 & 66.73 ± 11.85 & 80.75 ± 4.12 \\
    RResNet-LRBN & \red{ 61.94 ± 2.28} & \red{ 63.54 ± 3.01} & \red{62.26 ± 3.48} & \red{60.38 ± 2.97} & 87.92 ± 2.67 \\
    \midrule
    \rowcolor{HilightColor} RResNet-GyroBN & \textbf{ 84.58 ± 5.44} & \textbf{ 87.67 ± 2.53} & \textbf{ 88.08 ± 2.90} & \textbf{86.50 ± 1.59} & \textbf{89.52 ± 3.42} \\
    \bottomrule
    \end{tabular}%
    }
    \caption{Results on the Cora data set.}
    \label{tab:res:rresnet:cora}
    \end{subtable}
    \caption{LRBN vs. GyroBN on the Poincaré RResNet with different hidden dimensions.}
    \label{tab:res:rresnet}
\end{table}

\mypara{Results on RResNet.}  
As an additional backbone, we adopt the Poincaré RResNet Horo\footnote{\url{https://github.com/CUAI/Riemannian-Residual-Neural-Networks}} \citep{katsman2024riemannian}, which extends the classical ResNet \citep{he2016deep} to the Poincaré ball. For brevity, we denote it RResNet. A brief review is provided in \cref{app:subsubsec:review_hnns_layers}. The input features are first mapped by a Euclidean linear layer into a hidden dimension and then projected onto the Poincaré ball. The resulting representations are processed by RResNet, consisting of two residual blocks parameterized by horosphere-induced vector fields. A normalization layer, LRBN or our GyroBN, is inserted after RResNet, followed by a ReLU activation as defined in \cref{eq:act_ccnn}. For a complete comparison, we vary the hidden dimension from $8$ to $128$. Training is performed with a learning rate of $1e^{-2}$ and a weight decay of $1e^{-3}$, while keeping other settings identical to \citet{katsman2024riemannian}. The 5-fold average results are summarized in \cref{tab:res:rresnet}, leading to the following findings.  

\begin{enumerate}
    \item 
    \mypara{Improved Performance.}
    LRBN often degrades the RResNet baseline, particularly on the Airport and Cora data sets. In contrast, GyroBN consistently improves accuracy across most hidden dimensions, with gains of up to $+11.85\%$ on Disease (from $54.19\%$ to $66.04\%$), $+1.38\%$ on Airport (from $92.90\%$ to $94.28\%$), and $+19.77\%$ on Cora (from $66.73\%$ to $86.50\%$).  

    \item 
    \mypara{Stabilized Training.}
    On the Cora data set, both RResNet and RResNet-LRBN are highly sensitive to hidden dimensions, leading to large performance fluctuations. In contrast, RResNet-GyroBN exhibits much stabler performance, indicating that GyroBN not only improves accuracy but also enhances robustness to architectural variations.
\end{enumerate}

\subsection{Experiments on Correlation Neural Networks}
\label{subsec:exp:cor_nn}

\subsubsection{Setup}

\mypara{Data sets and Preprocessing.}
Following \citet{chen2025cornet}, we use the Radar, HDM05 and FPHA data sets, modeling each input sequence as multichannel correlation matrices. Please refer to \cref{app:subsec:dataset_preprocessing_cor,app:subsubsec:cor_inputs} for more details.

\mypara{Comparative Methods.}
Similar to \cref{subsec:exp:ccs_nn}, we compare GyroBN against LRBN.

\mypara{Backbone Networks.}
We adopt CorNet-PHCM \citep{chen2025cornet}, abbreviated as CorNet, as the backbone network. CorNet identifies each correlation matrix $C \in \cor{n}$ with a Poincaré vector and applies Poincaré layers. Specifically, $C$ is mapped to the poly-Poincaré space $\bbPPB{n-1}$ via \cref{eq:iso_cor_phc}. The resulting multi-channel Poincaré vectors are then merged into a single Poincaré representation through $\beta$-concatenation \citep[Section~3.3]{shimizu2020hyperbolic}, as recapped in \cref{app:subsubsec:review_hnns_layers}.Then, a Poincaré FC layer followed by a Poincaré \gls{MLR} layer constructs the network.

\mypara{Implementation Details.}
We follow all the implementation settings in \citet{chen2025cornet}. Since CorNet operates in the Poincaré geometry, both GyroBN and LRBN are instantiated in the Poincaré model and applied after the Poincaré FC layer. To stabilize training, we scale the learning rate of the shift parameter $s$ by factors of $0.1$ and $0.5$ for Radar and HDM05, respectively. On FPHA, we further regularize the scaling by clamping: $\min \left(\tfrac{s}{\sqrt{v^2+\epsilon}},4\right)$. For better efficiency, the number of Fréchet mean iterations is set to 2.

\subsubsection{Main Results}
\begin{table}[tbp]
  \centering
  \resizebox{0.99\linewidth}{!}{
    \begin{tabular}{c|ccc|ccc|ccc}
    \toprule
    \multirow{2}[4]{*}{Methods} & \multicolumn{3}{c|}{Radar} & \multicolumn{3}{c|}{HDM05} & \multicolumn{3}{c}{FPHA} \\
\cmidrule{2-10}          & Acc   & Fit Time &  \#Params & Acc   & Fit Time &  \#Params & Acc   & Fit Time &  \#Params \\
    \midrule
    CorNet & 96.56 ± 0.86 & 2.12  & 0.0438  & \textbf{82.26 ± 0.92} & 0.74  & 0.4091  & 90.03 ± 0.63 & 0.70  & 1.1210  \\
    \midrule
    CorNet-LRBN & \red{92.85 ± 2.46} & 2.33  & 0.0439  & \red{N/A} & 1.64  & 0.4094  & \red{81.53 ± 0.72} & 1.12 & 1.1213  \\
    \midrule
    \rowcolor{HilightColor} CorNet-GyroBN & \textbf{97.67 ± 0.36} & 2.19  & 0.0439  & \red{80.82 ± 0.86} & 1.33  & 0.4094  & \textbf{92.88 ± 0.20} & 1.07  & 1.1213  \\
    \bottomrule
    \end{tabular}%
    }  
    \caption{Comparison of CorNet with or without RBN layers.}
    \label{tab:res:cornet_gyrobn}%
\end{table}

We summarize the comparison of CorNet with or without normalization layers in \cref{tab:res:cornet_gyrobn}. Overall, GyroBN demonstrates clear benefits on Radar and FPHA with negligible parameter cost and modest efficiency trade-offs. On HDM05, however, neither GyroBN nor LRBN improves the baseline, with LRBN even diverging and failing to converge.

\subsection{Discussions}
\label{subec:exp:discussions}

\subsubsection{Ablations on Covariate Shifts}
The covariate shift is the central motivation behind the classical Euclidean BN \citep{ioffe2015batch}. Similar issues also arise in Riemannian networks, which can distort data distributions across layers, highlighting the importance of normalization. To examine this phenomenon, we conduct numerical experiments
on Grassmannian and hyperbolic networks.

\begin{table}[tbp]
    \centering
    \begin{tabular}{c|cc}
        \toprule
        & Pooling & Transformation \\
        \midrule
        $d(M _{out}, \idonb)$    & 2.90 ± 0.11 & 4.18 ± 0.04 \\
        $\Delta$ (\%) & 58.47\% ± 2.28\% & 84.22\% ± 0.72\% \\
        \bottomrule
    \end{tabular}%
    \caption{Ten-fold results for geodesic distance $d(M _{out}, \idonb)$ and shift $\Delta = \frac{d(M _{out}, \idonb)}{\sqrt{p} \nicefrac{\pi}{2}} \times 100$.}
    \label{app:tab:cov_shift_grass}%
\end{table}%

\mypara{Grassmannian.}
The transformation and pooling layers in the GyroGr baseline, which are recapped in \cref{app:eq:gyrotranslation,app:eq:grass_pooling}, can be expressed as
\begin{align*}
    f _{\mathrm{trans}}: \grasonb{p,n} \rightarrow \grasonb{p,n},\\
    f _{\mathrm{pooling}}: \grasonb{p,n} \rightarrow \grasonb{p,\nicefrac{n}{2}}.
\end{align*}
For simplicity, we assume $n$ is even for the pooling layer. Since the ONB Grassmannian $\grasonb{p,n}$ is a quotient manifold and pooling changes dimensions, we use the geodesic distance between the batch mean and the identity element $\idonb = (I _p, \mathbf{0})^\top \in \mathbb{R}^{n \times p}$ as a consistent measure. Note that the geodesic distance on $\grasonb{p,n}$ is bounded by $\sqrt{p}\frac{\pi}{2}$ \citep[Theorem 8]{wong1967differential}. We randomly generate 30 Grassmannian matrices of size $100 \times 10$ with an initial batch mean as the identity element $\idonb$. We denote the resulting batch mean after transformation or pooling as $M _{out}$. If the distribution were preserved, the geodesic distance $d(M _{out}, \idonb)$ (or $d(M _{out}, I _{p, n/2})$ for pooling) would be zero. \cref{app:tab:cov_shift_grass} shows that $M _{out}$ significantly deviates from $\idonb$, indicating the covariate shift.

\mypara{Hyperbolic Spaces.}
We examine the covariate shift of the transformation layer in the Poincaré HNN (\cref{eq:tans_ccnn}). We focus on the canonical Poincaré ball (curvature $K=-1$). We randomly generate 30 5-dimensional Poincaré vectors. The batch mean vectors of input and output in the transformation layer are
\begin{align*}
    M_{in} = [-0.0055,  0.0503, -0.0913, -0.0493,  0.0652]^\top,\\
    M_{out} = [-0.0969, -0.0443,  0.0385,  0.0936, -0.0770]^\top.
\end{align*}
The geodesic distance between $M_{in}$ and $M_{out}$ is $0.55$, indicating the covariate shift in HNN.

\subsubsection{Ablations on Condition Numbers}

In the GyroGr network, the transformation layer outputs $n \times p$ Grassmannian matrices whose singular values are all $1$, yielding trivial condition numbers. Similarly, the transformation layers in constant curvature networks output vectors, which also lead to trivial condition numbers. For CorNet, where each correlation is identified as a Poincaré vector, the condition numbers are likewise trivial. Hence, we focus our analysis on the condition numbers of the weight matrices in the hidden transformation layer and the overall network Jacobian. Both analyses are conducted on the Grassmannian GyroGr under the 1-block architecture. To cover different scales, we report results on small-scale HDM05 and large-scale NTU120.

\begin{table}[tbp]
    \centering
    \begin{subtable}[t]{0.48\linewidth}
    \centering
    \resizebox{\linewidth}{!}{
    \begin{tabular}{ccccc}
    \toprule
    Data set & BN & Mean  & Min   & Max \\
    \midrule
    \multirow{4}[2]{*}{HDM05} & None  & 3.81 ± 0.23 & 3.46  & 4.17 \\
          & ManifoldNorm & 1.97 ± 0.15 & 1.83  & 2.30 \\
          & LRBN & 3.49 ± 0.46 & 2.96  & 4.43 \\
          & \cellcolor{HilightColor} GyroBN & \cellcolor{HilightColor} 2.37 ± 0.18 & \cellcolor{HilightColor} 2.12  & \cellcolor{HilightColor} 2.67 \\
    \midrule
    \multirow{4}[2]{*}{NTU120} & None  & 3.35 ± 0.28 & 2.97  & 3.72 \\
          & ManifoldNorm & 1.91 ± 0.10 & 1.80   & 2.14 \\
          & LRBN & 2.22 ± 0.11 & 2.00     & 2.36 \\
          & \cellcolor{HilightColor} GyroBN & \cellcolor{HilightColor} 2.16 ± 0.11 & \cellcolor{HilightColor} 2.00     & \cellcolor{HilightColor} 2.33 \\
    \bottomrule
    \end{tabular}%
    }
    \caption{Weight matrices in the transformation layer. }
    \label{tab:con_num_weight}%
    \end{subtable}
    \hfill
    \begin{subtable}[t]{0.48\linewidth}
    \centering
    \resizebox{\linewidth}{!}{
    \begin{tabular}{ccccc}
    \toprule
    Data set & BN & Mean  & Min   & Max \\
    \midrule
    \multirow{4}[2]{*}{HDM05} & None  & 61.05 ± 4.15 & 52.19 & 74.29 \\
          & ManifoldNorm & 67.38 ± 6.82 & 51.15 & 88.81 \\
          & LRBN & 64.16 ± 6.44 & 50.54 & 80.07 \\
          & \cellcolor{HilightColor} GyroBN & \cellcolor{HilightColor} 59.85 ± 4.33 & \cellcolor{HilightColor} 48.46 & \cellcolor{HilightColor} 70.47 \\
    \midrule
    \multirow{4}[2]{*}{NTU120} & None  & 122.24 ± 68.45 & 61.8  & 399.28 \\
          & ManifoldNorm & 116.75 ± 65.17 & 75.46 & 443.01 \\
          & LRBN & 108.75 ± 52.27 & 63.19 & 349.25 \\
          & \cellcolor{HilightColor} GyroBN & \cellcolor{HilightColor} 97.01 ± 41.88 & \cellcolor{HilightColor} 61.19 & \cellcolor{HilightColor} 262.55 \\
    \bottomrule
    \end{tabular}%
    }
    \caption{Network Jacobian (output w.r.t. input).}
    \label{tab:con_num_network}%
    \end{subtable}
    \caption{Condition numbers under different normalization methods on the GyroGr baseline.}
    \label{tab:res:condition_number}
\end{table}

\mypara{Transformation Layers.}
Since the weight matrices have 8 channels, we report the mean, standard deviation (std), minimum, and maximum values of their condition numbers. The dimensions of the 8-channel transformation matrices are $83 \times 10$ and $140 \times 10$ on HDM05 and NTU120, respectively. As shown in \cref{tab:con_num_weight}, GyroBN consistently reduces the condition numbers of the weight matrices across both data sets. Interestingly, although ManifoldNorm achieves the smallest condition numbers, its performance can be worse than the vanilla GyroGr baseline. This suggests that excessively reducing the condition number may constrain the model’s expressive capacity.

\mypara{Network Jacobian.}
We randomly select 100 samples and feed them into the trained models to compute Jacobian statistics. The dimensions of the Jacobian matrices are $117 \times 930$ on HDM05 and $11 \times 9600$ on NTU120. As reported in \cref{tab:con_num_network}, GyroBN achieves lower condition numbers than competing methods on both data sets, indicating that it stabilizes network training. Moreover, the reduction in maximum condition numbers, especially on NTU120, highlights GyroBN’s ability to avoid extreme outliers in conditioning, further underscoring its advantage. Notably, on HDM05, both ManifoldNorm and LRBN actually increase the Jacobian condition numbers.

\subsubsection{Ablations on the Number of Fréchet Mean Iterations}

\begin{figure}[tbp]
\centering
\includegraphics[width=\linewidth,trim={0cm 1cm 0cm 1cm}]{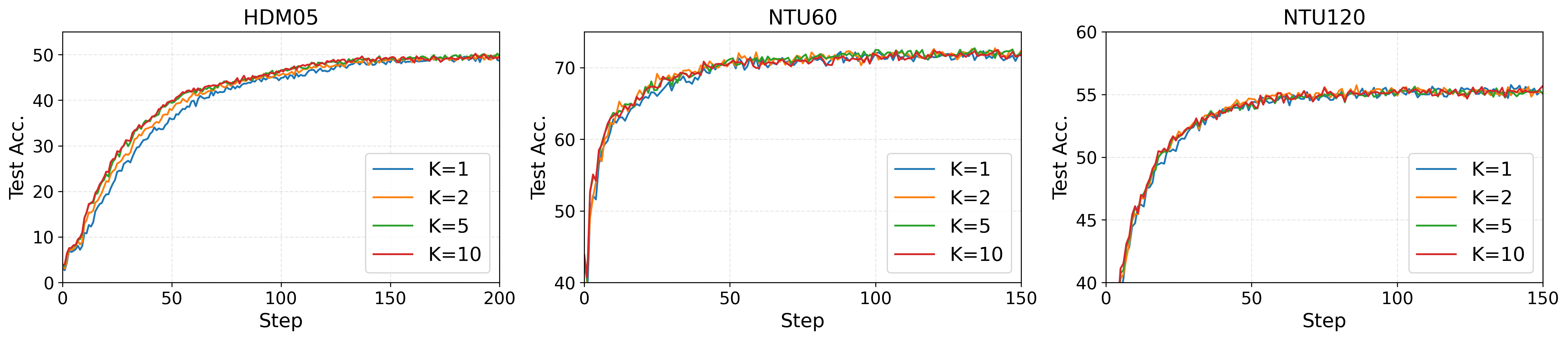}
\caption{Testing accuracy curve of the Grassmannian GyroGr-GyroBN under different Karcher steps.}
\label{fig:ab:mean_iteration_grass}
\end{figure}

\begin{table}[tbp]
  \centering
  \resizebox{0.99\linewidth}{!}{
    \begin{tabular}{c|c|c|c|c|c|c}
    \toprule
    Geometry & Model & Iteration & Disease & Airport & \textcolor[rgb]{ .122,  .137,  .157}{Pubmed} & \textcolor[rgb]{ .122,  .137,  .157}{Cora} \\
    \midrule
    \multirow{15}[11]{*}{Hyperbolic} & \multirow{5}[4]{*}{Poincaré Ball} & 1     & 67.47 ± 2.76 & 91.74 ± 2.52 & 94.93 ± 0.39 & 92.39 ± 0.27 \\
          &       & 2     & 74.76 ± 3.65 & \textbf{96.14 ± 0.10} & 95.93 ± 0.17 & 94.01 ± 0.23 \\
          &       & 5     & 80.92 ± 1.19 & 96.06 ± 0.05 & \textbf{96.08 ± 0.08} & 94.03 ± 0.21 \\
          &       & 10    & \textbf{81.44 ± 1.96} & 96.13 ± 0.12 & 96.05 ± 0.24 & 94.03 ± 0.21 \\
\cmidrule{3-7}          &       & \cellcolor{HilightColor} $\infty$ & \cellcolor{HilightColor} 81.18 ± 0.93 & \cellcolor{HilightColor}  95.40 ± 0.17 & \cellcolor{HilightColor} 95.83 ± 0.11 & \cellcolor{HilightColor} \textbf{94.32 ± 0.22} \\
\cmidrule{2-7}          & \multirow{5}[4]{*}{Hyperboloid} & 1     & 89.29 ± 0.52 & 94.50 ± 0.16 & \textbf{96.03 ± 0.03} & 93.72 ± 0.08 \\
          &       & 2     & 88.96 ± 0.58 & 94.58 ± 0.21 & 96.02 ± 0.20 & 93.78 ± 0.25 \\
          &       & 5     & 87.62 ± 1.49 & 94.60 ± 0.15 & 95.91 ± 0.11 & \textbf{93.80 ± 0.23} \\
          &       & 10    & \textbf{89.35 ± 0.57} & 94.44 ± 0.17 & 96.01 ± 0.10 & \textbf{93.80 ± 0.23} \\
\cmidrule{3-7}          &       & \cellcolor{HilightColor} $\infty$ & \cellcolor{HilightColor} 88.87 ± 0.33 & \cellcolor{HilightColor} \textbf{ 95.06 ± 0.12} & \cellcolor{HilightColor} 95.89 ± 0.11 & \cellcolor{HilightColor} 93.66 ± 0.30 \\
\cmidrule{2-7}          & \multirow{5}[3]{*}{Beltrami--Klein} & 1     & 81.56 ± 1.71 & 96.06 ± 0.08 & 95.71 ± 0.19 & 93.27 ± 0.18 \\
          &       & 2     & \textbf{81.62 ± 1.33} & 96.16 ± 0.08 & 95.83 ± 0.23 & 93.42 ± 0.13 \\
          &       & 5     & 81.38 ± 0.80 & \textbf{96.17 ± 0.17} & 95.88 ± 0.08 & 93.40 ± 0.15 \\
          &       & 10    & 80.84 ± 1.08 & 96.15 ± 0.05 & 95.81 ± 0.19 & 93.38 ± 0.12 \\
\cmidrule{3-7}          &       & \cellcolor{HilightColor} $\infty$ & \cellcolor{HilightColor} 81.56 ± 0.70 & \cellcolor{HilightColor} 96.14 ± 0.05 & \cellcolor{HilightColor} \textbf{96.23 ± 0.12} & \cellcolor{HilightColor} \textbf{93.48 ± 0.25} \\
\midrule
    \multirow{10}[5]{*}{Spherical} & \multirow{5}[3]{*}{Projected Hypersphere} & 1     & \textbf{74.07 ± 0.83} & \textbf{91.32 ± 0.82} & 91.08 ± 0.26 & 92.30 ± 0.20 \\
          &       & 2     & 71.64 ± 1.24 & 91.08 ± 0.43 & 91.34 ± 0.23 & 92.95 ± 0.46 \\
          &       & 5     & 72.01 ± 1.20 & 90.90 ± 0.29 & \textbf{92.30 ± 0.24} & 92.76 ± 0.62 \\
          &       & 10    & 72.25 ± 1.97 & 90.44 ± 0.93 & 91.89 ± 0.49 & 92.92 ± 0.16 \\
\cmidrule{3-7}          &       & \cellcolor{HilightColor} $\infty$ & \cellcolor{HilightColor} 72.26 ± 0.61 & \cellcolor{HilightColor} 90.44 ± 0.93 & \cellcolor{HilightColor} 92.06 ± 0.64 & \cellcolor{HilightColor} \textbf{93.26 ± 0.42} \\
\cmidrule{2-7}          & \multirow{5}[4]{*}{Projected Hypersphere} & 1     & 70.49 ± 3.18 & \textbf{91.57 ± 1.00} & 93.24 ± 0.31 & 92.43 ± 0.90 \\
          &       & 2     & \textbf{72.96 ± 2.00} & 90.72 ± 1.90 & 93.24 ± 0.22 & 93.09 ± 0.53 \\
          &       & 5     & 72.82 ± 1.74 & 89.61 ± 1.81 & \textbf{93.46 ± 0.10} & 92.85 ± 0.41 \\
          &       & 10    & 71.73 ± 0.91 & 91.12 ± 1.57 & 93.31 ± 0.08 & 92.64 ± 0.35 \\
\cmidrule{3-7}          &       & \cellcolor{HilightColor} $\infty$ & \cellcolor{HilightColor} 71.84 ± 0.89 & \cellcolor{HilightColor} 91.12 ± 1.57 & \cellcolor{HilightColor} 93.31 ± 0.08 & \cellcolor{HilightColor} \textbf{93.16 ± 0.32} \\
    \bottomrule
    \end{tabular}%
    }
    \caption{Ablation study on the number of Fréchet mean iterations for different constant curvature spaces. The symbol $\infty$ indicates that iterations are performed until convergence, which is the setting used in our main experiments.}
    \label{tab:ab:mean_iteration_ccs}%
\end{table}%

\begin{table}[tbp]
  \centering
    \resizebox{0.5\linewidth}{!}{
    \begin{tabular}{c|c|ccc}
    \toprule
    Geometry & Iteration & HDM05 & NTU60 & NTU120 \\
    \cmidrule{2-5}
    \multirow{4}[2]{*}{Grassmannian} & \cellcolor{HilightColor}  1 & \cellcolor{HilightColor}  3.04  & \cellcolor{HilightColor}  35.85  & \cellcolor{HilightColor}  67.37  \\
    \midrule
          & 2     & 3.70  & 40.07 & 69.23  \\
          & 5     & 5.51  & 54.49 & 93.03  \\
          & 10    & 8.64  & 74.67 & 130.77  \\
    \bottomrule
    \end{tabular}%
    }
    \caption{Efficiency (s/epoch) of Grassmannian GyroBN under different Fréchet mean iterations. Entries in \colorbox{HilightColor}{\strut\hspace{1em}} are the setting used in our main experiments.}
    \label{tab:ab:efficiency_iteration_grass}%
\end{table}
We ablate the number of iterations used by the Fréchet mean solver in GyroBN across the Grassmannian and five constant curvature spaces. \cref{fig:ab:mean_iteration_grass} shows that on the Grassmannian, a single iteration is sufficient. Similarly, \cref{tab:ab:mean_iteration_ccs} indicates that GyroBN over constant curvature spaces is generally saturated with two iterations, except for the Poincaré ball on the Disease data set, where ten iterations yield further improvement. Meanwhile, \cref{tab:ab:efficiency_iteration_grass} highlights that reducing iterations substantially improves the efficiency of GyroBN.

\section{Conclusion}
\label{sec:conclusion}
We have introduced the pseudo-reductive gyrogroup, a novel gyro-structure on manifolds that establishes the prerequisites for principled normalization over gyrospaces. Building on this foundation, we proposed GyroBN, a general framework for batch normalization on gyrogroups that enables the normalization of non-Euclidean statistics. Our analysis reveals that several existing RBN methods, including LieBN on Lie groups and AIM-based SPDBNs, arise as special cases of GyroBN. We further instantiated GyroBN on the Grassmannian, five constant curvature spaces, and the correlation manifold, and demonstrated its effectiveness through extensive experiments. These contributions position GyroBN as a unified normalization paradigm for manifold-valued data. A natural direction for future work is to broaden this paradigm beyond gyrogroups, extending normalization to manifolds that do not admit gyro-structures.

\acks{
This work was partly supported by the MUR PNRR project FAIR (PE00000013) funded by the NextGenerationEU, and the EU Horizon project ELIAS (No. 101120237). The authors also gratefully acknowledge the CINECA award under the ISCRA initiative for the availability of HPC resources support.
}

\clearpage
\appendix
\startcontents[appendices]
\printcontents[appendices]{l}{1}{\section*{Appendix Contents}}
\newpage

\numberwithin{equation}{section}
\counterwithin{figure}{section}
\counterwithin{table}{section}

\renewcommand{\thetheorem}{\thesection.\arabic{theorem}}
\renewcommand{\thelemma}{\thesection.\arabic{lemma}}
\renewcommand{\theproposition}{\thesection.\arabic{proposition}}
\renewcommand{\thecorollary}{\thesection.\arabic{corollary}}
\renewcommand{\thedefinition}{\thesection.\arabic{definition}}
\renewcommand{\theassumption}{\thesection.\arabic{assumption}}
\renewcommand{\theremark}{\thesection.\arabic{remark}}

\setcounter{theorem}{0} 

\printglossary[type=acronym, title={List of Acronyms}]

\section{Notations} \label{app:notations}

\cref{tab:sum_notaitons} summarizes the key notations in the main paper.

\begin{table}[tbp]
    \centering
    \resizebox{0.85\linewidth}{!}{
    \begin{tabular}{ll}
    \toprule
    \rowcolor{RowGray} \multicolumn{2}{c}{General notations} \\
    \midrule
    $\struct{\calM,g}$ & Riemannian manifold with metric $g$ \\
    $T_x\calM$ & Tangent space at $x\in\calM$ \\
    $g_x(\cdot,\cdot)$,\ $\inner{\cdot}{\cdot}_x$ & Riemannian metric at $x$; $\norm{v}_x=\sqrt{g_x(v,v)}$ \\
    $\dist(\cdot,\cdot)$ & Geodesic distance on $\calM$ \\
    $\rieexp_x,\ \rielog_x$ & Riemannian exponential / logarithm at $x$ \\
    $f_{*,x}$ & Differential of $f$ at $x$ \\
    $\pt{x}{y}(\cdot)$ & Parallel transport along the geodesic connecting $x$ and $y$ \\
    $\fm$,\ $\wfm$ & (Weighted) Fréchet mean under $\dist$ \\
    $(G,\oplus,\odot)$ & Gyrovector space: $\oplus$ gyroaddition, $\odot$ scalar gyromultiplication \\
    $e,\ \ominus x,\ \gyr[x,y]$ & Gyro identity, gyroinverse, gyration \\
    $\gyrinner{\cdot}{\cdot}$, $\gyrnorm{\cdot}$, $\gyrdist(x,y)$ & Gyro inner product, gyronorm, gyrodistance \\
    \midrule
    $\bbR{n}$,\ $\bbR{n\times n}$ & Euclidean vector / matrix spaces \\
    $\inner{\cdot}{\cdot}$ & Standard (resp. Frobenius) inner product for vectors (resp. matrices) \\
    $\norm{\cdot}$ &  Norm induced by $\inner{\cdot}{\cdot}$ \\
    $\Rzero$ & Zero vector or matrix \\
    \midrule
    \rowcolor{RowGray} \multicolumn{2}{c}{Matrix manifolds (SPD, Grassmannian, Correlation)} \\
    \midrule
    $\sym{n}$,\ $\spd{n}$ & Symmetric matrices, SPD matrices \\
    $I_n$ & $n \times n$ identity matrix \\
    $\mexp(\cdot)$,\ $\mlog(\cdot)$ & Matrix exponential / logarithm \\
    $\chol(\cdot)$,\ $\dlog(\cdot)$ & Cholesky decomposition; diagonal element-wise log \\
    $\clog$ & $\clog=\dlog\circ\chol$  \\
    $\oplusGyrAI,\ \oplusLieLE,\ \oplusLieLC$ & SPD gyroaddition induced by AIM/LEM/LCM \\
    \midrule
    $\grasonb{p,n},\ \graspp{p,n}$ & Grassmannian (ONB / projector perspectives) \\
    $\idonb,\ \idpp$ & Identity elements in ONB / projector perspectives \\
    $\pi:\grasonb{p,n}\!\to\!\graspp{p,n}$ & Isometry $U\mapsto UU^\top$ \\
    $\oplusGyrONB,\ \ominusGyrONB,\ \odotGyrONB$ & Grassmannian gyroaddition / inverse / scalar product (ONB) \\
    $\overline{(\cdot)}$ & PP logarithm at $\idpp$: $\rielog_{\idpp}(\cdot)$ \\
    \midrule
    $\cor{n}$ & Full-rank correlation manifold \\
    $\covtocor(\cdot)$ & $\mathrm{Cor}(\Sigma)=D(\Sigma)^{-1/2}\,\Sigma\,D(\Sigma)^{-1/2}$ \\
    \midrule
    \rowcolor{RowGray} \multicolumn{2}{c}{Constant curvature spaces} \\
    \midrule
    $\stereo{n}$ & $K$-stereographic model (unifies $\pball{n}$ for $K<0$, $\bbR{n}$ for $K=0$, $\projhs{n}$ for $K>0$) \\
    $\pball{n},\ \projhs{n}$ & Poincaré ball ($K<0$) and projected hypersphere ($K>0$) \\
    $\stoplus,\ \stominus,\ \stodot$ & Stereographic gyroaddition / inverse / scalar product \\
    $\tank,\ \sink,\ \cosk$ & Curvature-aware $\tan_K,\ \sin_K,\ \cos_K$ \\
    $\Moplus,\ \Mominus,\ \Modot$ & Möbius operations on $\pball{n}$ \\
    \midrule
    $\calMK{n}$ & $K$-radius model (unifies $\bbh{n}$ for $K<0$, $\bbR{n}$ for $K=0$, $\sphere{n}$ for $K>0$) \\
    $\bbh{n},\ \sphere{n}$ & Hyperboloid ($K<0$) and sphere ($K>0$) \\
    $\Kinner{\cdot}{\cdot}$ & Euclidean inner product for $K>0$ and Lorentz inner product for $K<0$ \\
    $\Knorm{\cdot}$ & $\Knorm{x} = \sqrt{\Kinner{x}{x}}$ \\
    $\MKoplus,\ \MKominus,\ \MKodot$ & Radius gyroaddition / inverse / scalar product \\
    $\MKzero$ & Origin of $\calMK{n}$ \\
    $\isoSTMK{n}$, $\isoSTMK{n}$ & Isometries between radius and stereographic models \\
    $\lambda^K_x$ & $\lambda^K_x =\frac{2}{\left(1+K\|x\|^2\right)}$ \\
    \midrule
    $\klein{n}$ & Beltrami--Klein model \\
    $\Eoplus,\ \Eominus,\ \Eodot$ & Einstein addition / inverse / scalar product \\
    $\pi_{\klein{n} \to \pball{n}}$, $\pi_{\pball{n} \to \klein{n} }$ & Isometries between the Poincaré ball and Beltrami--Klein models\\
    $\gamma^K_x$ & $\gamma^K_x =\frac{1}{\sqrt{1+K\|x\|^2}}$ \\
    \midrule
    $\hs{n}$, $\pi _{\hs{n} \rightarrow \unitpball{n}}$ & Open hemisphere and its isometry to the unit Poincaré ball \\
    $\bbPPB{n-1} = \prod_{i=1}^{n-1}\unitpball{i}$ & Product of unit Poincaré balls \\
    $\pi_{\hs{i} \rightarrow \unitpball{i}}$ & Isometry from \\
    $\Phi$ & Diffeomorphism from $\cor{n}$ to $\bbPPB{n-1}$ \\
    \bottomrule
    \end{tabular}
    }
    \caption{Summary of key notations.}
    \label{tab:sum_notaitons}
\end{table}

\section{Existence and Uniqueness of the Weighted Fréchet Mean}
\label{app:sec:exist_unique_wfm}

Let $(\calM, g)$ be an orientable complete Riemannian manifold equipped with a Riemannian metric $g$. The induced distance is denoted as $\dist(\cdot,\cdot)$. We denote the supremum of the sectional curvatures of $\calM$ by $\Delta$. We recover the theorem on the existence and uniqueness of the Weighted Fréchet Mean (WFM) \citep{afsari2011riemannian}. We acknowledge that \citet[Appendix A]{chakraborty2020manifoldnet} has also provided a summary of the following discussions.

\begin{definition}[Geodesic Ball \citep{do1992riemannian}]
Let $x \in \calM$ and $r > 0$. Then $B_r(x) = \{y \in \calM \mid \dist(x, y) < r\}$ is the open geodesic ball at $x$ of radius $r$.
\end{definition}

\begin{definition}[Injectivity Radius \citep{manton2004globally}]
The local injectivity radius at $x \in \calM$, $r_{\text{inj}}(x)$, is the largest radius $r$ for which $\rieexp_P : T_P \calM \supset B_r(\Rzero) \to \calM$ is a diffeomorphism onto its image. The injectivity radius of $\calM$ is defined as $r_{\text{inj}}(\calM) = \inf_{x \in \calM} \{r_{\text{inj}}(x)\}$.
\end{definition}

Within the local injectivity radius, the exponential map is invertible and we call the inverse map as Riemannian logarithmic map, $\rielog _{x} : B_{r_{\text{inj}}(x)} (x) \rightarrow B _{r_{\text{inj}}(x)}(\Rzero) \subset T _P \calM$.

\begin{definition}[Strong Convexity \citep{chavel1995riemannian}]
A subset $U \subset \calM$ is strongly convex if for all $x, y \in U$, there exists a unique length-minimizing geodesic segment between $x$ and $y$, and the geodesic segment lies entirely in $U$.
\end{definition}

\begin{definition}[Convexity Radius \citep{groisser2004newton}]
The local convexity radius at $x \in \calM$, $r_{\text{cvx}}(x)$, is defined as
\begin{equation}
    r_{\text{cvx}}(x) = \sup \{r \le r_{\text{inj}}(x) \mid B_r(x) \text{ is strongly convex}\}.
\end{equation}
The convexity radius of $\calM$ is defined as $r_{\text{cvx}}(\calM) = \inf_{x \in \calM} \{r_{\text{cvx}}(x)\}$.
\end{definition}

\begin{theorem}[Existence and Uniqueness of WFM \citep{afsari2011riemannian}]
The WFM exists and is unique inside a geodesic ball of radius $r_{\text{cvx}}(\calM)$.
\end{theorem}

In the main paper, we always assume the involved $\rieexp$, $\rielog$, and WFM are well-defined.

\section{Grassmannian GyroBN under the projector perspective}
\label{app:sec:gyro_grass_pp}

Given two isometric manifolds $\{ \calM_1, g^1 \}$ and $\{ \calM_2, g^2 \}$, the induced gyro-structures in \cref{eq:gyro_addtion}-\cref{eq:gyro_distance} have the following relations. 
We first present a useful lemma.

\begin{lemma} \label{lem:gyrogroups_isometry}
     Given manifolds $\{ \calM_1, g^1 \}$ and $\{ \calM_2, g^2 \}$ and a Riemannian isometry $f:\calM_1 \rightarrow \calM_2$, we have the following:
     \begin{enumerate}
         \item
         The groupoid $\{\calM_1, \oplus_1 \}$ induced by $g^1$ is pseudo-reductive (left-invariant), iff the groupoid $\{ \calM_2, \oplus_2 \}$ induced by $g^2$ is pseudo-reductive (left-invariant);
         \item
         Any gyration in $\{ \calM_1, g^1 \}$ preserves gyronorm iff any gyration in $\{ \calM_2, g^2 \}$ preserves gyronorm;
         \item 
         $f$ preserves the gyrodistance.
     \end{enumerate} 
\end{lemma}
\begin{proof}
    First, we review some facts about gyrogroups under the Riemannian isometry.
    As shown by \citet[Theorem 2.5]{nguyen2023building}, $\{\calM_1, \oplus_1 \}$ satisfies (G1-G3) in \cref{def:gyrogroups} iff $\{\calM_2, \oplus_2 \}$ satisfies (G1-G3).
    Besides, $f^{-1}$ is an isomorphism satisfying
    \begin{align}
        \label{app:eq:gyro_addition_isometry}
        f^{-1}(P \oplus_2  Q) &= f^{-1}(P) \oplus_1 f^{-1}(Q), \\
        \label{app:eq:gyro_scalar_product_isometry}
        f^{-1}(t \odot_2  P) &= t \oplus_1 f^{-1}(P), \\
        \label{app:eq:gyroinverse_isometries}
        f^{-1}(\ominus_2  P) &= \ominus_1 f^{-1}(P), \\
        \label{app:eq:gyration_isometries}
        \gyr_2 [P, Q] (R) &= f \left(\gyr_1[ f^{-1} (P), f^{-1}(Q)] (f^{-1}(R)) \right).
    \end{align}
    where $P,Q,R \in \calM_2$ are arbitrary points, $t \in \bbRscalar$ is a real scalar, $\ominus_i$, $\gyr_i$ are the gyroinverses and gyrations on $\calM_i$ for $i=1,2$.
    $f$ is also an isomorphism with similar properties.
    
    We only need to prove one direction for the iff condition for the pseudo-reduction or norm invariance. We focus on $\Rightarrow$ and follow the above notations in the following. 

    \textbf{Pseudo-reduction: }
    \begin{equation} 
        \begin{aligned}
            \gyr_2[\ominus P, P] 
            &= f \circ \gyr_1[ f^{-1} (\ominus_2 P), f^{-1} (P)] \circ f^{-1} \text{ (\cref{app:eq:gyration_isometries})}\\
            &= f \circ \gyr_1[ \ominus_1 f^{-1} (P), f^{-1} (P)] \circ f^{-1} \text{ (\cref{app:eq:gyroinverse_isometries})}\\
            &= \id.
        \end{aligned}
    \end{equation}

    \textbf{Gyronorm invariance under gyrations: }
    For simplicity, we denote the gyronorm, identity element, and Riemannian logarithm on $\calM_i$ as $\norm{}_i$ $E_i$, and $\rielog^i$, respectively.
    First, the following demonstrates that the Riemannian isometry $f$ preserves gyronorm:
    \begin{equation}
        \label{app:eq:f_preserve_gyronorm}
        \begin{aligned}
            \norm{P}_2
            &= \norm{\rielog^2_{E_2} \left( P \right)}_{E_2}\\
            &\stackrel{(1)}{=} \norm{ \rielog^1_{f^{-1}(E_2)} \left( f^{-1} (P) \right)}_{f^{-1} (E_2)}\\
            &\stackrel{(2)}{=} \norm{ \rielog_{E_1} \left( f^{-1} (P) \right)}_{E_1}\\
            &= \norm{f^{-1} (P)}_1
        \end{aligned}
    \end{equation}
    \begin{enumerate}[label=(\arabic*)]
        \item 
        As $f: \calM_1 \rightarrow \calM_2$ is a Riemannian isometry, we have the following equations:
        \begin{align}
            \rielog_P^2 Q &= f_{*,f^{-1}(P)} \left( \rielog^1_{f^{-1}(P)} \left( f^{-1}(Q) \right)  \right), \forall P,Q \in \calM_2,\\
            g^2_P(V,W) &= g^1_{f^{-1}(P)}((f^{-1})_{*,P}(V),(f^{-1})_{*,P}(W)), \forall V,W \in T_P\calM_2,
        \end{align}
        where $(\cdot)_{*}$ is the differential map;
        \item 
        $E_2=f(E_1)$.
    \end{enumerate}

    Then we have the following:
    \begin{equation}
        \begin{aligned}
            \norm{\gyr_2[P,Q] (R)}_2
            &\stackrel{(1)}{=} \norm{f \left(\gyr_1[ f^{-1} (P), f^{-1}(Q)] (f^{-1}(R)) \right)}_2\\
            &\stackrel{(2)}{=} \norm{ \gyr_1[ f^{-1} (P), f^{-1}(Q)] (f^{-1}(R)) }_1\\
            &\stackrel{(3)}{=} \norm{ f^{-1}(R) }_1\\
            &\stackrel{(4)}{=} \norm{ R }_2
        \end{aligned}
    \end{equation}
    The above derivation comes from the following.
    \begin{enumerate}[label=(\arabic*)]
        \item 
        \cref{app:eq:gyration_isometries};
        \item 
        \cref{app:eq:f_preserve_gyronorm};
        \item 
        Any gyration on $\calM_1$ can preserve the gyronorm;
        \item 
        \cref{app:eq:f_preserve_gyronorm}.
    \end{enumerate}

    \textbf{Invariance of gyrodistance under $f$: }
    Denoting the gyrodistance on $\calM_i$ as $\dist_i$, for any $U,V \in \calM_1$, we have the following:
    \begin{equation}
        \begin{aligned}
            \dist_{1}(U,V) 
            &= \norm{\ominus_1 U \oplus_1 V}_1 \\
            &\stackrel{(1)}{=} \norm{f (\ominus_1 P \oplus_1 Q) }_2 \\
            &\stackrel{(2)}{=} \norm{ \ominus_1  f (P) \oplus_2 f(Q) }_2 \\
            &= \dist_2(f (P), f(Q) )_2 \\
        \end{aligned}
    \end{equation}
    The above derivation comes from the following.
    \begin{enumerate}[label=(\arabic*)]
        \item 
        $f$ preserves gyronorm;
        \item
        $f$ is an isomorphism.
    \end{enumerate}
\end{proof}

Given a batch of activations $\{P_{1\cdots N}\}$ on a gyrogroup $\{\calM, \oplus \}$, we denote the GyroBN as 
\begin{equation}
    \gyrobn(\{ P_i \};B,s,\epsilon,\eta),
\end{equation}
where $B \in \calM$ and $s$ are biasing and scaling parameters, $\epsilon$ is a small positive value, and $\eta$ is the momentum.

\begin{theorem} \label{thm:liebn_pullback}
    Given manifolds $\{ \calM_1, g^1 \}$ and $\{ \calM_2, g^2 \}$ and a Riemannian isometry $f:\calM_1 \rightarrow \calM_2$, for a batch of activation $\{P_{1 \ldots N}\}$ in $\calM_1$, $\gyrobn_1(P_i;B,s,\epsilon,\gamma)$ in $\calM_1$ can be calculated as
    \begin{equation}
        \gyrobn_1(P_i;B,s,\epsilon,\gamma) = f^{-1} \left( \gyrobn_2(f(P_i);f(B),s,\epsilon,\gamma) \right),
    \end{equation} 
    where $\gyrobn_2$ is the GyroBN in  $\calM_2$.
\end{theorem}
\begin{proof}
    This theorem is inspired by \citet[Theorem 5.3]{chen2024liebn}, which characterizes the LieBNs under isometric manifolds.
    As gyrogroups are natural generalizations of Lie groups, our GyroBN is expected to have similar results.
    The following proof follows a similar logic to the one by \citet[Theorem 5.3]{chen2024liebn}, except that all operations are gyro operations.
    
    For $i=1,2$, we denote \cref{eq:gyrobn_centering} and binary gyro barycenter on $\calM_i$ as $\xi^i(\cdot | M,v^2,B,s)$ and $\barcenter^i_\eta (\cdot,\cdot)$.
    Let $\calB=\{P_{1 \ldots N}\}$ and $f(\calB)=\{f(P_{1 \ldots N})\}$. We only need to show the following:
    \begin{align}
        \label{app:eq:mean_varaince_isometry}
        M_2&=f(M_1), v_1=v_2,\\
        \label{app:eq:xi_isometry}
        \xi^1(P_i|M_1,v^2,B,s) &= f^{-1} (\xi^2(f(P_i)|M_2,v^2,f(B),s)),\\
        \label{app:eq:barycenter_isometry}
        \barcenter^1_\eta (P,Q) 
        &= f^{-1} \left(\barcenter^2_\eta (f(P),f(Q)) \right), \forall P,Q \in \calM_1.
    \end{align}
    where $M_i$ and $v_i$ are the batch Fréchet mean and variance over $\calM_i$ for $i=1,2$. \cref{app:eq:barycenter_isometry,app:eq:mean_varaince_isometry} can be directly obtained by the invariance of gyrodistance under $f$ (\cref{lem:gyrogroups_isometry}).
    We only need to show \cref{app:eq:xi_isometry}.

    We have the following:
    \begin{equation}
        \begin{aligned}
            f^{-1} (\xi^2(f(P_i)|M_2,v^2,f(B),s))
            &= f^{-1} ( f(B) \oplus_2  \left( t \odot_2 \left( \ominus_2 M_2  \oplus_2 f(P_i) \right) \right)\\
            &\stackrel{(1)}{=} f^{-1} \circ f \left( B \oplus_1  \left( t \odot \left( \ominus_1 M_1  \oplus_1 P_i \right) \right) \right)\\
            &= B \oplus_1  \left( t \odot \left( \ominus_1 M_1  \oplus_1 P_i \right) \right) \\
            &= \xi^1(P_i|M,v^2,B,s),\\
        \end{aligned}
    \end{equation}
    where $t=\frac{s}{\sqrt{v^2+\epsilon}}$.
    The above derivation comes from the following.
    \begin{enumerate}[label=(\arabic*)]
        \item 
        $f$ is an isomorphism preserving gyro operations.
    \end{enumerate}
\end{proof}

As $\pi^{-1}: \graspp{p,n} \rightarrow \grasonb{p,n}$ is a Riemannian isometry, \cref{thm:liebn_pullback} indicates that the GyroBN under the projector perspective can be calculated by the ONB perspective by the following process:
\begin{enumerate}
    \item 
    mapping data into the ONB perspective by $\pi^{-1}: \graspp{p,n} \rightarrow \grasonb{p,n}$;
    \item 
    normalizing data by the GyroBN under $\grasonb{p,n}$;
    \item 
    mapping normalized data back to $\graspp{p,n}$ by $\pi$.
\end{enumerate}
Besides, both \cref{lem:gyrogroups_isometry} or \cref{thm:gyroinvariance} can guarantee theoretical control over the gyromean and gyrovariance under the projector perspective.

\section{Experimental Details}

\subsection{Reviews of ManifoldNorm and LRBN}
\label{app:subsec:manifoldnorm_lrbn}
The key differences between Riemannian normalization methods lie in the way in which they implement centering, scaling, and biasing. We briefly review the corresponding formulations of ManifoldNorm \citep[Algorithm~1--2]{chakraborty2020manifoldnorm} and LRBN \citep[Algorithm~2]{lou2020differentiating}. 

\mypara{ManifoldNorm.}
Let $\{x_{i} \}_{i=1}^N \subset \calM$ be a batch of activations and $G$ the isometry group of the homogeneous Riemannian space $\calM$. The core operations in ManifoldNorm are
\begin{align}
    \text{Centering: }& x^1_i =  \rieexp _e (\pt{\mu}{e}(\rielog_{\mu} (P_i))), \\
    \label{app:eq:manifoldnorm_scaling}
    \text{Scaling: }& x^2_i =  \rieexp _e (S (\rielog_e (x^1_i))), \\
    \text{Biasing: }& x^3_i =  g \cdot x^2_i,
\end{align}
where $\mu \in \calM$ is the Fréchet batch mean, $g \in G$ is the bias parameter with $\cdot$ denoting the group action, 
and $S$ is a diagonal scaling matrix. 
Here, \cref{app:eq:manifoldnorm_scaling} assumes the identification $T_e \calM \cong \bbR{m}$ with $m = \dim(\calM)$. 

\mypara{LRBN.}
Following the same notation, the core operation in LRBN is
\begin{equation}
    x^1_i =  \rieexp _\beta \left(\frac{s}{v} \pt{\mu}{\beta}(\rielog_{\mu} (P_i)) \right),
\end{equation}
where $v^2$ is the Fréchet variance, $s$ is the scaling parameter, and $\beta \in \calM$ is the bias parameter. 

\subsection{Details on the Grassmannian Experiments}
\label{app:subsec:exp_details_grass}

\subsubsection{Data Sets and Preprocessing}
\label{app:subsubsec:datasets_preprocessing_grass}

\mypara{HDM05\footnote{\url{https://resources.mpi-inf.mpg.de/HDM05/}} \citep{muller2007documentation}.}
It consists of 2,273 skeleton-based motion capture sequences executed by different actors. Each frame records the 3D coordinates of 31 joints. We remove under-represented clips, trimming the data set down to 2,086 instances over 117 classes. 

\mypara{NTU60\footnote{\url{https://github.com/shahroudy/NTURGB-D}} \citep{shahroudy2016ntu}.} This data set contains 56,880 skeleton sequences classified into 60 classes, where each frame includes 3D coordinates of 25 or 50 joints. We focus on mutual actions and adopt the cross-view protocol \citep{shahroudy2016ntu}.

\mypara{NTU120\footnote{\url{https://github.com/shahroudy/NTURGB-D}} \citep{liu2019ntu}.}
This data set contains 114,480 sequences in 120 action classes. We again use mutual actions and adopt the cross-setup protocol \citep{liu2019ntu}.

Following \citet{nguyen2023building}, each sequence is represented as a Grassmannian matrix of size $93 \times 10$, $150 \times 10$, and $150 \times 10$ for HDM05, NTU60, and NTU120, respectively.

\subsubsection{Basic Layers in GyroGr}
\label{app:subsubsec:gyrogr}

GyroGr \citep{huang2018building} mimics conventional densely connected feedforward networks and is composed of three basic building blocks. Given an ONB Grassmannian matrix $U^{k-1}$, the gyrotranslation and ProjMap layers are defined as
\begin{align}
    \label{app:eq:gyrotranslation}
    &\text{Gyrotranslation: }  U^{k} = W^{k} \oplus_{\mathrm{Gr}} U^{k-1}, \quad W^{k} \in \grasonb{p,n}, \\
    &\text{ProjMap: }  P^{k} = (U^{k-1})^\top U^{k-1}.
\end{align}

In addition, the pooling is performed via the space of projection matrices. Specifically, Grassmannian data are first mapped to the space of projection matrices via the ProjMap layer. A standard mean pooling operation is applied to the resulting projection matrices. Finally, the pooled matrices are projected back to the ONB Grassmannian by SVD. The entire procedure can be expressed as
\begin{equation}
    \label{app:eq:grass_pooling}
    \begin{aligned}
        P^k &= f_{\mathrm{p}} \left( ({U}^{k-1})^\top {U}^{k-1} \right), \\
        U^{k} &= O^k_{1:p}, \quad \text{where } P^k \stackrel{\mathrm{SVD}}{:=} O^k \Sigma^k (O^k)^\top,
    \end{aligned}
\end{equation}
where $f_{\mathrm{p}}$ is a regular mean pooling.

\subsubsection{Trivialization}
\label{app:subsubsec:trivialization_details_grass}

Following \citet{nguyen2023building}, we adopt the trivialization trick \citep{lezcano2019trivializations} for the Grassmannian parameters in the gyrotranslation and our GyroBN layers. Each Grassmannian parameter $U \in \grasonb{p,n}$ is parameterized by a matrix $\frakU \in \bbR{(n-p) \times p}$ such that
\begin{equation*}
    \left[\begin{array}{cc}
    0 & -\frakU^T \\
    \frakU & 0
    \end{array}\right]=\left[\overline{UU^\top }, \idpp \right] .
\end{equation*}
where $\overline{(\cdot)}=\rielog_{\idpp}(\cdot)$. The parameter $U$ can be retrieved by
\begin{equation}
    \label{app:eq:trivilization_grass}
    U = \mexp \left(\left[\overline{UU^\top }, \idpp \right]\right) \idonb =\mexp \left(\left[\begin{array}{cc}
    0 & -\frakU^T \\
    \frakU & 0
    \end{array}\right]\right) \idonb.
\end{equation}
This reparameterization ensures that all variables lie in Euclidean space, thereby allowing the direct use of PyTorch optimizers \citep{paszke2019pytorch} and avoiding the additional computational burden of Riemannian optimization.

\subsubsection{Analysis on the Efficiency of GyroBN}
\label{app:subsubsec_analysis_gyrobn_efficiency}
As shown in \cref{subsubsec:main_results_grass}, our Grassmannian GyroBN is more efficient than the ManifoldNorm \citep[Algorithm 1--2]{chakraborty2020manifoldnorm} or RBN \citep[Algorithm 2]{lou2020differentiating}. The key difference lies in their methods for centering, biasing, and scaling. GyroBN uses gyro operations, while ManifoldNorm and RBN rely on Riemannian operators, such as parallel transport and the Riemannian logarithmic and exponential maps. This distinction underpins the efficiency of GyroBN. The three primary contributing factors, ranked by importance, are as follows:
\begin{enumerate}
    \item
    \mypara{Riemann vs. Gyro.}
    The Riemannian operators over the Grassmannian involve computationally expensive processes like SVD decomposition or matrix inversion (see \cref{tab:riem-gyro-operators-grass-onb} for Riemannian exp and log, and \citet[Theorem 2.4]{edelman1998geometry} for parallel transport). Consequently, ManifoldNorm and RBN require multiple SVD or matrix inversion operations. In contrast, GyroBN is relatively simpler. As discussed in \cref{subsec:grassmannian_gyrobn}, the gyro operation can be further simplified, and the involved SVD is performed on a reduced $p \times p$ matrix instead of $n \times p$. Additionally, computationally intensive matrix exponentiation is efficiently approximated using the Cayley map.
    \item
    \mypara{Reduced Matrix Products.} 
    As shown in \cref{tab:riem-gyro-operators-grass-onb}, each Riemannian operator involves several matrix products over $n \times p$ matrices. GyroBN reduces these to matrix products over $(n-p) \times p$ or $p \times p$ matrices, as shown in \cref{prop:fast_bracket_grassmannian}.
    \item 
    \mypara{Optimization.}
    The optimization of bias parameter in GyroBN is simpler. ManifoldNorm uses an $n \times n$ orthogonal matrix for biasing, and RBN employs an $n \times p$ Grassmannian matrix, both requiring Riemannian optimization. In contrast, GyroBN applies trivialization tricks (\eqnref{app:eq:trivilization_grass}), making the bias parameter a $(n-p) \times p$ Euclidean matrix. Furthermore, \cref{app:eq:trivilization_grass} and \cref{eq:gyrobn_grass_biasing} share a similar form, allowing them to be jointly simplified. Although RBN could adopt similar trivialization for biasing, this would introduce an additional Riemannian exp step, leaving little advantage over the Riemannian optimization. In summary, GyroBN benefits from joint simplification with trivialization, whereas the other two Grassmannian BN methods require additional Riemannian optimization.
\end{enumerate}

\subsection{Details on the Constant Curvature Space Experiments}
\label{app:subsec:exp_details_ccs}

\subsubsection{Data Sets}
\label{app:subsec:datasets_ccs}

\mypara{Cora \citep{sen2008collective}.}
It is a citation network where the nodes represent scientific papers in the area
of machine learning, the edges are citations between them, and the labels of the nodes are academic (sub)areas.

\mypara{Disease \citep{anderson1991infectious}. }
It represents a disease propagation tree, simulating the SIR disease transmission model, with each node representing either an infection or a non-infection state.

\mypara{Airport \citep{zhang2018link}. } 
It is a transductive data set where nodes represent airports and edges represent airline routes as from OpenFlights.org.

\mypara{Pubmed \citep{namata2012query}. }
This is a standard benchmark that describes citation networks where nodes represent scientific papers in the area of medicine, the edges are citations between them, and the node labels are academic (sub)areas.

\subsubsection{Brief Review of Hyperbolic Neural Networks}
\label{app:subsubsec:review_hnns_layers}

\mypara{Hyperboloid Neural Networks.}
Let $x \in \bbh{n}$ be the input vector and $W \in \bbR{m \times n+1}, v \in \bbR{n+1}$ the weight parameters. The Lorentz Fully-Connected (FC) layer \citep[Equation 3]{chen2021fully} and activation layer \citep[Equation 13]{bdeir2024fully} are defined in a spacetime manner:
\begin{align}
    \label{app:eq:act_lnn}
    \text{Activation: } y 
    &=\begin{bmatrix}
    \sqrt{\left\|\psi\left(x_s\right)\right\|^2-1 / K} \\
    \psi\left(x_s\right)
    \end{bmatrix},  \\  
    \text{FC: } y 
    &=\begin{bmatrix}
    \sqrt{\|\phi(W x, v)\|^2-1 / K} \\
    \phi(W x, v)
    \end{bmatrix}, \\
    & \text{ with } 
    \phi(W x, v) =\lambda \sigma\left(v^T x+b^{\prime}\right) \frac{W \psi(x)+b}{\|W \psi(x)+b\|},
\end{align}
where $\psi$ is an activation, $\lambda>0$ is a learnable scaling parameter, and $b \in \bbR{n}, \psi, \sigma$ denote the bias, activation, and sigmoid function, respectively.

\mypara{Poincaré MLR.}
\citet{lebanon2004hyperplane} first reformulated the Euclidean Multinomial Logistic Regression (MLR) $p(y{=}k\mid x) \propto \exp(\inner{a_k}{x} - b_k)$ via the point-to-hyperplane distance: 
\begin{align*}
    p(y{=}k\mid x) &\propto \exp \left(\sign(\inner{a_k}{x}{-}b_k) \norm{a_k} d(x,H_{a_k,b_k}) \right), \\
    H_{a, b} &=\left\{x \in \bbR{n}:\inner{a}{x}-b=0\right\}, \quad \text { where } a \in \bbR{n}, \text { and } b \in \bbRscalar.    
\end{align*}
\citet[Equations~24--25]{ganea2018hyperbolic} generalized this formulation to the Poincaré ball via geometric reinterpretation, and \citet[Section~3.1]{shimizu2020hyperbolic} further simplified it using trivialization. The resulting closed form is
\begin{equation*}
    v_k(x) = \frac{2\|z_k\|}{\sqrt{|K|}}\,
    \operatorname{asinh} \left(
    \lambda_x^{K} \langle \sqrt{|K|} x, [z_k]\rangle \cosh(2\sqrt{|K|} r_k)
    -\left(\lambda_x^{K} - 1\right)\sinh(2\sqrt{|K|} r_k)\right),
\end{equation*}
where $\lambda_x^{K}=2\,(1-|K|\|x\|^2)^{-1}$ is the conformal factor, $p(y{=}k\mid x)\propto\exp\left(v_k(x)\right)$, and $[z_k] = \frac{z_k}{\norm{z_k}}$. Here, $z_k \in \bbR{n}$ and $r_k \in \bbRscalar$ are parameters. Note that $\lim_{K\to 0}v_k(x)=4(\inner{a_k}{x}-b_k)$.

\mypara{Poincaré FC Layer.}
\citet{shimizu2020hyperbolic} extended the Euclidean FC layer to the Poincaré ball via point-to-hyperplane distances. In Euclidean spaces, the FC can be written element-wise as $y_k = \inner{a_k}{x} - b_k$ with $x, a_k\in\bbR{n}$ and $b_k\in\bbRscalar$, which may be interpreted as a linear transform whose output coordinate is the signed distance to the hyperplane passing through the origin and orthogonal to the $k$-th axis. Combining this view, the Poincaré FC layer takes the closed form:
\begin{equation*}
y = \frac{w}{1+\sqrt{1+|K| \|w\|^2}}, 
\qquad
w_k = |K|^{-1/2} \sinh \left(\sqrt{|K|} v_k(x)\right),
\end{equation*}
where $c=|K|$ is the magnitude of the curvature. Here, $Z=\{z_k\}_{k=1}^m$ and $r=\{r_k\}_{k=1}^m$ parameterize the orientations and biases, and $v_k(x)$ is the Poincaré MLR. 

\mypara{Poincaré $\beta$-Concatenation.} 
It generalizes the Euclidean concatenation into the hyperbolic Poincaré ball, stabilizing the norm of the Poincaré vector \citep[Section 3.3]{shimizu2020hyperbolic}. Given inputs $\{x_i \in \pball{n_i}\}_{i=1}^N$, it is defined as
\begin{equation*}
    \rieexp_{\Rzero} \left( \beta_n \left( \beta_{n_1}^{-1} v_1^\top, \cdots, \beta_{n_N}^{-1} v_{N}^\top \right) \right)^\top \in \pball{n},
\end{equation*}
where $v_i = \rielog_{\Rzero}(x_i)$, $n=\sum_{i=1}^N n_i$, and $\beta_\alpha = \mathrm{B}\left(\nicefrac{\alpha}{2}, \nicefrac{1}{2}\right)$ are the beta function.

\mypara{RResNet.}
Euclidean residual blocks can be written as
\begin{equation}\label{app:eq:euc_res_block}
    x^{(i)} = x^{(i-1)} + n_i \left(x^{(i-1)}\right),
\end{equation}
where $n_i$ is a network. \citet{katsman2024riemannian} generalized this to manifolds by replacing addition with the Riemannian exponential map:
\begin{equation} \label{app:eq:riem_res_block}
x^{(i)} = \rieexp _{x^{(i-1)}}\left( \ell_i(x^{(i-1)}) \right),
\end{equation}
where $\ell_i: \calM \rightarrow T\calM$ outputs a vector field parameterized by the neural network. As $\rieexp_x(v) = x+v$ for the Euclidean space, it can be immediately shown that \cref{app:eq:riem_res_block} naturally extends \cref{app:eq:euc_res_block} to manifolds.

\subsection{Details on the Correlation Experiments}
\label{app:subsec:exp_details_cor}

\subsubsection{Data Sets and Preprocessing}
\label{app:subsec:dataset_preprocessing_cor}

\mypara{HDM05.}
Please refer to \cref{app:subsec:exp_details_grass}.

\mypara{Radar\footnote{\url{https://www.dropbox.com/s/dfnlx2bnyh3kjwy/data.zip?dl=0}} \citep{brooks2019riemannian}.}
It consists of 3,000 synthetic radar signals equally distributed in 3 classes.

\mypara{FPHA\footnote{\url{https://github.com/guiggh/hand_pose_action}} \citep{garcia2018first}.}  
It includes 1,175 skeleton-based first-person hand gesture videos of 45 different categories with 600 clips for training and 575 for testing. Each frame contains the 3D coordinates of 21 hand joints.

For the HDM05 and FPHA data sets, we preprocess each sequence using the code\footnote{\url{https://ravitejav.weebly.com/kbac.html}} provided by \citet{vemulapalli2014human} to normalize the lengths of the body parts and ensure invariance with scale and view.

\subsubsection{Correlation Input in CorNets}
\label{app:subsubsec:cor_inputs}
Following \citet{chen2025cornet}, we model each sequence into multichannel correlation matrices. Specifically, according to \citet{wang2024grassatt,nguyen2024matrix}, each sample is modeled as a multichannel SPD tensor. Then, each SPD matrix is transformed to their correlation matrix by 
\begin{equation*}
    \coropt: \spd{n} \ni \Sigma \longmapsto C= \bbD(\Sigma)^{-\frac{1}{2}} \Sigma \bbD(\Sigma)^{-\frac{1}{2}} \in \cor{n}.
\end{equation*}
After preprocessing, the input correlation tensor shapes are $[7,20,20]$, $[3,28,28]$, and $[9,28,28]$ on the Radar, HDM05, and FPHA data sets, respectively. The following introduces SPD modeling.

\mypara{HDM05 and FPHA}.
We first identify the closest left (right) neighbor of every joint based on their distance to the hip (wrist) joint, and then combine the 3D coordinates of each joint and those of its left (right) neighbor to create a feature vector for the joint. For a given frame $t$, we compute its Gaussian embedding \citep{lovric2000multivariate}:
\begin{equation*}
    Y_t=(\operatorname{det} \Sigma_{t})^{-\frac{1}{n+1}}
    \left[\begin{array}{cc} \Sigma_t+\mu_t\left(\mu_t\right)^T & \mu_t \\
    \left(\mu_t\right)^T & 1
\end{array}\right],
\end{equation*}
where $\mu_t$ and $\Sigma_t$ are the mean vector and covariance matrix computed from the set of feature vectors within the frame. The lower part of the matrix $\log \left(Y_t\right)$ is flattened to obtain a vector $\tilde{v}_t$. All vectors $\tilde{v}_t$ within a time window $[t, t+c-1]$, where $c$ is determined from a temporal pyramid representation of the sequence (the number of temporal pyramids is set to 2 in our experiments), are used to compute a covariance matrix:
\begin{equation*}
    \widetilde{\Sigma}_t=\frac{1}{c} \sum_{i=t}^{t+c-1}\left(\tilde{v}_i-\overline{v}_t\right)\left(\tilde{v}_i-\overline{v}_t\right)^T,
\end{equation*}
where $\overline{v}_t=\frac{1}{c} \sum_{i=t}^{t+c-1} \tilde{v}_i$. The resulting $\{\widetilde{\Sigma}_t\}$ are the covariance matrices that we need. On the FPHA data set, we generate covariance based on three sets of neighbors: left, right, and vertical (bottom) neighbors. 

\mypara{Radar.} 
We follow \citep{wang2024grassatt} to use the temporal convolution followed by a covariance pooling layer to obtain a multi-channel covariance tensor of shape $[c,20,20]$. 

\subsection{Hardware}
All experiments are conducted on a single NVIDIA Quadro RTX A6000 48GB GPU.

\section{Proofs}
\label{app:sec:proofs}

\linkofproof{prop:grassmannian_pseudo_reductive_gyrogroups}
\subsection{Proof of \cref{prop:grassmannian_pseudo_reductive_gyrogroups}}
\begin{proof}  
    By \citet[Lemma 2.3]{nguyen2023building}, easy computations show that \cref{eq:pseudo_reduction} holds for $\grasonb{p,n}$ iff it holds for $\graspp{p,n}$.
    Without loss of generality, we prove the case for the projector perspective.

    Given any $P, Q\in \graspp{p,n}$, Definition 3.18 by \citet{nguyen2022gyro} gives the expression for gyration:
    \begin{equation*}
        \gyr[\ominus P,  P] Q=F(\ominus P, P) Q\left(F(\ominus P, P)\right)^{-1},
    \end{equation*}
    with $F(\ominus P, P)$ defined as
    \begin{equation*}
        F(\ominus P, P)= \mexp \left(-\left[\overline{\ominus P \oplus  P}, \idpp\right]\right) \mexp \left(\left[\overline{\ominus P}, \idpp\right]\right) \mexp \left(\left[\overline{P}, \idpp\right]\right),
    \end{equation*}
    where $\overline{(\cdot)} = \rielog_{\idpp}(\cdot)$.
    This equation can be further simplified as
    \begin{equation*}
        \begin{aligned}
            F(\ominus P, P)
            &\stackrel{(1)}{=} \mexp (0) \mexp \left(\left[\overline{\ominus P}, \idpp\right]\right) \mexp \left(\left[\overline{P}, \idpp\right]\right) \\
            &\stackrel{(2)}{=} \mexp \left(\left[\overline{\ominus P}, \idpp\right]\right) \mexp \left(\left[\overline{P}, \idpp\right]\right) \\
            &\stackrel{(3)}{=} I_{n}.
        \end{aligned}
    \end{equation*}
    The above derivation follows from
    \begin{enumerate}[label=(\arabic*)]
        \item 
        $\overline{\ominus P \oplus P} = \overline{\idpp} = 0 \in \bbR{n \times n}.$
        \item 
        $\mexp (0) = I_n$.
        \item 
        $\overline{\ominus P}= - \overline{P}$ and $\mexp \left(\left[-\overline{P}, \idpp\right]\right)=\mexp \left(\left[\overline{P}, \idpp\right]\right)^{-1}$.
    \end{enumerate}

    Therefore, $\gyr[\ominus P, P]$ is the identity map.
\end{proof}

\linkofproof{thm:pseudo_reductive_gyrogroups_properties}
\subsection{Proof of \cref{thm:pseudo_reductive_gyrogroups_properties}}
\label{app:subsec:prf:pseudo_reductive_gyrogroups_properties}
\begin{proof} 
This theorem follows \citep[Theorems. 2.10--2.11]{ungar2022analytic}, which presents some useful properties for gyrogroups. 
We argue that all the properties except $\gyr[a, a]=\id$ are independent of the left reduction law (G4), and are therefore satisfied on pseudo-reductive gyrogroups. 
All the properties can be proven in the same way as the ones for Theorems. 2.10--2.11 by \citet{ungar2022analytic}. We summarize the logic in the following:
\begin{itemize}
    \item 
    left gyroassociativity $\Rightarrow$ \ref{enu:prgp_1} 
    \item 
    left gyroassociativity + \ref{enu:prgp_1} $\Rightarrow$ \ref{enu:prgp_2} 
    \item 
    definition $\Rightarrow$ \ref{enu:prgp_3} 
    \item 
    left gyroassociativity + \ref{enu:prgp_1} + \ref{enu:prgp_3}  $\Rightarrow$ \ref{enu:prgp_4} 
    \item 
    definition $\Rightarrow$ \ref{enu:prgp_5} 
    \item 
    left gyroassociativity + (G2) + \ref{enu:prgp_1} + \ref{enu:prgp_3} + \ref{enu:prgp_4} + \ref{enu:prgp_5} $\Rightarrow$ \ref{enu:prgp_6} 
    \item 
    \ref{enu:prgp_1}+\ref{enu:prgp_6} $\Rightarrow$ \ref{enu:prgp_7} 
    \item 
    left gyroassociativity +\ref{enu:prgp_3} $\Rightarrow$ \ref{enu:prgp_8} 
    \item 
    left gyroassociativity + left cancellation in \ref{enu:prgp_8} $\Rightarrow$ \ref{enu:prgp_9}
    \item 
    gyro identity in \ref{enu:prgp_9} $\Rightarrow$ \ref{enu:prgp_10}
    \item 
    \ref{enu:prgp_10} $\Rightarrow$ \ref{enu:prgp_11}
    \item 
    left cancellation in \ref{enu:prgp_8} + gyro identity in \ref{enu:prgp_9}   $\Rightarrow$ \ref{enu:prgp_12}
    \item 
    left cancellation in \ref{enu:prgp_8} + gyro identity in \ref{enu:prgp_9} $\Rightarrow$ \ref{enu:prgp_13}
\end{itemize}
\end{proof}

\linkofproof{lem:gyro_geodesic_dist}
\subsection{Proof of \cref{lem:gyro_geodesic_dist}}
\begin{proof}
    \begin{equation}
        \begin{aligned}
          y &\stackrel{(1)}{=}  x \oplus ( \ominus x \oplus y) \\
          &\stackrel{(2)}{=} \rieexp_{x} \left( \pt{e}{x} \left( \rielog_e \left( \ominus x \oplus y   \right)\right)\right)\\
          &\stackrel{(3)}{\Rightarrow} \rielog_{x}(y) = \pt{e}{x} \left( \rielog_e \left( \ominus x \oplus y   \right)\right).
        \end{aligned}
    \end{equation}
        The above comes from the following.
    \begin{enumerate}[label=(\arabic*)]
        \item 
        Left cancellation law.
        \item 
        Definition of gyroaddition.
        \item 
        Applying both sides with $\rielog_x(\cdot)$.
    \end{enumerate}
    By the last equation, we have
    \begin{equation}
        \begin{aligned}
            \dist(x,y) 
            &= \norm{\rielog_x (y)}_x \\
            &= \norm{\pt{e}{x} \left( \rielog_e \left( \ominus x \oplus y   \right)\right)}_x \\
            &\stackrel{(1)}{=} \norm{ \rielog_e \left( \ominus x \oplus y \right) }_e \\
            &= \gyrdist(x,y),
        \end{aligned}
    \end{equation}
    where (1) comes from 
    \begin{itemize}
        \item 
        Parallel transport preserving the norm \citep[Section 3.1]{do1992riemannian}.
        \item 
        $\pt{x}{e} \circ \pt{e}{x}(v)=v, \forall v \in T_e\calM$.
    \end{itemize}
\end{proof}

\linkofproof{lem:isometry_for_gyro}
\subsection{Proof of \cref{lem:isometry_for_gyro}}
\begin{proof}
We denote $\rielog$, $\gyr$ and $\gyrnorm{\cdot}$ as the Riemannian logarithm and gyronorm on $\{\calM,g \}$, while  $\widetilde{\rielog}$, $\gyrw$ and $\gyrnormw{\cdot}$ are the counterparts on $\{\wcalM, \widetilde{g}\}$. We recall the following from \citet[Lemmas 2.1--2.3]{nguyen2023building}:
\begin{align}
    \label{app:eq:iso_gyro_add}
    x \oplus y &= \phi^{-1} \left( \phi(x) \widetilde{\oplus} \phi(y) \right), \\
    \label{app:eq:iso_gyro_scalarprod}
    t \otimes x &= \phi^{-1} \left( t \widetilde{\otimes} \phi(x) \right), \\
    \label{app:eq:iso_gyro_gyration}
    \gyr[x, y] z &= \phi^{-1} \left( \gyrw[\phi(x), \phi(y)] \phi(z) \right),
\end{align}
where $x,y,z \in \calM$.

\mypara{Gyrodistance.}
\begin{equation*}
    \begin{aligned}
        \widetilde{\gyrdist}(\phi(x), \phi(y)) 
        &= \gyrnormw{ \widetilde{\ominus} \phi(x) \widetilde{\oplus} \phi(y) } \\
        &\stackrel{(1)}{=} \gyrnormw{ \phi(\ominus x \oplus y) } \\
        &= \inner{ \widetilde{\rielog _{\widetilde{e}}} (\phi(\ominus x \oplus y))}{\widetilde{\rielog _{\widetilde{e}}} (\phi(\ominus x \oplus y))}_{\widetilde{e}} \\
        &\stackrel{(2)}{=} \inner{ \rielog _{e} (\ominus x \oplus y) }{ \rielog _{e} (\ominus x \oplus y) }_{e} \\
        &= \gyrnorm{\ominus x \oplus y} \\
        &= \gyrdist(x, y).
    \end{aligned}
\end{equation*}
The derivation above comes from the following.
\begin{enumerate}[label=(\arabic*)]
    \item
    By \cref{app:eq:iso_gyro_add,app:eq:iso_gyro_scalarprod}:
    \begin{equation*}
        \phi(\ominus x \oplus y) = \phi(\ominus x) \widetilde{\oplus} \phi(y).
    \end{equation*}
    \item
    By the isometry:
    \begin{align*}
        \rielog_{x} (y)
        &= (\phi_{*, x})^{-1} \left( \widetilde{\rielog}_{ \phi(x) } (\phi(y))  \right), \forall x,y \in \calM, \\
        \inner{v}{w}_{x} 
        &= \inner{\phi_{*,x} (v)}{ \phi_{*,x}(w)}_{\phi(x)}, \forall x \in \calM \text{ and } \forall v,w \in T_x\calM,
    \end{align*}
    where $\phi_{*,x}$ is the differential map. Here, the RHSs contain the operators over $\widetilde{\calM}$, where the LHSs involve the ones over $\calM$.
\end{enumerate}

\mypara{Gyroisometry.}
Given any $x,y,z,a \in \calM$, we have the following by the isometry of $\phi$.

For the gyroinverse:
\begin{equation*}
    \begin{aligned}
        \gyrdistw(\widetilde{\ominus} \phi(x), \widetilde{\ominus} \phi(y)) 
        &= \gyrdistw(\phi(\ominus x), \phi(\ominus y)) \\
        &= \gyrdist( \ominus x, \ominus y) \\
        &= \gyrdist( x, y) \\
        &= \gyrdistw(\phi(x), \phi(y)).
    \end{aligned}
\end{equation*}
For the gyration:
\begin{equation*}
    \begin{aligned}
        & \gyrdistw(\gyrw[\phi(z), \phi(a)] \phi(x), \gyrw[\phi(z), \phi(a)] \phi(y)) \\ 
        &= \gyrdistw( \phi(\gyr[z, a] x), \phi(\gyr[z, a] y)) \\
        &= \gyrdist( \gyr[z, a] x, \gyr[z, a] y) \\
        &= \gyrdist( x, y ) \\
        &= \gyrdistw( \phi(x), \phi(y)).
    \end{aligned}
\end{equation*}
For the left gyrotranslation:
\begin{equation*}
    \begin{aligned}
        & \gyrdistw(\phi(z) \widetilde{\oplus} \phi(x), \phi(z) \widetilde{\oplus} \phi(y) ) \\ 
        &= \gyrdistw(\phi(z \oplus x), \phi(z \oplus y) ) \\
        &= \gyrdist(z \oplus x, z \oplus y ) \\
        &= \gyrdist( x, y ) \\
        &= \gyrdistw( \phi(x), \phi(y)).
    \end{aligned}
\end{equation*}
\end{proof}

\linkofproof{thm:iff_gyroauto_gyroisometries}
\subsection{Proof of \cref{thm:iff_gyroauto_gyroisometries}}
We first prove a useful lemma.

 \begin{lemma}[Left Gyrotranslation Law]
 \label{app:lem:left_gyrotranslation}
    Every pseudo-reductive gyrogroup $\left\{ G, \oplus \right\}$ verifies the left gyrotranslation law:
    \begin{equation*}
        \ominus \left(x \oplus y \right) \oplus \left(x \oplus z \right) = \gyr[x, y]\left(\ominus y \oplus z\right), \quad \forall x, y, z \in G.
    \end{equation*}
\end{lemma}
\begin{proof}
    This lemma generalizes Lemmas I.1 and L.1 by \citet{nguyen2023building}, which prove the left gyrotranslation law on the specific gyrogroups of the SPD and Grassmannian manifolds. 
    Their proof only relies on the left cancellation and the basic axioms (G1-3).
    Note that the original proof of left gyrotranslation on the Grassmannian \citep[Lemma I.1]{nguyen2023building} is questionable, as it relies on the left cancellation of gyrogroups, and the Grassmannian is not a gyrogroup but a non-reductive gyrogroup. Fortunately, as we show in \cref{thm:pseudo_reductive_gyrogroups_properties}, the general pseudo-reductive gyrogroups, including the Grassmannian, enjoy left cancellation.
    Therefore, all the proof by \citet[Lemma I.1]{nguyen2023building} can be readily generalized into the general pseudo-reductive gyrogroups.
\end{proof}
    
\begin{proof}[Proof of \cref{thm:iff_gyroauto_gyroisometries}]
    $\Rightarrow$:
    For any $z,a \in G$, the gyroautomorphism can be expressed by the gyrator identity in \cref{thm:pseudo_reductive_gyrogroups_properties}:
    \begin{align*}
        \gyr[x,y] z &= X \oplus \bar{z},\\
        \gyr[x,y] a &= X \oplus \bar{a},
    \end{align*}
    where $X = \ominus (x \oplus y)$, $\bar{z} = x \oplus (y \oplus z)$, and $\bar{a} = x \oplus (y \oplus a)$. Then Equation (30) in \citep{nguyen2023building} for the specific Grassmannian can be directly extended into the pseudo-reductive gyrogroup, as it only relies on left gyrotranslation, invariance of the norm under gyroautomorphisms, and the axioms of (G1-G3).

    $\Leftarrow$:
    \begin{equation*}
        \begin{aligned}
            \gyrnorm{\gyr[x,y](z)} 
            &= \gyrnorm{\gyr[x,y](\ominus e \oplus z)} \quad \text{(\ref{enu:prgp_7} in \cref{thm:pseudo_reductive_gyrogroups_properties} indicates $\ominus e = e$)}\\
            &= \gyrnorm{ \ominus \gyr[x,y](e) \oplus  \gyr[x,y](z)} \quad \text{(automorphism)}\\
            &= \dist \left(\gyr[x,y](e), \gyr[x,y](z) \right) \\
            &= \dist \left(e, z \right) \\
            &= \gyrnorm{\ominus e \oplus z} \\
            &= \gyrnorm{z}.
        \end{aligned}
    \end{equation*} 
\end{proof}

\linkofproof{thm:gyroisometries}
\subsection{Proof of \cref{thm:gyroisometries}}

\begin{proof}
    Given any $x, y, z \in G$, we make the following proof.
    
    \mypara{Gyroisometry of the Left Gyrotranslation.}
    This property generalizes Theorems 2.12 and 2.16 by \citet{nguyen2023building}, which deal with the gyrotranslation in the SPD and Grassmannian, respectively. We have the following:
    \begin{equation*}
        \begin{aligned}
            \dist(L_{x} (y), L_{x} (z))
            &= \dist(x \oplus y, x \oplus z) \\
            &= \gyrnorm{\ominus (x \oplus y) \oplus (x \oplus z)} \\
            &= \gyrnorm{\gyr[x, y]\left(\ominus y \oplus z\right)} \quad \text{ (left gyrotranslation law)} \\
            &= \gyrnorm{\ominus y \oplus z} \quad \text{ (gyroisometry of the automorphism)} \\
            &= \dist(y, z).
        \end{aligned}
    \end{equation*}

    \mypara{Gyroisometry of the Gyroinverse.}
    \begin{equation*}
        \begin{aligned}
            & \dist(\ominus x, \ominus y)\\ 
            &= \gyrnorm{x \ominus y} \\
            &= \gyrnorm{\ominus y \oplus x} \quad \text{ (gyrocommutativity and gyroisometry of the automorphism)} \\
            &= \dist(y, x) \\
            &= \dist(x, y) \quad \text{ (Symmetry of the geodesic distance)}.\\
        \end{aligned}
    \end{equation*}
\end{proof}

\linkofproof{thm:gyroinvariance}
\subsection{Proof of \cref{thm:gyroinvariance}}
\label{app:subsec:proof_gyroinvariance}

Different from our conference version \citep[Appendix~G.4]{chen2025gyrogroup}, we give a much more concise argument based on \cref{lem:gyro_geodesic_dist,thm:iff_gyroauto_gyroisometries,thm:gyroisometries}.

\begin{proof}
    First, all these gyrospaces are characterized by \cref{eq:gyro_addtion,eq:gyro_scalar_product}. By \cref{lem:gyro_geodesic_dist}, their gyrodistances agree with the geodesic distances. It thus remains to establish the gyroisometries. 

    As shown by \cref{thm:iff_gyroauto_gyroisometries,thm:gyroisometries}, it suffices to show that gyrations in each space preserve the gyronorm. This argument on the SPD and ONB Grassmannian has already been proven \citep[Lemmas L.2 and I.2]{nguyen2023building}. Since the ONB Grassmannian is isometric to the PP Grassmannian via \cref{eq:iso_grass}, \cref{lem:isometry_for_gyro} implies that the same arguments apply for the PP. We therefore only need to treat $\stereo{n}$ with $K \leq 0$. As the Euclidean case $\bbR{n}$ is trivial, we only need to show the Poincaré ball. In the following, $a,b,x,y$ are arbitrary points in $\pball{n}$.  

    \mypara{Norm Invariance under Gyrations.}
    As the Poincaré ball forms a real inner product gyrovector spaces \citep[Definition 6.2 and Theorem 6.85]{ungar2022analytic}, any gyration preserves the Euclidean norm: 
    \begin{equation*}
        \norm{\gyr[a,b](x)} = \norm{x}, \quad \forall x \in \stereo{n}.
    \end{equation*}
    For the gyronorm, we further have
    \begin{equation*}
        \begin{aligned}
            \gyrnorm{\gyr[a,b] x}
            &= 2\| \rielog _{\Rzero}  (\gyr[a,b] x )\|_{\Rzero} \\
            &= \tfrac{2}{\sqrt{|K|}} \tanh^{-1}\!\left(\sqrt{|K|}\,\|\gyr[a,b] x \|\right) \\
            &= \tfrac{2}{\sqrt{|K|}} \tanh^{-1}\!\left(\sqrt{|K|}\,\|x\|\right) \\
            &= \gyrnorm{x}.
        \end{aligned}
    \end{equation*}
\end{proof}

\linkofproof{thm:gyrobn}
\subsection{Proof of \cref{thm:gyrobn}}
\begin{proof}
        According to \cref{thm:gyroisometries}, any left gyrotranslation is a gyroisometry. For any $y \in \calM$, we have the following:
    \begin{equation}
        \label{eq:gyobn_homogeneity_intermediate}
        \begin{aligned}
             \dist\left(\beta \oplus x_i, y \right)
            &\stackrel{(1)}{=} \dist \left(\ominus \beta \oplus (\beta \oplus x_i), \ominus \beta \oplus  y \right) \\
            &\stackrel{(2)}{=} \dist \left((\ominus \beta \oplus \beta) \oplus \gyr[\ominus \beta,\beta] (x_i) , \ominus \beta \oplus  y \right) \\
            &\stackrel{(3)}{=} \dist \left( x_i, \ominus \beta \oplus  y \right).
        \end{aligned}
    \end{equation}
    The above comes from the following.
    \begin{enumerate}[label=(\arabic*)]
        \item Any left gyrotranslation is a gyroisometry.
        \item Left gyroassociative law.
        \item $\ominus \beta \oplus \beta = e$ and pseudo-reduction.
    \end{enumerate}

    Denoting the gyromean of $\{x_i\}$ and $\{\beta \oplus x_i \}$ as $\mu$ and $\widetilde{\mu}$, we have the following:
    \begin{equation}
        \begin{aligned}
           \beta \oplus \mu 
           &\stackrel{(1)}{=} \beta \oplus (\ominus \beta \oplus  \widetilde{\mu}) \\
           &\stackrel{(2)}{=} \gyr[\beta, \ominus \beta] (\widetilde{\mu}) \\
           &\stackrel{(3)}{=} \widetilde{\mu}. \\
        \end{aligned}
    \end{equation}
    The above comes from the following.
    \begin{enumerate}[label=(\arabic*)]
        \item \cref{eq:gyobn_homogeneity_intermediate} indicates that $\mu = \ominus \beta \oplus \widetilde{\mu}$.
        \item Left gyroassociative law.
        \item Pseudo-reduction.
    \end{enumerate}

    Now, we proceed to deal with the second property.
    We have the following:
    \begin{equation}
        \label{eq:dispersion_intermediate}
        \begin{aligned}
            \dist(t \odot x_i, e) 
            &\stackrel{(1)}{=} \gyrnorm{ \ominus e \oplus (t \odot x_i) } \\
            &\stackrel{(2)}{=} \gyrnorm{ t \odot x_i } \\
            &= \norm{ t \rielog_e (x_i) }_e\\
            &= |t| \norm{\rielog_e (x_i) }_e\\
            &= |t| \gyrnorm{ x_i }\\
            &\stackrel{(3)}{=} |t| \gyrnorm{ \ominus e \oplus x_i }\\
            &= |t| \dist( e, x_i)\\
            &\stackrel{(4)}{=} |t| \dist(x_i, e)\\
    \end{aligned}
    \end{equation}
    The above follows from the following.
    \begin{enumerate}[label=(\arabic*)]
        \item Symmetry of gyrodistance (as geodesic distance).
        \item $\ominus e = e$.
        \item $x_i = \ominus e \oplus x_i$.
        \item Symmetry of gyrodistance (as geodesic distance).
    \end{enumerate}
    The last equation in \cref{eq:dispersion_intermediate} indicates the homogeneity of dispersion from $e$.
\end{proof}

\linkofproof{prop:fast_bracket_grassmannian}
\subsection{Proof of \cref{prop:fast_bracket_grassmannian}}

We first review a fast and stable algorithm for the ONB Grassmannian logarithm \citep[Algorithm 5.3]{bendokat2024grassmann}, and the calculation of Grassmannian logarithm under the projector perspective by the ONB Grassmannian logarithm \citep[Proposition 3.12]{nguyen2024matrix}.

\begin{algorithm}[tbp]
\caption{ONB Grassmann logarithm\citep[Algorithm 5.3]{bendokat2024grassmann}}
\label{alg:modgrasslog}
\KwIn{$U, Y \in \grasonb{p,n}$ are Stiefel representatives under ONB perspective.}
$QSR^T \stackrel{\mathrm{SVD}}{:=} Y^T U$ with $S$ in ascending order, and $Q$ and $R$ column-wisely flipped accordingly\;
$\hat{S} = \sqrt{I_n -S^2}$\;
$\Delta = (I_n - UU^\top)YQ \frac{\arcsin(\hat{S})}{\hat{S}} R^T$\;
\BlankLine
\KwOut{$\rielog_U(Y)=\Delta$}
\end{algorithm}

\cref{alg:modgrasslog} reviews a fast and stable algorithm for the Grassmannian Riemannian logarithm under the ONB perspective $\grasonb{p,n}$.
The vanilla Riemannian logarithm in \cref{tab:riem-gyro-operators-grass-onb} requires an $n \times p$ SVD and a $p \times p$ matrix inverse, while \cref{alg:modgrasslog} only requires an $p \times p$ SVD.
Therefore, \cref{alg:modgrasslog} is more efficient than the vanilla logarithm.
Besides, \cref{alg:modgrasslog} can also return a unique tangent vector when $Y$ is in the cut locus of $U$. For more details, please refer to \citet[Section 5.2]{bendokat2024grassmann}.

As the projector perspective is isometric to the ONB perspective, the Grassmannian logarithm under the projector perspective can be calculated by the ONB Grassmannian logarithm \citep[Proposition 3.12]{nguyen2024matrix}.

\begin{proposition}[\citep{nguyen2024matrix}] \label{app:prop:grass_logarithm_onb_pp}
    Given any $P, Q \in \graspp{p,n}$ with $U=\pi^{-1}(P)$ and $V=\pi^{-1}(Q)$, the Riemannian logarithm $\widetilde{\rielog}_{P}(Q)$ on $\graspp{p,n}$ is given as
    \begin{equation*}
        \widetilde{\rielog}_{P}(Q) = \pi_{*,U} \left( \rielog_U V \right),
    \end{equation*}
    where $\rielog$ is the Riemannian logarithm under the ONB perspective, $\pi_{*,U}: T_{U} \grasonb{p,n} \rightarrow T_{P} \graspp{p,n}$ is the differential map of $\pi$ at $U$, which is defined as
    \begin{equation*}
        \pi_{*,U} (\Delta) = \Delta U^\top + U \Delta^\top, \forall \Delta \in  T_{U} \grasonb{p,n}. 
    \end{equation*}
\end{proposition}

Now, we begin to present the proof. 
\begin{proof}[Proof of \cref{prop:fast_bracket_grassmannian}]
    We first show the expression for $\rielog_{\idonb}$ and $\widetilde{\rielog}_{\idpp}$.
    
    First note the following:
    \begin{equation*}
        (I_n - \idonb \idonb^\top) = 
        \left( \begin{array}{cc} 
            0 & 0 \\
            0 & I_{n-p}
        \end{array} \right),
    \end{equation*}
    \begin{equation*}
        \begin{aligned}
             U^\top \idonb 
            &= \left(U_1^\top,U_2^\top \right)
            \left(\begin{array}{c}
                    I_p \\
                    0 
                \end{array} \right) \\
            &= U_1^\top,
        \end{aligned}
    \end{equation*}
    By the above two equations, the ONB Grassmannian logarithm at $\idonb$ is 
    \begin{equation}
        \label{eq:gras_log_i_onb}
        \begin{aligned}
            \rielog_{\idonb} (U)
            &=  \left( \begin{array}{cc} 
                0 & 0 \\
                0 & I_{n-p}
            \end{array} \right) 
            \left(\begin{array}{c}
                U_1 \\
                U_2
            \end{array} \right)
            Q \frac{\arcsin(\hat{S})}{\hat{S}} R^T \text{ (\cref{alg:modgrasslog})}\\
            &= \left(\begin{array}{c}
                0 \\
                U_2 Q \frac{\arcsin(\hat{S})}{\hat{S}} R^T
            \end{array} \right) \\
            &= \left(\begin{array}{c}
                0 \\
                \widetilde{U}_2
            \end{array} \right),
        \end{aligned}
    \end{equation}
    where $QSR^T \stackrel{\mathrm{SVD}}{:=} U_1^\top$ with $S$ in ascending order, and $Q$ and $R$ column-wisely flipped accordingly, and $\hat{S} = \sqrt{I_n -S^2}$. 

    For $\widetilde{\rielog}_{\idpp}$, we have
    \begin{equation*}
        \begin{aligned}
            \widetilde{\rielog}_{\idpp}(UU^\top) 
            &\stackrel{(1)}{=} \pi_{*,\idonb} \left( \rielog_{\idonb} (U) \right) \\
            &\stackrel{(2)}{=} \pi_{*,\idonb} \left( \left(\begin{array}{c}
                0 \\
                \widetilde{U}_2
            \end{array} \right) \right) \\
            &\stackrel{(3)}{=} \left(\begin{array}{cc}
                0 & \widetilde{U}_2^\top \\
                \widetilde{U}_2 & 0
            \end{array} \right).
        \end{aligned}
    \end{equation*}
    The above derivation comes from the following.
    \begin{enumerate}[label=(\arabic*)]
        \item 
        \cref{app:prop:grass_logarithm_onb_pp}
        \item
        \cref{eq:gras_log_i_onb}
        \item
        For any $\Delta = (\Delta_1^\top,\Delta_2^\top)^{\top} \in T_{\idonb}\grasonb{p,n}$, where $\Delta_1$ is $p \times p$, we have the following
        \begin{equation*}
            \begin{aligned}
                 \pi_{*,\idonb} \left( 
                 \left( \begin{array}{c}
                      \Delta_1 \\
                      \Delta_2
                 \end{array}
                 \right)\right) 
                 &=  
                 \left( \begin{array}{c}
                      \Delta_1 \\
                      \Delta_2
                 \end{array}
                 \right) \left(I_p,0 \right) + 
                 \left( \begin{array}{c}
                      I_p \\
                      0
                 \end{array}
                 \right) \left(\Delta_1^\top, \Delta_2^\top \right) \\
                 &= \left( \begin{array}{cc}
                      \Delta_1 & 0  \\
                      \Delta_2 & 0
                 \end{array} \right) + \left( \begin{array}{cc}
                      \Delta_1^\top & \Delta_2^\top  \\
                      0 & 0
                 \end{array} \right) \\
                 &=\left( \begin{array}{cc}
                      \Delta_1 + \Delta_1^\top & \Delta_2^\top  \\
                      \Delta_2 & 0
                 \end{array} \right).
            \end{aligned}
        \end{equation*} 
    \end{enumerate}
    Combining all the above results together, we have the following:
    \begin{equation*}
        \begin{aligned}
            [\overline{U U^\top},\idpp] 
            &= \left[ \widetilde{\rielog}_{\idpp}(UU^\top), \idpp \right]\\
            &= \left[ \left(\begin{array}{cc}
                0 & \widetilde{U}_2^\top \\
                \widetilde{U}_2 & 0
            \end{array} \right), \idpp \right]\\
            &= \left(\begin{array}{cc}
                0 & \widetilde{U}_2^\top \\
                \widetilde{U}_2 & 0
            \end{array} \right)
            \left( \begin{array}{cc} 
            I_{p} & 0 \\
            0 & 0
        \end{array} \right) - 
        \left( \begin{array}{cc} 
            I_{p} & 0 \\
            0 & 0
        \end{array} \right)
        \left(\begin{array}{cc}
                0 & \widetilde{U}_2^\top \\
                \widetilde{U}_2 & 0
            \end{array} \right) \\
        &=\left(
        \begin{array}{cc}
             0 & -\widetilde{U}_2^T \\
             \widetilde{U}_2^T & 0
        \end{array} \right).
        \end{aligned}
    \end{equation*}
\end{proof}

\linkofproof{prop:stereographic_gyro_from_riemannian}
\subsection{Proof of \cref{prop:stereographic_gyro_from_riemannian}}
\begin{proof}
    Given $v \in T_\Rzero \stereo{n}$, we have the following:
    \begin{align*}
        \lambda^K_{\Rzero} &= 2, \\
        \rielog _{\Rzero}(y) 
        &= \tank ^{-1} \left( \sqrt{|K|} \norm{y} \right) \frac{y}{ \sqrt{|K|} \norm{y}}, \\
        \pt{\Rzero}{x} (v) 
        &= \frac{\lambda_{\Rzero} ^K}{\lambda_{x}^K} \gyr[y,-\Rzero] (v) = \frac{2}{\lambda_{x}^K} v, \\
        \rieexp _{\Rzero}(v)
        &= \tank \left(\sqrt{|K|} \norm{v}\right) \frac{v}{\sqrt{|K|} \norm{v}}.
    \end{align*}
    Therefore, we have
    \begin{equation*}
        \begin{aligned}
            \rieexp_{x}\left(\pt{\Rzero}{x} \left(\rielog _{\Rzero}(y)\right)\right)
            &= \rieexp_{x}\left( \frac{2}{\sqrt{|K|} \lambda_{x}^K } \tank ^{-1} \left( \sqrt{|K|} \norm{y} \right) \frac{y}{\norm{y}} \right) \\
            &= x \stoplus y,
        \end{aligned}
    \end{equation*}
    which implies
    \begin{equation*}
        \begin{aligned}
            \rieexp _{\Rzero} \left( t \rielog _{\Rzero} (x) \right) = t \stodot x.
        \end{aligned}
    \end{equation*}
\end{proof}

\linkofproof{thm:stereographic_gyro}
\subsection{Proof of \cref{thm:stereographic_gyro}}
\begin{proof}
As discussed in the main paper, we always assume the gyro operations are well-defined. We only need to show the case of $K > 0$. Let $x, y, z, w$ be any vectors in $\stereo{n}$, and $s, t \in \bbRscalar$ be real scalars. 

We first notice the gyroaddition is a linear combination:
\begin{equation*} 
    x \oplus_K y = \frac{(1 - 2K \inner{x}{y} - K \norm{y}^2)x + (1 + K \norm{x}^2)y}{1 - 2K \inner{x}{y} + K^2 \norm{x}^2 \norm{y}^2}
    = \frac{A x + B y}{D},
\end{equation*}
where
\begin{equation}
    \label{app:eq:gyro_addition_setero_rewritten}
    \begin{aligned}
    A &= 1 - 2K \inner{x}{y} - K \norm{y}^2, \\
    B &= 1 + K \norm{x}^2, \\
    D &= 1 - 2K \inner{x}{y} + K^2 \norm{x}^2 \norm{y}^2.
    \end{aligned}
\end{equation}
Recalling \cref{eq:stereographic_gyration}, the gyration $\gyr[x, y](z)$ is also a linear expression:
\begin{equation*}
    \gyr[x, y](z) = z + f_1 \cdot x + f_2 \cdot y.
\end{equation*}

\mypara{Axioms G1-G3.}
Noting that $e=\Rzero$ and $\stominus x = -x$, G1 and G2 can be immediately verified. 
The left gyroassociative law follows from the definition of gyration \citep[Equation 36]{bachmann2020constant}:
\begin{equation*}
    \gyr[x, y] (z)= -\left(x \stoplus y\right) \stoplus\left(x \stoplus\left(y \stoplus z\right)\right).
\end{equation*}
To confirm that any gyration is an automorphism of $(\stereo{n}, \stoplus)$, we verify the the following identity using symbolic computation:
\begin{equation} \label{app:eq:stereographic_gyration_identity}
    \gyr[x,y](w \stoplus z) = \gyr[x,y](w) \stoplus \gyr[x,y](z).
\end{equation}

Expanding all necessary inner products (e.g., $\inner{x}{w \stoplus z}$, $\inner{y}{w \stoplus z}$) and norms in terms of $\inner{x}{w}$, $\inner{x}{z}$, etc., we express the both sides of \cref{app:eq:stereographic_gyration_identity} as linear combinations:
\begin{align*}
    \gyr[x,y](w \stoplus z) = f_1 x + f_2 y + f_3 w + f_4 z, \\
    \gyr[x,y](w) \stoplus \gyr[x,y](z) = f_1' x + f_2' y + f_3' w + f_4' z.
\end{align*}
We use \texttt{SymPy} to compare the coefficients of $x$, $y$, $w$, and $z$ on both sides. The above is exposed in \verb|stereographic_gyr_automorphism.py|.

\mypara{(G4) Left Reduction Law.}
Similarly, we expand $\gyr[x \stoplus z,y](w)$ and $\gyr[x,y](w)$ and compare the coefficients, as implemented in \verb|stereographic_left_reduction.py|.

\mypara{Gyrocommutative Law.}
This has been verified by \citet[Lemma 11]{bachmann2020constant}.

\mypara{(V1) Identity Scalar Multiplication.}
This can be directly verified by definition:
\begin{equation*}
    t \stodot x= \frac{\tan \left(t \tan ^{-1}(\sqrt{-K}\norm{x})\right)}{\sqrt{-K}} \frac{x}{\norm{x}}, \forall x \neq 0.
\end{equation*}

\mypara{(V2) Scalar Distributive Law.}
We want to verify
\begin{equation} \label{app:eq:stereographic_axiom_v2}
    (s + t) \stodot x = (s \stodot x) \stoplus (t \stodot x).
\end{equation}
We only need to show the case of $x \neq \Rzero$.
Let \( \theta = \tan^{-1}(\sqrt{K} \, \norm{x} ) \). Then scalar multiplication is simplified as
\begin{equation*}
    t \stodot x = \frac{\tan(t \theta)}{\sqrt{K} \norm{x}} x.
\end{equation*}
Using symbolic computation (see \verb|stereographic_gyr_v2.py|), we obtain the following w.r.t. \cref{app:eq:stereographic_axiom_v2}:
\begin{align*}
    \text{LHS} &= \frac{\tan\left((s + t)\theta\right)}{\sqrt{K} \norm{x}}, \\
    \text{RHS} &= \frac{\tan(s\theta) + \tan(t\theta)}{\sqrt{K} \norm{x} \left( 1 - \tan(s\theta)\tan(t\theta) \right)}.
\end{align*}
The identity follows from the tangent addition formula:
\begin{equation*}
    \tan((s + t)\theta) = \frac{\tan(s\theta) + \tan(t\theta)}{1 - \tan(s\theta)\tan(t\theta)}.
\end{equation*}

\mypara{(V3) Scalar Associative Law.}
This can be directly verified by definition.

\mypara{(V4) Gyroautomorphism.}
We now verify
\begin{equation} \label{app:eq:stereographic_axiom_v4}
    \gyr[x,y](t \stodot z) = t \stodot \gyr[x,y](z).
\end{equation}
Let us denote
\begin{equation*}
    \alpha_t^{(z)} = \frac{\tan \left(t \tan^{-1}\left(\sqrt{K} \norm{z}\right)\right)}{\sqrt{K} \norm{z}}.
\end{equation*}
By the linearity of gyration \citep[Lemma 11]{bachmann2020constant}, the left-hand side of \cref{app:eq:stereographic_axiom_v4} becomes
\begin{equation*}
    \text{LHS} = \alpha_t^{(z)} \gyr[x,y](z),
\end{equation*}
while the right-hand side reads
\begin{equation*}
    \text{RHS} = \alpha_t^{\gyr[x,y](z)} \gyr[x,y](z).
\end{equation*}
Since $\alpha_t^{(\cdot)}$ depends only on the norm of its argument, it suffices to show
\begin{equation*}
    \norm{z} = \norm{\gyr[x,y](z)},
\end{equation*}
which holds as proven by \citet[Lemma 11, iv)]{bachmann2020constant}.

\mypara{(V5) Identity Gyroautomorphism.}
In each of the three special cases when (i) $x=\Rzero$, or (ii) $y=\Rzero$, or (iii) $x$ and $y$ are parallel in $\mathbb{V}, x \| y$, we have
\begin{align*}
    \gyr[\Rzero, x] (z) &\stackrel{(1)}{=} z, \\
    \gyr[x, \Rzero] (z) &\stackrel{(2)}{=} z, \\
    \gyr[x, y] (z) &\stackrel{(3)}{=} z, \quad x \| y,
\end{align*}
where (1--3) come from $Ax +By = 0$ in \cref{eq:stereographic_gyration}. Therefore, we have 
\begin{equation*}
    \gyr[s \stodot x, t \stodot x] = \gyr[\alpha_s^{(x)} x, \alpha_t^{(x)} x] = \id.
\end{equation*}
Note that (1--2) are also implied by the first gyrogroup theorem \citep[Theorem 2.10]{ungar2022analytic}.
\end{proof}

\linkofproof{thm:gyroinvariance_stereo}
\subsection{Proof of \cref{thm:gyroinvariance_stereo}}
This proof largely follows the one for \cref{thm:gyroinvariance}.
\begin{proof}
    \mypara{Norm Invariance under Gyrations.}
    By \citet[Lemma~11]{bachmann2020constant}, any gyration preserves the Euclidean norm: 
    \begin{equation*}
        \| \gyr[a,b](x) \| = \| x \|, \quad \forall x \in \stereo{n}.
    \end{equation*}
    For the gyronorm, we further have
    \begin{equation*}
        \begin{aligned}
            \gyrnorm{\gyr[a,b] x}
            &= 2\| \rielog _{\Rzero}  (\gyr[a,b] x )\|_{\Rzero} \\
            &= \tfrac{2}{\sqrt{|K|}} \tan _K^{-1}\!\left(\sqrt{|K|}\,\|\gyr[a,b] x \|\right) \\
            &= \tfrac{2}{\sqrt{|K|}} \tan _K^{-1}\!\left(\sqrt{|K|}\,\|x\|\right) \\
            &= \gyrnorm{x},
        \end{aligned}
    \end{equation*}
    with $\tan_K=\tanh$ for $K<0$ and $\tan_K=\tan$ for $K>0$.
    
    \mypara{Isometry of Left Gyrotranslation and Gyroinverse.}
    As shown in \cref{thm:stereographic_gyro}, $\stereo{n}$ forms a gyrocommutative gyrogroup. By \cref{thm:gyroisometries}, the left gyrotranslation and gyroinverse are gyroisometries.
\end{proof}

\linkofproof{prop:calmk_scalar_prod_inv}
\subsection{Proof of \cref{prop:calmk_scalar_prod_inv}}

\begin{proof}
    \mypara{Non-Singular Cases.}
    Denoting $\norm{x}_s = \norm{x_s}, \forall x \in \calMK{n}$, we have
    \begin{align*}
    \left( t  \log_{\MKzero} x \right)_s 
    &= t  \frac{\cosk^{-1}(\sqrt{|K|} x_t)}{\sqrt{|K|} \norm{x_s}} x_s, \\
    \left\| t  \log_{\MKzero} x \right\|_s 
    &= t  \frac{\cosk^{-1}(\sqrt{|K|} x_t)}{\sqrt{|K|}}, \\
    \sqrt{-K}  \left\| t  \log_{\MKzero} x \right\|_s 
    &= t \cosk^{-1}(\sqrt{|K|} x_t),
    \end{align*}
    Putting the above into $\rieexp_{\MKzero}$ in \cref{tab:riem-operator-radius}, one can obtain the result.

    For $t=-1$, we have 
    \begin{equation*}
        \begin{aligned}
        -1 \MKominus x 
        &= 
        \frac{1}{\sqrt{|K|}} 
        \begin{bmatrix}
        \cosk \left( - \cosk^{-1}(\sqrt{|K|} x_t) \right) \\
        \frac{\sink \left( - \cosk^{-1}(\sqrt{|K|} x_t) \right)}{\norm{x_s}}  x_s
        \end{bmatrix} \\
        &\stackrel{(1)}{=}
        \begin{bmatrix}
        x_t \\
        - \frac{\sink \left(\cosk^{-1}(\sqrt{|K|} x_t) \right)}{\sqrt{|K|}\norm{x_s}} x_s
        \end{bmatrix},
        \end{aligned}
    \end{equation*}
    where (1) comes from $\cosk(- \theta) = \cosk(\theta)$ and $\sink(-\theta)=-\sink(\theta)$. The rest is to show $\frac{\sink \left(\cosk^{-1}(\sqrt{|K|} x_t) \right)}{\sqrt{|K|}\norm{x_s}} = 1$:
    \begin{equation*}
        \frac{\sink \left(\cosk^{-1}(\sqrt{|K|} x_t) \right)}{\sqrt{|K|}\norm{x_s}} 
        \stackrel{(1)}{=} \frac{\sqrt{ \sign(K) (1- |K| x^2_t) }}{\sqrt{|K|} \norm{x_s}}
        \stackrel{(2)}{=}1.
    \end{equation*}
    The derivation above is based on the following:
    \begin{enumerate}[label=(\arabic*)]
        \item
        $\cosk ^2 (\theta) + \sign(K) \sink ^2 (\theta) = 1$ implies
        \begin{equation*}
            \sink (\theta) = \sqrt{\sign(K) (1 - \cosk ^2 (\theta))}, \; \forall \theta>0.
        \end{equation*}
        Considering $\cosk^{-1}(\theta) \geq 0$ for any $\theta \in \operatorname{dom}(\cosk^{-1}(\cdot))$, we have (1).
        \item 
        \begin{equation*}
            \begin{aligned}
                &\Knorm{x} = \sign(K) x_t^2 + \norm{x_s}^2 = \frac{1}{K} \\
                & \Rightarrow K \norm{x_s}^2 = 1 - |K| x_t^2 \\
                & \Rightarrow |K| \norm{x_s}^2 = \sign(K) (1 - |K| x_t^2).
            \end{aligned}
        \end{equation*}

    \end{enumerate}

    \mypara{Equivalence in the Singular Cases.}
    By \cref{eq:iso_calMK_to_stereo},
    \begin{equation*}
    \sqrt{K} \norm{u} = \frac{\sqrt{K}  \norm{x_s}}{1+\sqrt{K} x_t}.
    \end{equation*}
    Writing $\theta=\cos^{-1}(\sqrt{K} x_t)$ or $\sqrt{K} x_t=\cos\theta$, the sphere constraint gives $\sqrt{K}  \norm{x_s}=\sin\theta$. Hence
    \begin{equation*}
    \sqrt{K} \norm{u} = \frac{\sin\theta}{1+\cos\theta} = \tan \frac{\theta}{2},
    \quad \Rightarrow 
    \tan^{-1}\!\left(\sqrt{K} \norm{u}\right)=\frac{\theta}{2}.
    \end{equation*}
    Therefore (i) $\iff$ (ii) since $t \tan^{-1}(\sqrt{K}\norm{u})=\frac{t\theta}{2}$.
    
    Next, we use the closed form of gyromultiplication \cref{eq:singular-case-radius}. For $K>0$, the gyromultiplication reads
    \begin{equation*}
    t \MKodot x = \frac{1}{\sqrt{K}}
    \begin{bmatrix}
    \cos\!\left(t\theta\right) \\
    \dfrac{\sin(t\theta)}{ \norm{x_s}}x_s
    \end{bmatrix}.
    \end{equation*}
    If (ii) holds, then $\sin(t\theta)=0$ and $\cos(t\theta)=-1$, whence $t \MKodot x=[-1/\sqrt{K},0]^\top=-\MKzero$.
\end{proof}

\linkofproof{prop:calmk_gyroaddition}
\subsection{Proof of \cref{prop:calmk_gyroaddition}}
\begin{proof}
We denote $x\MKoplus y=[z_t,z_s^\top]^\top$. As the results are trivial under $x=\MKzero$ or $y=\MKzero$, we assume $x \neq \MKzero$ and $y \neq \MKzero$ in the following.

\mypara{Non-Singular Cases.}
We first consider the non-singular case: i) $K<0$; ii) $K>0$, $x,y \neq \pm \MKzero$, and $u \neq \frac{v}{K \norm{v}^2}$. Since gyroadditions on $\calMK{n}$ and $\stereo{n}$ are both defined by \cref{eq:gyro_addtion}, we have the following under isometries \citep[Lemma 2.2]{nguyen2023building}:
\begin{equation*}
x\MKoplus y = \isoSTMK{n} \left( \isoMKST{n}(x)\ \stoplus \isoMKST{n}(y) \right).
\end{equation*}

Following \cref{app:eq:gyro_addition_setero_rewritten}, we rewrite the gyroaddition on the stereographic model as
\begin{equation*}
u\stoplus v=\frac{(1-2K\langle u,v\rangle-K\|v\|^2)u+(1+K\|u\|^2)v}{1-2K\langle u,v\rangle+K^2\|u\|^2\|v\|^2}= \frac{A u + B v}{\Delta}.
\end{equation*}
where
\begin{equation*}
A=1-2K\langle u,v\rangle-K\|v\|^2,\quad B=1+K\|u\|^2,\quad \Delta=1-2K\langle u,v\rangle+K^2\|u\|^2\|v\|^2.
\end{equation*}
Using $\langle u,v\rangle=s_{xy}/(ab)$, $\|u\|^2=n_x/a^2$, $\|v\|^2=n_y/b^2$, one obtains
\begin{equation}\label{app:eq:ABD-simplified}
A=\frac{ab^2-2Kb\,s_{xy}-Ka\,n_y}{ab^2},\quad B=\frac{a^2+K n_x}{a^2},\quad \Delta=\frac{D}{a^2 b^2}.
\end{equation}
Hence
\begin{equation}\label{app:eq:w}
\stereo{n} \ni w=u\stoplus v=\frac{A u+B v}{\Delta}
=\frac{1}{\Delta} \left(\frac{A}{a}x_s+\frac{B}{b}y_s \right).
\end{equation}

Apply $\isoSTMK{n}$ to $w$:
\begin{equation}\label{app:eq:ztzs-pre}
z_t=\frac{1}{\sqrt{|K|}}\frac{1-K\|w\|^2}{1+K\|w\|^2},\qquad
z_s=\frac{2w}{1+K\|w\|^2}.
\end{equation}
A direct expansion (\texttt{radius\_gyroaddition.py}) yields
\begin{equation*}
\|w\|^2=\frac{A^2\|u\|^2+B^2\|v\|^2+2AB\langle u,v\rangle}{\Delta^2}=\frac{N}{D}.
\end{equation*}
Substituting the above into \cref{app:eq:ztzs-pre} gives
\begin{equation}\label{app:eq:zt}
z_t=\frac{1}{\sqrt{|K|}}\frac{1-K N/D}{1+K N/D}=\frac{1}{\sqrt{|K|}}\frac{D-KN}{D+KN}.
\end{equation}
For $z_s$, using \cref{app:eq:w}, \cref{app:eq:ABD-simplified} and $1+K\|w\|^2=(D+KN)/D$,
\begin{equation*}
    \begin{aligned}
        z_s
        &=\frac{2}{1+K N/D}\cdot\frac{1}{\Delta}\left(\frac{A}{a}x_s+\frac{B}{b}y_s\right) \\
        &= \frac{2}{D+KN}\left((A\,ab^2)x_s+(B\,a^2 b)y_s\right) \\
        &= \frac{2 \left( A_s x_s + A_y y_s \right)}{D+KN}.
    \end{aligned}
\end{equation*}

\mypara{Gyroaddition in Singular Cases ($K>0$).}
We show that the definition \cref{eq:gyroadd_calmk} indeed returns $-\MKzero$ in this case.

\noindent \emph{Step 1: \(\rielog_{\MKzero}(y)\).}
Write $R=1/\sqrt{K}$ and choose the polar angle $\theta\in(0,\pi)$ so that
\begin{equation}\label{app:eq:x_param}
x=\begin{bmatrix}R\cos\theta\\ R\sin\theta\,\hat s\end{bmatrix},\qquad
y=\begin{bmatrix}-R\cos\theta\\ R\sin\theta\,\hat s\end{bmatrix},\qquad
\hat s=\frac{x_s}{\|x_s\|}.
\end{equation}
Using $\rielog_{\MKzero}$ in \cref{tab:riem-operator-radius},
\begin{equation*}
\rielog_{\MKzero}(y)
=\begin{bmatrix}
0\\ \dfrac{\cos^{-1}(\sqrt{K}\,y_t)}{\sqrt{K}\,\|y_s\|}\,y_s
\end{bmatrix}
=
\begin{bmatrix}
0\\ (\pi-\theta)\,R\,\hat s
\end{bmatrix}.
\end{equation*}

\noindent\emph{Step 2: Parallel transport.}
With $1+\sqrt{K}\,x_t=1+\cos\theta\neq 0$ (since $x\neq-\MKzero$) and $\pt{\MKzero}{x}$ in \cref{tab:riem-operator-radius}, we have
\begin{equation}\label{app:eq:PT_compute}
\pt{\MKzero}{x}\left(\rielog_{\MKzero}(y)\right)
= \begin{bmatrix}
0\\ (\pi-\theta)R\,\hat s
\end{bmatrix}
-\frac{K\langle x_s,(\pi-\theta)R\,\hat s\rangle}{1+\sqrt{K}\,x_t}
\begin{bmatrix}
x_t+\tfrac{1}{\sqrt{K}}\\ x_s
\end{bmatrix}.
\end{equation}
Using $\langle x_s,\hat s\rangle=\|x_s\|=R\sin\theta$, $K R^2=1$, and $x_t=R\cos\theta$, the scalar factor in \cref{app:eq:PT_compute} equals
\[
\frac{K(\pi-\theta)R\langle x_s,\hat s\rangle}{1+\sqrt{K}\,x_t}
=\frac{(\pi-\theta)\sin\theta}{1+\cos\theta}.
\]
A short simplification then yields the compact form
\begin{equation}\label{app:eq:w_tangent}
w = \pt{\MKzero}{x}\left(\rielog_{\MKzero}(y)\right)
= R(\pi-\theta)
\begin{bmatrix}
-\sin\theta\\ 
\cos\theta\,\hat s
\end{bmatrix}.
\end{equation}
Note that $\|w\|_x=R(\pi-\theta)$, so with $\alpha=\sqrt{K}\,\|w\|_x$ we have $\alpha=\pi-\theta$.

\noindent\emph{Step 3: Exponential at \(x\).}
By \cref{tab:riem-gyro-operators-grass-onb},
\begin{equation}\label{app:eq:Expx_standard}
\rieexp_x(w) \;=\; \cos(\alpha)\,x + \frac{\sin(\alpha)}{\alpha}\,w,
\qquad \alpha=\sqrt{K}\,\|w\|_x,
\end{equation}
and here $\cos(\alpha)=\cos(\pi-\theta)=-\cos\theta$, $\sin(\alpha)=\sin(\pi-\theta)=\sin\theta$.
Substituting \cref{app:eq:x_param} and \cref{app:eq:w_tangent} into \cref{app:eq:Expx_standard},
\begin{equation*}
\rieexp_x(w)
= R\!\left[
-\cos\theta\begin{bmatrix}\cos\theta\\ \sin\theta\,\hat s\end{bmatrix}
+ \sin\theta\begin{bmatrix}-\sin\theta\\ \cos\theta\,\hat s\end{bmatrix}
\right]
= R\begin{bmatrix}-1\\ 0\end{bmatrix}
= -\,\MKzero.
\end{equation*}
We conclude that $x\MKoplus y=-\MKzero$ in the singular configuration.

\mypara{Equivalence in Singular Cases ($K>0$).}
Recall the isometry between $\calMK{n}$ and $\stereo{n}$, we rewrite $u=\tfrac{x_s}{a}$, $v=\tfrac{y_s}{b}$.

\emph{(i) $\Rightarrow$ (ii).}
Write $\alpha=K \norm{v}^2 >0$. From $u=\frac{v}{\alpha}$ and \cref{eq:iso_stereo_to_calMK} we get
\begin{equation*}
    \begin{aligned}
        y_t
        &=\frac{1}{\sqrt{K}}\frac{1-\alpha}{1+\alpha}, \\
        x_t
        &=\frac{1}{\sqrt{K}}\frac{1-\frac{1}{\alpha}}{1+\frac{1}{\alpha}}
        =-\,\frac{1}{\sqrt{K}}\frac{1-\alpha}{1+\alpha}=-y_t, \\
        x_s
        &=\frac{2u}{1+K\|u\|^2}=\frac{2v/\alpha}{1+K\|v\|^2/\alpha^2}=\frac{2v}{1+\alpha}=y_s, 
    \end{aligned}
\end{equation*}
Hence $x_s=y_s$ and $x_t=-y_t$.

\emph{(ii) $\Rightarrow$ (iii).}
Under $x_s=y_s$ and $x_t=-y_t$, set $n= \norm{x_s}^2 = \norm{y_s}^2$ and note
$s_{xy}=n$, $n_x=n_y=n$. Then,
\begin{equation*}
D=a^2b^2-2Kab\,s_{xy}+K^2n_xn_y=(ab-Kn)^2.
\end{equation*}
On the sphere $\sphere{n}$, we have the constraint $x_t^2+ \norm{x_s}^2 = \nicefrac{1}{K}$. Since $y_t=-x_t$ and
$\norm{y_s}^2=\norm{x_s}^2=n$,
\begin{equation*}
\begin{aligned}
    ab
    &=(1+\sqrt{K}\,x_t)(1+\sqrt{K}\,y_t) \\
    &=(1+\sqrt{K}\,x_t)(1-\sqrt{K}\,x_t) \\
    &=1-Kx_t^2 \\
    &=1-K\Bigl(\frac{1}{K}-n\Bigr)=Kn.
\end{aligned}
\end{equation*}
Therefore $D=(ab-Kn)^2=0$.

\emph{(iii) $\Rightarrow$ (i).}
This can by obtained by the Cauchy-Schwarz’s inequality as \citet[Appendix C.2.1.]{bachmann2020constant}.

\mypara{Consequences in Singular Cases ($K>0$).}
Under (ii) we also have $a+b=2$, so
\begin{equation*}
N=a^2n+2ab\,n+b^2n=(a+b)^2n=4n>0.
\end{equation*}
Using $D=0$ and $N>0$, \cref{app:eq:zt} gives
\begin{equation*}
z_t=\frac{1}{\sqrt{K}}\frac{-KN}{KN}=-\frac{1}{\sqrt{K}}
\end{equation*}
Besides, since for $x_s=y_s$ one checks $A_s+A_y=(a+b)(ab-Kn)=0$, \cref{app:eq:zt} yields $z_s=0$. On the other hand, $u\stoplus v = \infty$.
\end{proof}

\linkofproof{cor:isomorphism_calmk_stereo}
\subsection{Proof of \cref{cor:isomorphism_calmk_stereo}}
\begin{proof}
    This has already been implied in the proof of \cref{prop:calmk_scalar_prod_inv,prop:calmk_gyroaddition}.
\end{proof}

\linkofproof{thm:calmk_gyrovector}
\subsection{Proof of \cref{thm:calmk_gyrovector}}

\begin{proof}
    The gyrovector space over the stereographic model is also defined by \cref{eq:gyro_addtion,eq:gyro_scalar_product}:
    \begin{equation*}
        \begin{aligned}
            x \stoplus y &= \rieexp_{x}\left(\pt{\Rzero}{x} \left(\rielog _{\Rzero}(y)\right)\right),\\
            r \stoplus x &= \rieexp_{\Rzero} \left( \rielog _{\Rzero}(x)\right),\\
        \end{aligned}
    \end{equation*}
    with $x, y \in \stereo{n}$ and $r \in \bbRscalar$. Noting that $\pi_{\calMK{n} \to \stereo{n}}(\MKzero)=\Rzero$. As $(\stereo{n},\Moplus, \Modot)$ is a gyrovector space, $(\calMK{n},\MKoplus, \MKominus)$ is also a gyrovector space \citep[Theorem 2.4]{nguyen2023building}.
\end{proof}

\linkofproof{props:klein_poincare_isometry_differentials}
\subsection{Proof of \cref{props:klein_poincare_isometry_differentials}} 

\begin{proof}
    \mypara{Isometries:}
    The Beltrami--Klein model is isometric to the hyperboloid model by the following diffeomorphisms \citep[Theorem 3.7]{lee2018introduction}:
    \begin{align}
        \label{app:eq:iso_klein_to_bbh}
        \pi_{\klein{n} \to \bbh{n}}
        &: \klein{n} \ni x  \longmapsto\left(\frac{1}{\sqrt{-K} \sqrt{1+K\|x\|^2}}, \frac{x}{\sqrt{1+K \|x\|^2}}\right) \in \bbh{n},\\
        \label{app:eq:iso_bbh_to_klein}
        \pi_{\bbh{n} \to \klein{n}} 
        &: 
        \bbh{n} \ni  
        \begin{bmatrix}
        x_t \\
        x_s
        \end{bmatrix} \longmapsto \frac{x_s}{\sqrt{-K}\xi} \in \klein{n}.
    \end{align}
    
    Combining the isometries \cref{app:eq:iso_klein_to_bbh,app:eq:iso_bbh_to_klein,eq:iso_calMK_to_stereo,eq:iso_stereo_to_calMK}, one can readily obtain the isometries between Beltrami--Klein and Poincaré ball models. Now, we turn to the differential maps. Given a curve over $c(t) \in \calM$ with $c(0)=x$ and $c'(0)=v$, the differential maps can be calculated by
    \begin{equation*}
        \begin{aligned}
        \left. \frac{d \pi_{\klein{n} \to \pball{n}} (c(t))}{d t} \right|_{t=0} 
        &=  \left. \frac{d }{d t}  \frac{c(t)}{ \left( 1 + \sqrt{1 + K \|c(t)\|^2} \right)} \right|_{t=0} \\
        &=\frac{\left(1+ \sqrt{1+K\|x\|^2} \right) v - \left(1+K\|x\|^2\right)^{-\frac{1}{2}} K \inner{x}{v} x}{\left(1 + \sqrt{1+K\|x\|^2} \right)^2} \\
        &=\frac{1}{1 + \sqrt{1+K\|x\|^2}} v 
        - \frac{K \inner{x}{v}}{\left(1 + \sqrt{1+K\|x\|^2} \right)^2 \sqrt{1+K\|x\|^2}} x, \\
        \left. \frac{d \pi_{\pball{n} \to \klein{n} } (c(t))}{d t} \right|_{t=0} 
        &= \left. \frac{d }{d t} \frac{2 c(t)}{1 - K \|c(t)\|^2} \right|_{t=0}  \\
        &= \frac{2 v\left(1- K\|x\|^2\right)+4 K \inner{x}{v} x}{\left(1-K\|x\|^2\right)^2 } \\
        &= \frac{2}{\left(1-K\|x\|^2\right)}v
        + \frac{4 K \inner{x}{v} }{\left(1-K\|x\|^2\right)^2 }x.
        \end{aligned}
    \end{equation*}  

    \mypara{Homomorphism:}
    The homomorphism w.r.t. the scalar product can be readily obtained by the Riemannian isometry. Therefore, we first address it before proceeding to the addition. For simplicity, we denote $\phi=\pi_{\pball{n} \to \klein{n}}$.

    \mypara{Scalar Product: }
    As shown by \citet{ungar2022analytic}, the geodesics under the Beltrami--Klein and Poincaré ball models are
    \begin{equation*}
    \begin{aligned}
        \gamma ^{\mathbb{K}} _{\phi(x) \rightarrow \phi(y)}(t)
        &=\phi(x) \Eoplus t \Eodot \left(-\phi(x) \Eoplus \phi(y)\right) , \\
        \gamma ^{\mathbb{P}} _{ x \rightarrow y}(t)
        &=x \Moplus t \Modot \left(- x \Moplus y \right).
    \end{aligned}
    \end{equation*}
    Here, we use the fact that the gyro inverses in the Möbius and Einstein gyrovector spaces are exactly the familiar vector inverse. The above geodesics satisfy
    \begin{equation*}
        \begin{aligned}
            \phi(x \Moplus t \Modot \left(-x \Moplus y\right))=\phi(x) \Eoplus  t \Eodot \left( -\phi(x) \Eoplus \phi(x) \right).
        \end{aligned}
    \end{equation*}
    The above comes from $\phi(\gamma ^{\mathbb{P}} _{x \rightarrow y}\left( t \right)) = \gamma ^{\mathbb{K}} _{ \phi(x) \rightarrow \phi(y)} \left( t \right)$. Especially, the geodesic starting from the identity element brings 
    \begin{equation*}
        \begin{aligned}
            \phi(x \Modot t) = \phi(\gamma ^{\mathbb{P}} _{\Rzero \rightarrow x}\left( t \right)) = \gamma ^{\mathbb{K}} _{\phi(\Rzero) \rightarrow \phi(x)}\left( t \right) \stackrel{(1)}{=} \phi(x) \Eodot t, \\
        \end{aligned}
    \end{equation*}
    where (1) comes from $\phi(\Rzero)=\Rzero$. The above holds for all $x \in \pball{n}$ and $\forall t \in \bbRscalar$, as every hyperbolic geometry is geodesically complete \citep[Page 139]{lee2018introduction}.
    
    \mypara{Addition: }
    We first expand the LHS of \cref{eq:iso_addition}. Inspired by \citet[Equations 73--76]{mao2024klein}, we express the Möbius addition as $x \Moplus y = \frac{B x +C y}{A}$ by denoting
    \begin{equation*}
        \begin{aligned}
            A &= 1-2 K\langle x, y\rangle+K^2\|x\|^2\|y\|^2 ,\\
            B &= 1-2 K\langle x, y\rangle-K\|y\|^2, \\
            C &= 1+K\|x\|^2.
        \end{aligned}
    \end{equation*}
    Then, the Möbius addition is
    \begin{equation} \label{app:eq:prf_moplus_simplify}
    \begin{aligned}
        \phi(x \Moplus y)&=\frac{2 \frac{B x+C y}{A}}{1 - K\left\|\frac{B x+C y}{A}\right\|^2} \\
        & =\frac{2 A B x + 2 A C y}{A^2 - K \left\|B x + C y\right\|^2} \\
        & =\frac{2 A B x+2 A C y}{A^2 - B^2 K \left\|x\right\|^2 - C^2 K \left\|y\right\|^2 - 2 B C K \inner{x}{y} } .
    \end{aligned}
    \end{equation}

    Inspired by \citet[Equations 77]{mao2024klein}, we denote $\|x\|=a, \|y\|=b$, and $\inner{x}{y}=a b \cos(\theta)$. In this way, we can resort to the symbolic computation package \texttt{SymPy} \citep{meurer2017sympy} for the heavy algebra computation, which brings
    \begin{equation}
    \label{app:prf:eq:iso_addition_lhs_final}
    \begin{aligned}
        \phi(x \Moplus y) 
         =&  \frac{2 \left(- 2 K a b \cos{\left(\theta \right)} - K b^{2} + 1\right)}{K^{2} a^{2} b^{2} - K a^{2} - 4 K a b \cos{\left(\theta \right)} - K b^{2} + 1} x \\
         & +   \frac{2 K a^{2} + 2}{K^{2} a^{2} b^{2} - K a^{2} - 4 K a b \cos{\left(\theta \right)} - K b^{2} + 1} y.
    \end{aligned}
    \end{equation}
    
    Now we turn to the RHS of \cref{eq:iso_addition}. For any $u,v \in \klein{n}$, the Einstein addition can be rewritten as
    \begin{equation} 
    \label{app:prf:eq:iso_einstein_addition_rewritten}
        \begin{aligned}
        u \oplus_{\mathrm{E}} v 
        &=\frac{1}{1 - K\inner{u}{v}}\left(u+ \frac{1}{\gamma_{u}} v -K \frac{\gamma_{u}}{1+\gamma_{u}} \inner{u}{v} u\right) \\
        &=\frac{1 -K \frac{\gamma_{u}}{1+\gamma_{u}} \inner{u}{v} }{1 - K\inner{u}{v}} u + \frac{1}{\gamma_{u} (1 - K\inner{u}{v}) } v.
        \end{aligned}
    \end{equation}
    The gamma factor and inner product under isometry can be rewritten as 
    \begin{equation*}
        \begin{aligned}
        r_\phi(x)
        &=\frac{1}{\sqrt{1+K\|\phi(x)\|^2}} \\
        &=\frac{1}{ \sqrt{1 + K\left( \frac{2}{1-K\|x\|^2} \right )^2\|x\|^2}} \\
        &=\frac{1-K\|x\|^2}{\sqrt{\left(1-K\|x\|^2\right)^2+4 K\|x\|^2}} \\
        &\stackrel{(1)}{=} \frac{1-K\|x\|^2}{1+K\|x\|^2}, \\
        \inner{x}{y} 
        &= \frac{4 }{(1-K\|x\|^2)(1-K\|y\|^2)} \inner{x}{y}.
        \end{aligned}
    \end{equation*}
    where (1) comes from $\|x\|^2 < -\frac{1}{K} \Rightarrow K\|x\|^2 + 1 > 0$. 
    Putting the above into \cref{app:prf:eq:iso_einstein_addition_rewritten} and following the same notation as \cref{app:prf:eq:iso_addition_lhs_final}, we can obtain the following by \texttt{SymPy}:
    \begin{equation*}
        \begin{aligned}
        \phi (x) \Eoplus \phi (y) 
        =&  \frac{2 \cdot \left(2 K a b \cos{\left(\theta \right)} + K b^{2} - 1\right)}{4 K a b \cos{\left(\theta \right)} - \left(K a^{2} - 1\right) \left(K b^{2} - 1\right)} x \\
        &+  \frac{- 2 K a^{2} - 2}{4 K a b \cos{\left(\theta \right)} - \left(K a^{2} - 1\right) \left(K b^{2} - 1\right)} y,
        \end{aligned}
    \end{equation*}
    which is clearly equal to \cref{app:prf:eq:iso_addition_lhs_final}. 
\end{proof}

\linkofproof{thm:einstein_klein}
\subsection{Proof of \cref{thm:einstein_klein}} 

\begin{proof}   
    For simplicity, we denote $\phi=\pi_{\pball{n} \to \klein{n}}$. The homomorphism and bijection of $\phi$ imply the homomorphism of its inverse $\phi^{-1}$. Also note that $\phi({\Rzero})=\phi^{-1}({\Rzero})={\Rzero}$. The above brings
    \begin{equation*}
        \begin{aligned}
            x \Eoplus y 
            &= \phi \left( \phi^{-1} (x) \Moplus \phi^{-1} (x) \right) \\
            &\stackrel{(1)}{=} \phi \left( \rieexp ^\mathbb{P} _ {\phi^{-1}(x)} (\pt{{\Rzero}}{\phi^{-1}(x)} ^\mathbb{P} \rielog ^\mathbb{P} _0 (\phi^{-1}(y))) \right) \\
            &\stackrel{(2)}{=}  \rieexp ^\mathbb{K} _ y (\pt{{\Rzero}}{y} ^\mathbb{K} \rielog ^\mathbb{K} _{\Rzero} (y) ). \\
        \end{aligned}
    \end{equation*}
    The above comes from the following.
    \begin{enumerate}[label=(\arabic*)]
        \item 
        For any Poincaré vectors $u, w \in \pball{n}$, the following holds:
        \begin{equation*}
            \pt{{\Rzero}}{u} ^\mathbb{P} (v) = \rielog ^\mathbb{P} _{u} (u \Moplus \rieexp ^\mathbb{P} _{{\Rzero}} (v)) 
            \Rightarrow 
            u \Moplus w 
            = \rieexp ^\mathbb{P} _u \left(\pt{{\Rzero}}{u} ^\mathbb{P} (\rielog ^\mathbb{P} _{\Rzero} (w)) \right),
        \end{equation*}
        where $v= \rielog ^\mathbb{P} _0 (w)$ and the LHS comes from \citet[Theorem 4]{ganea2018hyperbolic};
        \item 
        It comes from the isometry:
        \begin{equation*}
            \begin{aligned}
                \rieexp ^\mathbb{P} _ x (v) 
                &=  \phi^{-1} \left( \rieexp ^\mathbb{K} _ {\phi(x)} (\phi_{*,x} (v)) \right), \\
                \rielog ^\mathbb{P} _ x (y) 
                &=  \phi_{*,x}^{-1} \left( \rielog ^\mathbb{K} _ {\phi(x)} (\phi (y)) \right), \\
                \pt{x}{y} ^\mathbb{P} (v) 
                &=  \phi_{*,y}^{-1} \left( \pt{\phi(x)}{\phi(y)} ^\mathbb{K} (\phi_{*,x} (v) ) \right).
            \end{aligned}
        \end{equation*}
    \end{enumerate}
    Similar with the gyro addition, isometry of $\phi$ implies the following w.r.t. the gyro scalar product:
    \begin{equation*}
        \begin{aligned}
            t \Eodot x 
            &= \phi \left( t \Modot \phi^{-1} (x) \right)\\
            &\stackrel{(1)}{=} \phi \left( \rieexp ^\mathbb{P} _{\Rzero} ( t \rielog ^\mathbb{P} _{\Rzero} (\phi^{-1} (x))) \right)\\
            &\stackrel{(2)}{=} \rieexp ^\mathbb{K} _{\Rzero} ( t \rielog ^\mathbb{K} _{\Rzero} (x)).
        \end{aligned}
    \end{equation*}
    The above comes from the following.
    \begin{enumerate}[label=(\arabic*)]
        \item 
        \citet[Lemma 3]{ganea2018hyperbolic};
        \item The isometry of $\phi$.
    \end{enumerate}
\end{proof}

\linkofproof{thm:riem_klein}
\subsection{Proof of \cref{thm:riem_klein}} 

\begin{proof} 
    Following \cref{thm:einstein_klein}, we denote $\phi = \pi_{\pball{n} \to \klein{n}}$ and $\psi =\pi_{\klein{n} \to \pball{n}}$. The results can be obtained by the properties of isometry and isomorphism of $\phi$.

    \mypara{Riemannian exponential and logarithmic maps at the zero vector.}
    First, we recall that the expression of $\rieexp ^{\mathbb{P}} _{\Rzero}(v)$ and $\rielog ^{\mathbb{P}} _{\Rzero}(v)$ under Poincaré ball model is the exactly \cref{eq:exp_0_klein,eq:log_0_klein} \citep[Equation 13]{ganea2018hyperbolic}.
    
    By Riemannian isometry, we have the following
    \begin{equation*}
        \begin{aligned}
            \rieexp ^\mathbb{K} _{\Rzero}(v) 
            &= \phi \left( \rieexp ^\mathbb{P} _{\Rzero}(\psi _{*,\Rzero } (v)) \right) \\
            &= \phi\left(\tanh \left(\frac{\sqrt{-K}}{2}\|v\|\right) \frac{v}{\sqrt{-K}\|v\|}\right) \\
            &= \frac{2}{1-K\| v \|^2\left(\frac{\tanh \left(\frac{\sqrt{-K}}{2}\|v\|\right)}{\sqrt{-K}\|v\|}\right)^2} \frac{\tanh \left(\frac{\sqrt{-K}}{2}\|v\|\right)}{\sqrt{-K}\|v\|} v \\
            & =\frac{2 \tanh \left(\frac{\sqrt{-K}}{2}\|v\|\right)}{1+\tanh \left(\frac{\sqrt{-K}}{2}\|v\|\right)^2} \frac{v}{\sqrt{-K}\|v\|} \\
            & \stackrel{(1)}{=}\tanh (\sqrt{-K}\|v\|) \frac{v}{\sqrt{-K}\|v\|}, \\
        \end{aligned}
    \end{equation*}
    where (1) comes from $\tanh (2 x)=\frac{2 \tanh x}{1+\tanh ^2 x}$.

    As the inverse of $\rieexp ^\mathbb{K} _{\Rzero}(v)$, $\rielog ^\mathbb{K} _{\Rzero}(x)$, therefore, shares the expression with its counterpart under the Poincaré ball model. Here, we use the properties of isometry to further validate this result:
    \begin{equation*}
        \begin{aligned}
            \rielog ^\mathbb{K} _{\Rzero}(x) 
            &= \phi _{*,0} \left( \rielog ^\mathbb{K} _{\Rzero}( \psi(x))\right) \\
            & =2 \tanh ^{-1}\left(\frac{\sqrt{-K}\|x\|}{1+\sqrt{1+K\left\|^x\right\|^2}}\right) \frac{x}{\sqrt{-K}\|x\|}.
        \end{aligned}
    \end{equation*}

    We only need to show 
    \begin{equation*}
        2 \tanh ^{-1}\left(\frac{\sqrt{-K}\|x\|}{1+\sqrt{1+K\|x\|^2}}\right)=\tanh ^{-1}(\sqrt{-K}\|x\|).
    \end{equation*}
    We denote $a = \sqrt{-K}\|x\|<1$. Both sides are 
    \begin{equation} \label{app:prf:eq:log_exp_ln_equation1}
        \begin{aligned}
            \text{LHS: }& 2 \tanh ^{-1}\left(\frac{a}{1+\sqrt{1-a^2}}\right) = \ln \left(\frac{1+\frac{a}{1+\sqrt{1-a^2}}}{1-\frac{a}{1+\sqrt{1-a^2}}}\right)=\ln \left(\frac{1+\sqrt{1-a^2}+a}{1+\sqrt{1-a^2}-a}\right), \\
            \text{RHS: }& \tanh ^{-1}(a) = \tanh \left( \frac{\sqrt{1+a}}{\sqrt{1-a}} \right) = \tanh \left( \frac{\sqrt{1-a^2}}{1-a} \right).
        \end{aligned}
    \end{equation}
    Note that the below equation holds:
    \begin{equation} \label{app:prf:eq:log_exp_ln_equation2}
        \frac{1+\sqrt{1-a^2}+a}{1+\sqrt{1-a^2}-a} = \frac{\sqrt{1-a^2}}{1-a}.
    \end{equation}

    \mypara{Geodesic distances.}
    \begin{equation*}
        \begin{aligned}
        \dist ^{\mathbb{K}} (x, y) 
        &\stackrel{(1)}{=} \dist ^{\mathbb{P}} (\psi(x), \psi(y)) \\
        &=\frac{2}{\sqrt{-K}} \tanh ^{-1}\left(\sqrt{-K}\left\|-\psi(x) \Moplus \psi(y)\right\|\right) \\
        & \stackrel{(2)}{=} \frac{2}{\sqrt{-K}} \tanh ^{-1}\left(\sqrt{-K}\left\|\psi\left(-x \Eoplus y\right)\right\|\right) \\
        & =\frac{2}{\sqrt{-K}} \tanh ^{-1}\left(\sqrt{-K} \frac{ \norm{-x \Eoplus y}}{1+\sqrt{1+K\left\|-x \Eoplus y\right\|^2}} \right),
        \end{aligned}
    \end{equation*}
    where (1) comes from the isometry, while (2) comes from the isomorphism.

    \mypara{Exponential maps.}
    \begin{equation*}
        \begin{aligned}
        \rieexp _x ^{\mathbb{K}} (v) 
        & \stackrel{(1)}{=} \psi^{-1} \left(\rieexp_{\psi(x)} ^{\mathbb{P}} \left( \psi_{*, x}(v) \right)\right) \\
        & \stackrel{(2)}{=} \psi^{-1} \left( \psi(x) \Moplus \rieexp _{\Rzero}\left(\frac{\lambda_{\psi(x)}^K}{2} \psi_{*, x}(v) \right) \right) \\
        &\stackrel{(3)}{=} x \Eoplus \psi^{-1} \circ \rieexp _{\Rzero}\left(\frac{\lambda_{\psi(x)}^K}{2} \psi_{*, x}(v) \right) \\
        &\stackrel{(4)}{=} x \Eoplus \rieexp _{\Rzero}\left( \lambda_{\psi(x)}^K \psi_{*, x}(v)  \right) . \\
        \end{aligned}
    \end{equation*}
    The above comes from the following.
    \begin{enumerate}[label=(\arabic*)]
        \item 
        The isometry of $\psi$;
        \item 
        $\rieexp _x ^\mathbb{P} (v) = x \Moplus \rieexp _\Rzero \left(\frac{\lambda_{x}^K}{2} v \right)$;
        \item 
        The isomorphism of $\psi$;
        \item 
        \begin{equation*}
            \begin{aligned}
                \psi^{-1} \circ \rieexp _{\Rzero} \left( v \right)
                &= \psi^{-1} \circ \rieexp ^{\mathbb{P}}_{\Rzero} \left( v \right) \\
                &= \psi^{-1} \circ \psi \circ \rieexp ^{\mathbb{K}}_{\Rzero} \left( 2v \right) \\
                &= \rieexp _{\Rzero} \left( 2v \right).
            \end{aligned}
        \end{equation*}
    \end{enumerate}

    The rest is to calculate $\lambda_{\psi(x)}^K \psi_{*, x}(v)$: 
    \begin{equation*}
        \begin{aligned}
            \lambda_{\psi(x)}^K
            &= \frac{2}{1+ K \frac{\norm{x}^2}{\left(1+\sqrt{1+K\|x\|^2}\right)^2}} \\
            &= \frac{2\left(1+\sqrt{1+K\|x\|^2}\right)^2}{\left(1+\sqrt{1+K\|x\|^2}\right)^2+K\|x\|^2} \\
            &= \frac{2\left(1+\sqrt{1+K\|x\|^2}\right)^2}{ 2 +2 \sqrt{1+K\|x\|^2}+2K\|x\|^2 } \\
            &= \frac{\left(1+\sqrt{1+K\|x\|^2}\right)^2}{ 1 + \sqrt{1+K\|x\|^2}+K\|x\|^2} \\
            &= \frac{\left(1+\sqrt{1+K\|x\|^2}\right)}{\sqrt{1+K\|x\|^2}}, \\
        \end{aligned}
    \end{equation*}
    \begin{equation*}
        \begin{aligned}
            \lambda_{\psi(x)}^K \psi_{*, x}(v) 
            &= \lambda_{\psi(x)}^K \left( \frac{1}{1 + \sqrt{1+K\|x\|^2}} v 
            - \frac{K \inner{x}{v}}{\left(1 + \sqrt{1+K\|x\|^2}\right)^2 \sqrt{1+K\|x\|^2}} x \right) \\
            &= \frac{1}{\sqrt{1+K\|x\|^2}} v 
            - \frac{K \inner{x}{v}}{\left(1 + \sqrt{1+K\|x\|^2}\right) (1+K\|x\|^2)} x .
        \end{aligned}
    \end{equation*}

    \mypara{Logarithmic maps.}  
    \begin{equation*}
        \begin{aligned}
            \rielog _x ^{\mathbb{K}} (y) 
            &\stackrel{(1)}{=} \psi ^{-1} _{*, x} \left( \rielog _{\psi(x)} ^{\mathbb{P}} \psi(y)\right) \\
            &\stackrel{(2)}{=} \psi ^{-1} _{*, x} \left( \frac{2}{ \lambda _{\psi(x)} ^K} \rielog _{\Rzero}\left(-\psi(x) \Moplus \psi(y) \right)\right) \\
            &\stackrel{(3)}{=} \psi ^{-1} _{*, x} \left( \frac{2}{ \lambda _{\psi(x)} ^K} \rielog _{\Rzero}\left(-\psi(x) \Moplus \psi(y)\right)\right) \\
            &\stackrel{(4)}{=} \frac{2}{\lambda_{\psi(x)}^K} \phi_{*, \psi(x)}\left(\rielog _{\Rzero}\left(-\psi(x) \Moplus \psi(y)\right)\right) \\
            &\stackrel{(5)}{=} \frac{1}{\lambda_{\psi(x)}^K} \phi_{*, \psi(x)}\left(\rielog _{\Rzero}\left( -x \Eoplus  y\right)\right) .
        \end{aligned}
    \end{equation*}  
    The above comes from the following.
    \begin{enumerate}[label=(\arabic*)]
        \item 
        The isometry of $\psi$;
        \item 
        $\rielog _{x} ^{\mathbb{P}} (y) = \frac{2}{ \lambda _x ^K} \rielog _{\Rzero} \left(  -x \Moplus y \right)$;
        \item 
        The isomorphism of $\psi$;
        \item 
        The Linearity of differential maps;
        \item 
        \begin{equation*}
            \begin{aligned}
                \rielog _{\Rzero}\left(-\psi(x) \Moplus \psi(y)\right) 
                &= \rielog ^\mathbb{P} _{\Rzero}\left(-\psi(x) \Moplus \psi(y)\right) \\
                &= \frac{1}{2}\rielog ^\mathbb{K} _{\Rzero}\left( \psi^{-1} \left(-\psi(x) \Moplus \psi(y)\right) \right) \\
                &= \frac{1}{2}\rielog ^\mathbb{K} _{\Rzero}\left( -x \Eoplus  y \right).
            \end{aligned}
        \end{equation*}
        
    \end{enumerate}
\end{proof}

\begin{remark}
   \citet[Theorem 9]{mao2024klein} extended the Möbius matrix-vector multiplication  \citep[Lemma 6]{ganea2018hyperbolic} into the Einstein space of the Beltrami--Klein model under $K=-1$, namely $\rieexp _{\Rzero} \left(M \rielog _{\Rzero} (x))\right)$. Although their presented formulations are different from the Möbius one, this theorem indicates that matrix-vector multiplications over these two spaces are identical under any negative curvature. The equality can also be readily observed by \cref{app:prf:eq:log_exp_ln_equation1,app:prf:eq:log_exp_ln_equation2}.
\end{remark}

\bibliography{ref}

\end{document}